\documentclass[10pt,twocolumn,letterpaper]{article}

\usepackage{cvpr}
\usepackage{times}
\usepackage{epsfig}
\usepackage{graphicx}
\usepackage{amsmath}
\usepackage{amssymb}

\usepackage{subfigure}
\usepackage{xcolor}

\usepackage{booktabs}
\usepackage{array}
\usepackage{multirow}

\newcommand{\mycaption}[2]{\caption{\textbf{#1.}\xspace#2}}

\newcommand{\f}{\mathbf{f}}
\newcommand{\bv}{\mathbf{v}}
\newcommand{\bw}{\mathbf{w}}

\usepackage[pagebackref=true,breaklinks=true,letterpaper=true,colorlinks,bookmarks=false]{hyperref}

\cvprfinalcopy 

\def\cvprPaperID{766} 

\ifcvprfinal\fi

\RequirePackage{snapshot}
\begin{document}

\title{Learning Sparse High Dimensional Filters:\\Image Filtering, Dense CRFs and Bilateral Neural Networks}

\author{
Varun Jampani$^1$, Martin Kiefel$^{1, 2}$ and Peter V. Gehler$^{1, 2}$\\
$^1$Max Planck Institute for Intelligent Systems, T{\"u}bingen, 72076, Germany\\
$^2$Bernstein Center for Computational Neuroscience, T{\"u}bingen, 72076, Germany \\
\texttt{\{varun.jampani, martin.kiefel, peter.gehler\}@tuebingen.mpg.de} \\
}

\maketitle

\begin{abstract}
Bilateral filters have wide spread use due to their edge-preserving properties. The common
use case is to manually choose a parametric filter type, usually a Gaussian filter. In this paper,
we will generalize the parametrization and in particular derive a gradient descent algorithm
so the filter parameters can be learned from data. This derivation allows to learn
high dimensional linear filters that operate in sparsely populated feature spaces. We build on
the permutohedral lattice construction for efficient filtering.
The ability to learn more general forms of high-dimensional filters can be used in
several diverse applications.
First, we demonstrate the use in applications where single filter applications are desired for runtime reasons.
Further, we show how this algorithm can be used to learn the pairwise potentials in
densely connected conditional random fields and apply these to different image segmentation tasks.
Finally, we introduce layers of bilateral filters in CNNs and propose
\emph{bilateral neural networks} for the use of high-dimensional
sparse data. This view provides new ways to encode model structure into network architectures. A diverse
set of experiments empirically validates the usage of general forms of filters.
\end{abstract}
\vspace{-0.5cm}

\section{Introduction}

Image convolutions are basic operations for many image processing and computer vision applications. In this paper
we will study the class of bilateral filter convolutions and propose a general image adaptive convolution that can be
learned from data.
 The bilateral filter~\cite{aurich1995non, smith97ijcv, tomasi1998bilateral} was originally introduced for the
 task of image denoising as an edge preserving filter. Since the bilateral filter contains the spatial
 convolution as a special case, we will in the following directly state the general case. Given an image $\bv=(\bv_1,\ldots,\bv_n), \bv_i\in\mathbb{R}^c$
 with $n$ pixels and $c$ channels,
 and for every pixel $i$, a $d$ dimensional feature vector $\f_i\in\mathbb{R}^d$ (\eg, the $(x,y)$ position in the image $\f_i=(x_i,y_i)^{\top}$).
 The bilateral filter then computes 
\begin{equation}
  \bv'_i = \sum_{j=1}^n \bw_{\f_i,\f_j} \bv_j.\label{eq:bilateral}
  \vspace{-0.2cm}
\end{equation}
for all $i$. Almost the entire literature refers to the bilateral filter as a synonym of the Gaussian parametric form
$\bw_{\f_i,\f_j} = \exp{(-\frac{1}{2}(\f_i-\f_j)^{\top}\Sigma^{-1}(\f_i-\f_j))}$. The features $\f_i$ are most commonly chosen to be
position $(x_i,y_i)$ and color $(r_i,g_i,b_i)$ or pixel intensity. To appreciate the edge-preserving effect of
the bilateral filter, consider the five-dimensional feature $\f={(x,y,r,g,b)}^{\top}$.
Two pixels $i,j$ have a strong influence $\bw_{\f_i,\f_j}$ on each other only if they are close in
position \emph{and} color. At edges the color changes, therefore pixels lying on opposite sides have low
influence and thus this filter does not blur across edges. This behaviour is
sometimes referred to as ``image adaptive'', since the filter has a different shape when evaluated at different
locations in the image. More precisely, it is the projection of the filter to the two-dimensional
image plane that changes, the filter values $\bw_{\f,\f'}$ do not change. The filter itself can
be of $c$ dimensions $\bw_{\f_i,\f_j}\in\mathbb{R}^c$,
in which case the multiplication in Eq.~\eqref{eq:bilateral} becomes an inner product. For the Gaussian case the filter can be
applied independently per channel. For an excellent review of
image filtering we refer to~\cite{milanfar2011tour}.


The filter operation of Eq.~\eqref{eq:bilateral} is a sparse high-dimensional convolution,
a view first developed in~\cite{paris2006fast}. An image $v$ is
not sparse in the spatial domain, we observe pixels values for all locations $(x,y)$. However, when
pixels are understood in a higher dimensional feature space, \eg, $(x,y,r,g,b)$, the image becomes a sparse signal,
since the $r,g,b$ values lie scattered in this five-dimensional space.
This view on filtering is the key difference of the bilateral filter compared to the common spatial convolution.
An image edge is not ``visible'' for a filter in the spatial domain alone, whereas in the 5D space it is.
The edge-preserving behaviour is possible due to the higher dimensional operation. Other data can naturally
be understood as sparse signals, \eg, 3D surface points.


The contribution of this paper is to propose a general and learnable sparse high dimensional convolution.
Our technique builds on efficient algorithms
that have been developed to approximate the Gaussian bilateral filter and re-uses them for
more general high-dimensional filter operations. Due to its practical importance (see related work in Sec.~\ref{sec:related}) several efficient algorithms for computing Eq.~\eqref{eq:bilateral}
have been developed, including the bilateral grid~\cite{paris2006fast}, Gaussian KD-trees~\cite{adams2009gaussian}, and the permutohedral
lattice~\cite{adams2010fast}. The design goal for these algorithms was to provide a) fast runtimes and b) small approximation errors for
the Gaussian filter case. The key insight of this paper is to use the permutohedral lattice and use it not as an approximation of a
predefined kernel but to freely parametrize its values. We relax the separable Gaussian filter case from~\cite{adams2010fast} and show
how to compute gradients of the convolution (Sec.~\ref{sec:learning}) in lattice space. This enables learning the filter from data.


This insight has several useful consequences. We discuss applications where the bilateral filter has been used before: image
filtering (Sec.~\ref{sec:filtering}) and CRF inference (Sec.~\ref{sec:densecrf}). Further we will demonstrate how the
free parametrization of the filters enables us to use them in Deep convolutional neural networks (CNN) and allow convolutions that go beyond the regular
spatially connected receptive fields (Sec.~\ref{sec:bnn}). For all domains, we present various empirical evaluations with
a wide range of applications.

\vspace{-0.1cm}
\section{Related Work}\label{sec:related}

We categorize the related work according to the three different generalizations of this work.

%
\vspace{-0.3cm}
\paragraph{Image Adaptive Filtering:} The literature in this area is rich
and we can only provide a brief overview. Important classes of image adaptive filters
include the bilateral filters~\cite{aurich1995non, tomasi1998bilateral,
smith97ijcv}, non-local means~\cite{buades2005non, awate2005higher}, locally
adaptive regressive kernels~\cite{takeda2007kernel}, guided image
filters~\cite{he2013guided} and propagation filters~\cite{rick2015propagated}.
The kernel least-squares regression problem can serve as a unified view of many
of them~\cite{milanfar2011tour}. In contrast to the present work that learns
the filter kernel using supervised learning, all these filtering schemes use a
predefined kernel.
Because of the importance of the bilateral filtering to many applications in
image processing, much effort has been devoted to derive fast algorithms; most
notably~\cite{paris2006fast, adams2010fast, adams2009gaussian, gastal2012adaptive}.
Surprisingly, the
only attempt to learn the bilateral filter we found is~\cite{hu2007trained} that
casts the learning problem in the spatial domain by rearranging pixels. However,
the learned filter does not necessarily obey the full region of influence of a
pixel as in the case of a bilateral filter.
The bilateral filter also has been proposed to regularize a large set of
applications in~\cite{barron2015bilateral, barron2015defocus} and the respective
optimization problems are parametrized in a bilateral space. They provide
gradients to use the filters as part of a learning system but unlike this work
restrict the filters to be Gaussian.

\vspace{-0.3cm}
\paragraph{Dense CRF:} The key observation of~\cite{krahenbuhl2012efficient} is
that mean-field inference update steps in densely connected CRFs with Gaussian
edge potentials require Gaussian bilateral filtering operations. This enables
tractable inference through the application of a fast filter implementation
from~\cite{adams2010fast}. This quickly found wide-spread use, \eg, the combination of CNNs with
a dense CRF is among the best performing segmentation
models~\cite{chen2014semantic, zheng2015conditional, bell2015minc}. These works
combine structured prediction frameworks on top of CNNs, to model the
relationship between the desired output variables thereby significantly
improving upon the CNN result. Bilateral neural networks, that are presented in
this work, provide a principled framework for encoding the output relationship,
using the feature transformation inside the network itself thereby alleviating
some of the need for later processing. Several works~\cite{krahenbuhl2013parameter,
domke2013learning, kiefel2014human} demonstrate how to learn free parameters of
the dense CRF model. However, the parametric form of the pairwise term always remains a
Gaussian. Campbell \etal~\cite{campbell2013fully} embed complex pixel
dependencies into an Euclidean space and use a Gaussian filter for pairwise connections.
This embedding is a pre-processing step and can not directly be learned. In
Sec.~\ref{sec:densecrf} we will discuss how to learn the pairwise
potentials, while retaining the efficient inference strategy of~\cite{krahenbuhl2012efficient}.

\vspace{-0.3cm}
\paragraph{Neural Networks:} In recent years, tremendous progress in a wide
range of computer vision applications has been observed due to the use of
CNNs. Most CNN architectures use
spatial convolution layers, which have fixed local receptive fields. This work
suggests to replace these layers with bilateral filters, which have a varying
spatial receptive field depending on the image content. The equivalent
representation of the filter in a higher dimensional space leads to sparse
samples that are handled by a permutohedral lattice data structure. Similarly,
Bruna \etal~\cite{bruna2013spectral} propose convolutions on irregularly sampled
data. Their graph construction is closely related to the high-dimensional
convolution that we propose and defines weights on local neighborhoods of nodes.
However, the structure of the graph is bound to be fixed and it is not
straightforward to add new samples. Furthermore, re-using the same filter
among neighborhoods is only possible with their costly spectral construction.
Both cases are handled naturally by our sparse convolution.
Jaderberg \etal~\cite{jaderberg2015spatial} propose a spatial
transformation of signals within the neural network to learn invariances for a
given task. The work of~\cite{ionescu2015matrix} propose matrix backpropagation
techniques which can be used to build specialized structrual layers such as normalized-cuts.
Graham \etal~\cite{graham2015sparse} propose extensions from 2D CNNs
to 3D sparse signals. Our work enables sparse 3D filtering as a special case,
since we use an algorithm that allows for even higher dimensional data.


\vspace{-0.1cm}
\section{Learning Sparse High Dimensional Filters}\label{sec:learning}
In this section, we describe the main technical contribution of
this work, we generalize the permutohedral convolution~\cite{adams2010fast} and show how
the filter can be learned from data.


Recall the form of the bilateral convolution from Eq.~\eqref{eq:bilateral}. A naive
implementation would compute for every pixel $i$ all associated filter values $\bw_{\f_i,\f_j}$ and
perform the summation independently. The view of $\bw$ as a linear filter in a higher dimensional
space, as proposed by~\cite{paris2006fast}, opened the way for new algorithms. Here,
we will build on the permutohedral lattice convolution developed
in Adams \etal~\cite{adams2010fast} for approximate Gaussian filtering.



\subsection{Permutohedral Lattice Convolutions}
We first review the permutohedral lattice convolution for
Gaussian bilateral filters from Adams \etal~\cite{adams2010fast} and
describe its most general case.

\begin{figure}[t]
\begin{center}
\centerline{\includegraphics[width=\columnwidth]{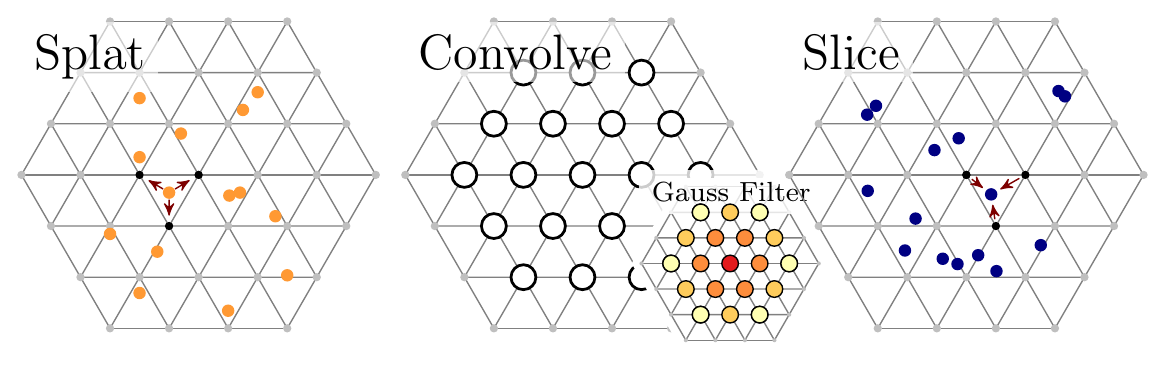}}
  \vspace{-0.5em}
  \mycaption{Schematic of the permutohedral convolution}
  {Left: splatting the input
  points (orange) onto the lattice corners (black); Middle: The extent of a filter on the lattice
   with a $s=2$ neighborhood (white circles), for reference we show a Gaussian filter,
   with its values color coded. The general case has a free scalar/vector parameter per circle.
    Right: The result of the convolution at the lattice corners (black) is projected back
    to the output points (blue). Note that in general the output and input points may be different. \label{fig:ops}}
  \vspace{-1.5em}
\end{center}
\vspace{-0.6cm}
\end{figure}

\begin{figure}[t]
  \centering
  \subfigure[\scriptsize Sample Image]{%
    \includegraphics[width=.23\columnwidth]{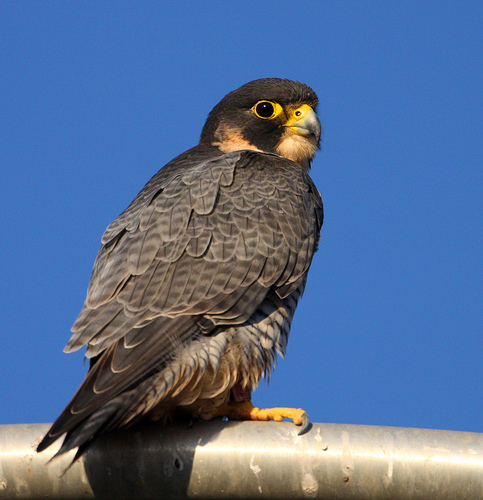}\label{fig:bird}
  }
  \subfigure[\scriptsize Position]{%
    \includegraphics[width=.23\columnwidth]{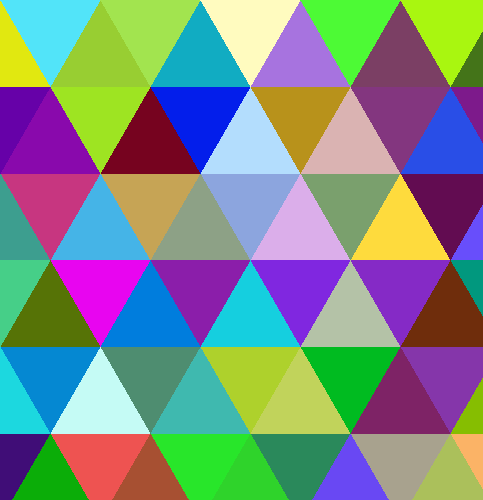}\label{fig:bird_lattice_1}
  }
  \subfigure[\scriptsize Color]{%
    \includegraphics[width=.23\columnwidth]{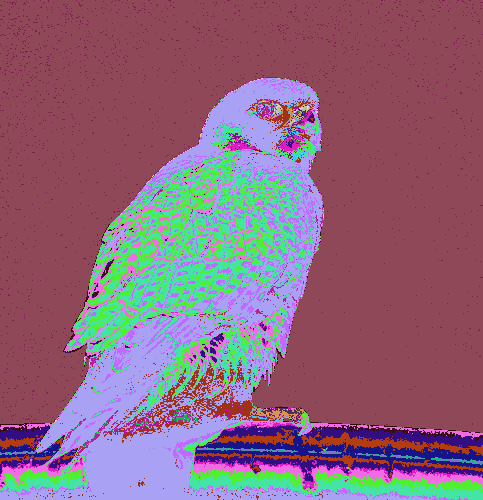}\label{fig:bird_lattice_2}
  }
  \subfigure[\scriptsize Position, Color]{%
    \includegraphics[width=.23\columnwidth]{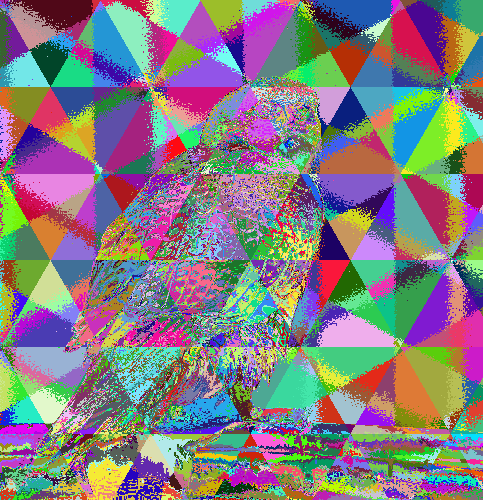}\label{fig:bird_lattice_3}
  }
  \mycaption{Visualization of the Permutohedral Lattice}
  {(a) Input image; Lattice visualizations for different feature spaces: (b) 2D position features: $0.01(x, y)$,
  (c) color features: $0.01(r, g, b)$ and (d) position and color features: $0.01(x, y, r, g, b)$. All pixels
  falling in the same simplex cell are shown with the same color.}
\label{fig:lattice}
\vspace{-0.4cm}
\end{figure}

As before, we assume that every image pixel $i$ is associated with a $d$-dimensional
feature vector $\f_i$. Gaussian bilateral filtering using a permutohedral lattice
approximation involves 3 steps. We begin with an overview of the algorithm, then discuss each step
in more detail in the next paragraphs. Fig.~\ref{fig:ops} schematically shows
the three operations for 2D features. First, interpolate the image signal on the
$d$-dimensional grid plane of the permutohedral lattice, which is called
\emph{splatting}. A permutohedral lattice is the tessellation of space into permutohedral simplices.
We refer to~\cite{adams2010fast} for details of the lattice construction and its properties.
In Fig.~\ref{fig:lattice}, we visualize
the permutohedral lattice in the image plane, where every simplex cell receives a different color. All pixels of the same
lattice cell have the same color. Second,
\emph{convolve} the signal on the lattice.
And third, retrieve the result by interpolating the signal at the original
$d$-dimensional feature locations, called \emph{slicing}. For example, if
the features used are a combination of position and color $\f_i = (x_i, y_i, r_i, g_i,
b_i)^{\top}$, the input signal is mapped into the 5D cross product space of position and
color and then convolved with a 5D tensor. The result is then mapped back to the
original space. In practice we use a feature scaling $D\f$ with a diagonal matrix $D$ and use separate scales
for position and color features. The scale determines the distance of points and thus the size
of the lattice cells.
More formally, the computation is written by $\mathbf{v'} =
S_{\text{slice}}BS_{\text{splat}} \mathbf{v}$ and all involved matrices are
defined below. For notational convenience we will assume scalar input signals $v_i$, the vector valued case
is analogous, the lattice convolution changes from scalar multiplications to inner products.

\vspace{-0.3cm}
\paragraph{Splat:} The splat operation (cf.\ left-most image in
Fig.~\ref{fig:ops}) finds the enclosing simplex in $\mathcal{O}(d^2)$ on the
lattice of a given pixel feature $\f_i$ and distributes its value $v_i$ onto the corners of the
simplex. How strong a pixel contributes to a corner $j$ is defined by its
barycentric coordinate $b_{i,j}\in\mathbb{R}$ inside the simplex. Thus, the value $l_j \in \mathbb{R}$ at a lattice point
$j$ is computed by summing over all enclosed input points; more precisely, we
define an index set $J_i$ for a pixel $i$, which contains all the lattice points
$j$ of the enclosing simplex
\vspace{-0.7em}
\begin{equation}
  \label{eq:splat}
  \mathbf{\ell} = S_{\text{splat}} \mathbf{v};%
  {(S_{\text{splat}})}_{j, i} = b_{i, j}, \text{ if } j\in J_i, \text{ otherwise } 0.
\vspace{-0.5em}
\end{equation}

\vspace{-0.3cm}
\paragraph{Convolve:} The permutohedral convolution is defined on the
lattice neighborhood $N_s(j)$ of lattice point $j$, \eg, only $s$ grid hops away. More formally
\vspace{-0.7em}
\begin{equation}
  \label{eq:blur}
  \mathbf{\ell'} = B \mathbf{\ell}; {(B)}_{j', j} =  w_{j, j'},%
  \text{ if } j' \in N_s(j), \text{ otherwise } 0.
\vspace{-0.5em}
\end{equation}
An illustration of a two-dimensional permutohedral filter is shown in Fig.~\ref{fig:ops} (middle).
Note that we already presented the convolution in the general form that we will make
use of. The work of~\cite{adams2010fast} chooses the filter weights such that
the resulting operation approximates a Gaussian blur, which is illustrated in Fig.~\ref{fig:ops}.
Further, the algorithm of~\cite{adams2010fast} takes advantage of the
separability of the Gaussian kernel. Since we are interested in the most general case, we
extended the convolution to include non-separable filters $B$.

\vspace{-0.3cm}
\paragraph{Slice:} The slice operation (cf.\ right-most image in
Fig.~\ref{fig:ops}) computes an output value $v'_{i'}$ for an output pixel $i'$
again based on its barycentric coordinates $b_{i, j}$ and sums over the corner
points $j$ of its lattice simplex
\vspace{-0.7em}
\begin{equation}
  \label{eq:slice}
  \mathbf{v'} = S_{\text{slice}} \mathbf{\ell'}; %
  {(S_{\text{slice}})}_{i, j} = b_{i, j}, \text{ if } j\in J_i, \text{ otherwise } 0
\vspace{-0.5em}
\end{equation}

The splat and slice operations take a role of an interpolation between the
different signal representations: the irregular and sparse distribution of
pixels with their associated feature vectors and the regular structure of the
permutohedral lattice points. Since high-dimensional spaces are usually sparse,
performing the convolution densely on all lattice points is inefficient. So, for
speed reasons, we keep track of the populated lattice points using a hash table
and only convolve at those locations.

\subsection{Learning Permutohedral Filters}\label{sec:backprop}
%
The \emph{fixed} set of filter weights $\mathbf{w}$ from~\cite{adams2010fast} in
Eq.~\ref{eq:blur} are designed to approximate a Gaussian filter. However, the
convolution kernel $\mathbf{w}$ can naturally be understood as a general filtering
operation in the permutohedral lattice space with free parameters. In the exposition above we
already presented this general case. As we will show in more detail later, this modification has non-trivial consequences
for bilateral filters, CNNs and probabilistic graphical models.

The size of the neighborhood $N_s(k)$ for the blur in Eq.~\ref{eq:blur} compares
to the filter size of a spatial convolution. The filtering kernel of a common
spatial convolution that considers $s$ points to either side in all dimensions
has ${(2s + 1)}^d\in\mathcal{O}(s^d)$ parameters. A comparable filter on the
permutohedral lattice with an $s$ neighborhood is specified by ${(s+1)}^{d+1} -
s^{d+1} \in\mathcal{O}(s^d)$ elements (cf. Sec.~\ref{sec:permconv}). Thus, both
share the same asymptotic size.

By computing the gradients of the filter elements we enable the use
of gradient based optimizers, \eg, backpropagation for CNN in the same
way that spatial filters in a CNN are learned. The gradients
with respect to $\mathbf v$ and the filter weights in $B$ of a
scalar loss $L$ are:
\vspace{-0.5em}
\begin{eqnarray}
  \frac {\partial L} {\partial \mathbf{v}} &=&
  S'_{\text{splat}} %
  B' %
  S'_{\text{slice}} %
  \frac {\partial L} {\partial \mathbf{v}'}, \label{eq:dv}\\
  \frac {\partial L} {\partial {(B)}_{i, j}} &=& %
  {\left(S'_{\text{slice}} \frac {\partial L} {\partial \mathbf{v}}\right)}_i %
  {(S_{\text{splat}} \mathbf{v})}_j.\label{eq:db}%
\vspace{-0.3em}
\end{eqnarray}
Both gradients are needed during backpropagation and in experiments,
we use stochastic backpropagation for learning the filter kernel.
The permutohedral lattice convolution is parallelizable, and scales linearly with the
filter size. Specialized implementations run at interactive speeds in image
processing applications~\cite{adams2010fast}.
Our implementation allows arbitrary filter parameters and the computation
of the gradients on both CPU and GPU. It is also integrated into the caffe deep learning
framework~\cite{jia2014caffe}.

%
%

\begin{figure*}[t!]
  \centering
  \subfigure{%
   \raisebox{2.0em}{
    \includegraphics[width=.12\columnwidth]{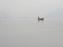}\label{fig:color_given}
   }
  }
  \subfigure{%
    \includegraphics[width=.36\columnwidth]{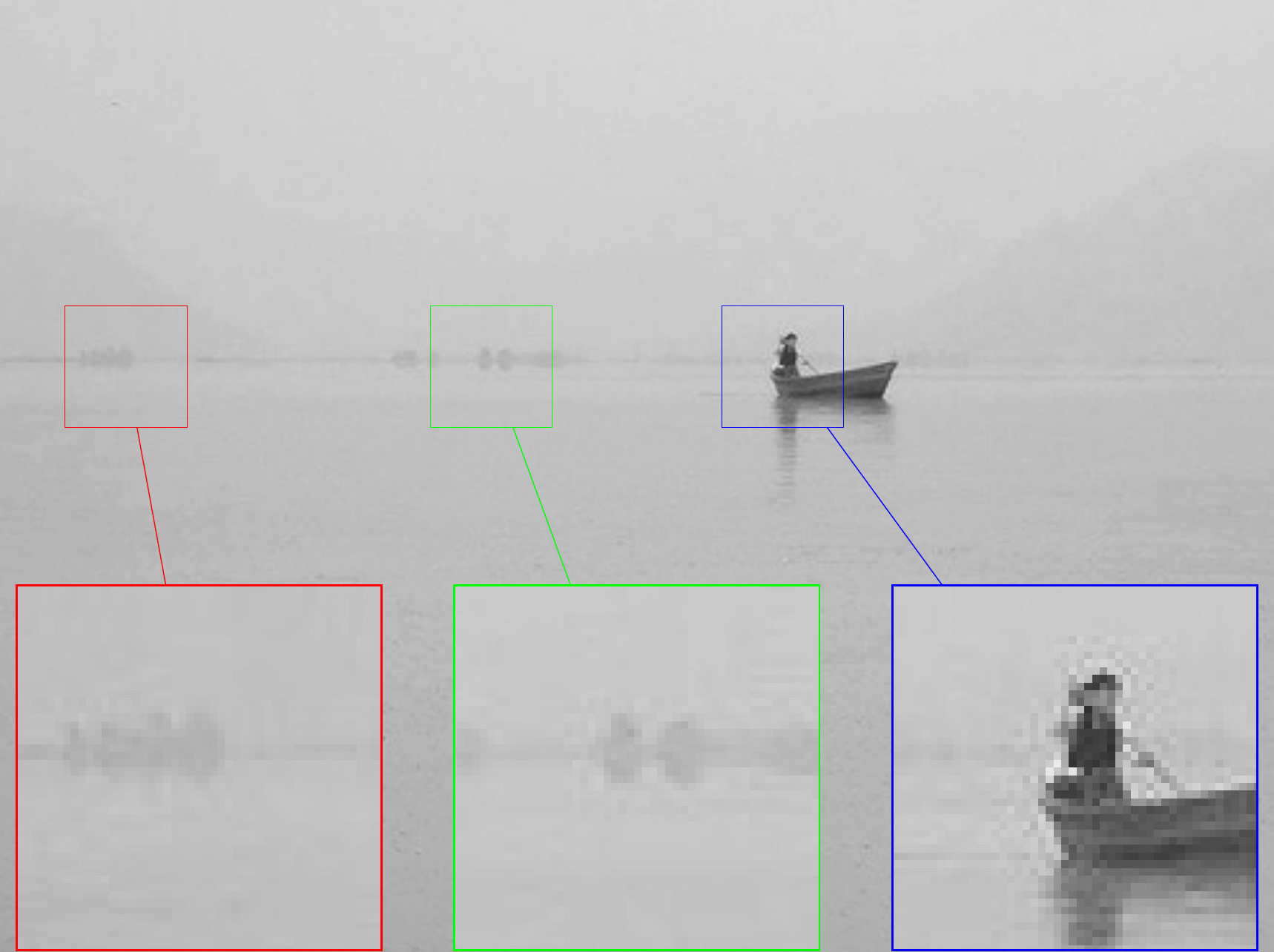}\label{fig:color_guidance}
  }
  \subfigure{%
    \includegraphics[width=.36\columnwidth]{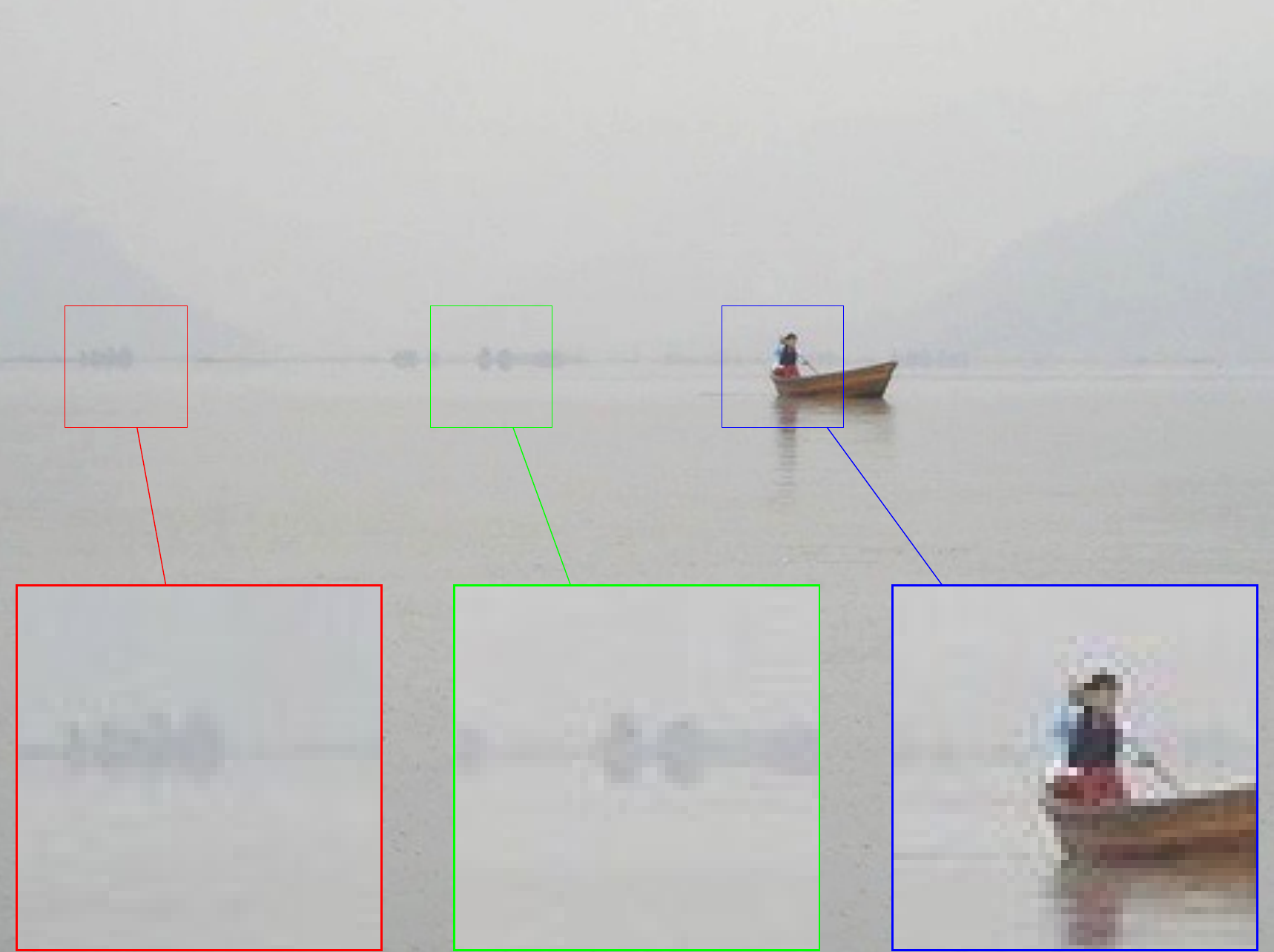}\label{fig:color_original}
  }
  \subfigure{%
    \includegraphics[width=.36\columnwidth]{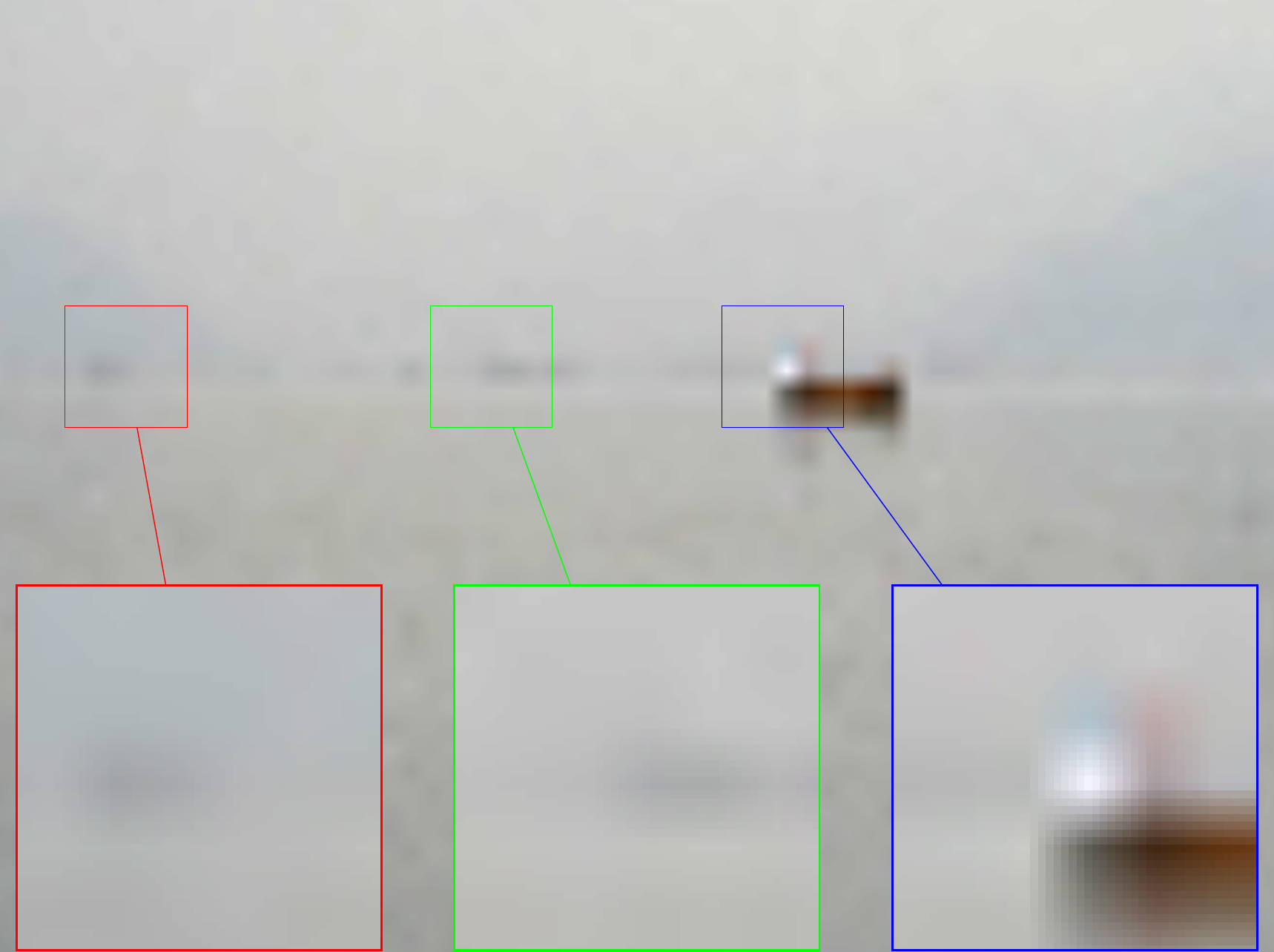}\label{fig:color_bicubic}
  }
  \subfigure{%
    \includegraphics[width=.36\columnwidth]{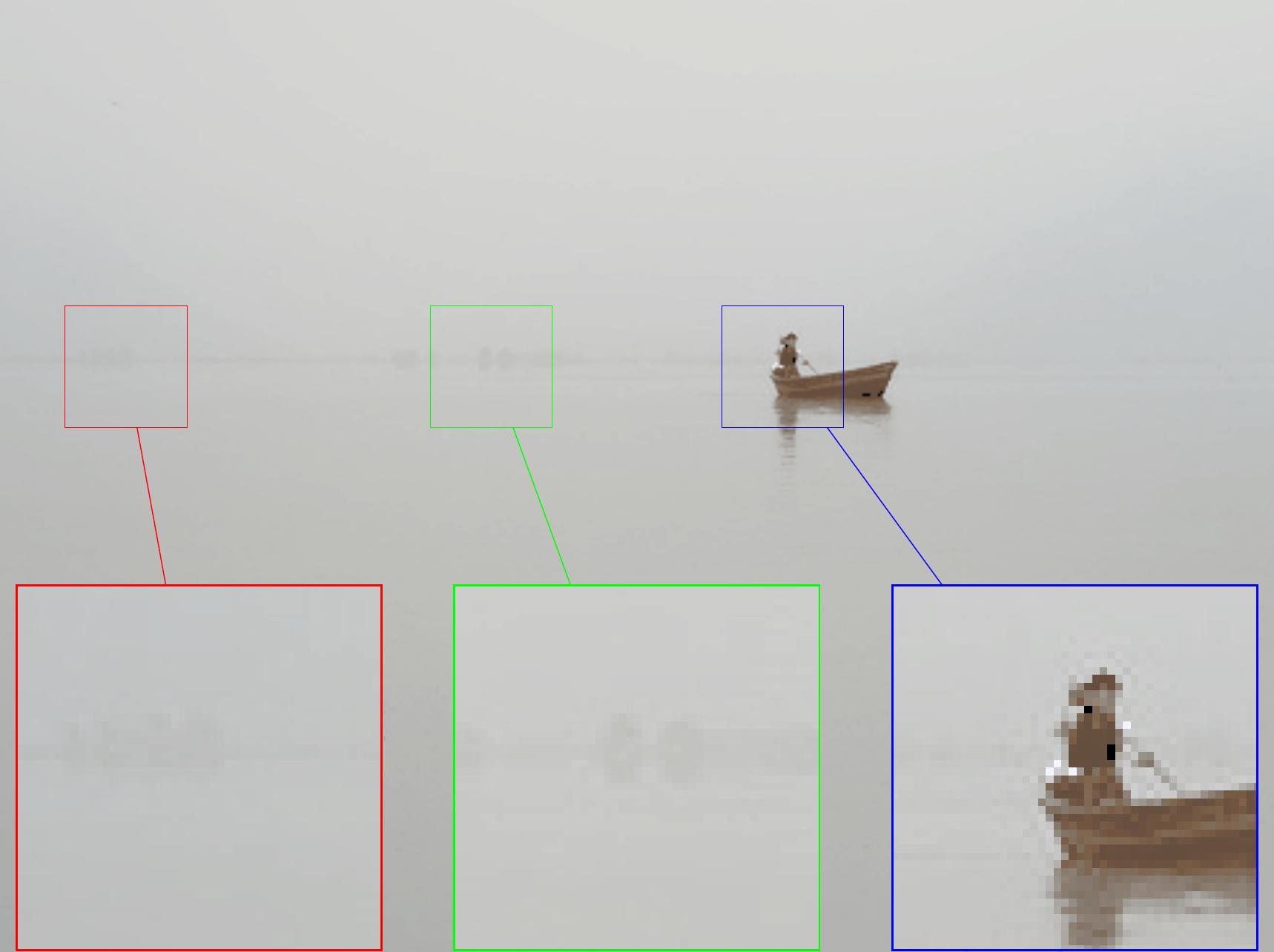}\label{fig:color_gauss}
  }
  \subfigure{%
    \includegraphics[width=.36\columnwidth]{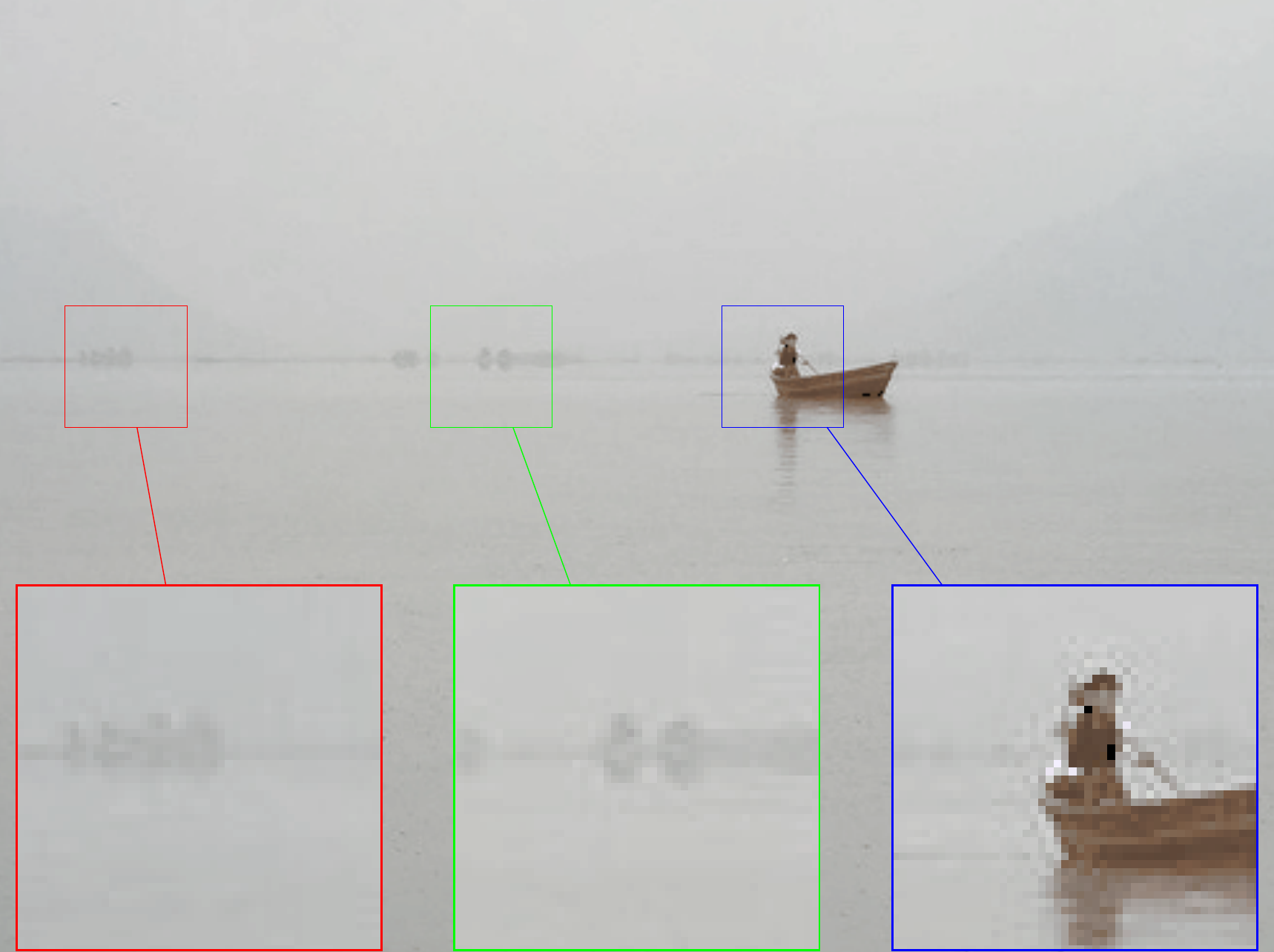}\label{fig:color_learnt}
  }\\[-2ex]
  \setcounter{subfigure}{0}
  \subfigure[Input]{%
  \raisebox{2.0em}{
    \includegraphics[width=.12\columnwidth]{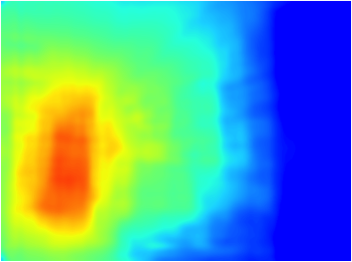}\label{fig:depth_given}
   }
  }
  \subfigure[Guidance]{%
    \includegraphics[width=.36\columnwidth]{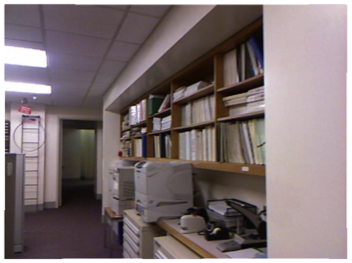}\label{fig:depth_guidance}
  }
   \subfigure[Ground Truth]{%
    \includegraphics[width=.36\columnwidth]{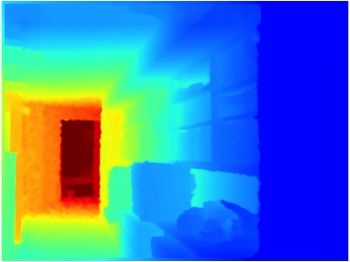}\label{fig:depth_gt}
  }
  \subfigure[Bicubic Interpolation]{%
    \includegraphics[width=.36\columnwidth]{figures/depth_bicubic.png}\label{fig:depth_bicubic}
  }
  \subfigure[Gauss Bilateral]{%
    \includegraphics[width=.36\columnwidth]{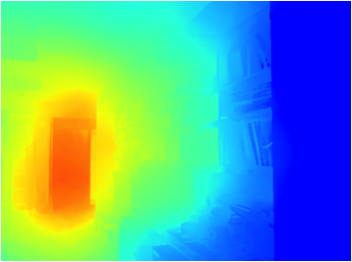}\label{fig:depth_gauss}
  }
  \subfigure[Learned Bilateral]{%
    \includegraphics[width=.36\columnwidth]{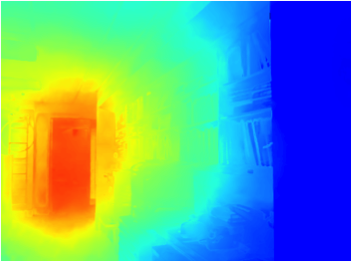}\label{fig:depth_learnt}
  }
  \mycaption{Guided Upsampling}{Color (top) and depth (bottom) $8\times$ upsampling results
  using different methods (best viewed on screen).}
  \vspace{-0.5cm}
\label{fig:upsample_visuals}
\end{figure*}

\section{Single Bilateral Filter Applications}\label{sec:filtering}

In this section we will consider the problem of joint bilateral upsampling~\cite{kopf2007joint} as a prominent
instance of a single bilateral filter application. See~\cite{paris2009bilateral} for a recent overview
of other bilateral filter applications. Further experiments on image denoising and
3D body mesh denoising are included in Sec.~\ref{sec:mesh_denoising}, together
with details about exact experimental protocols and more visualizations.


\subsection{Joint Bilateral Upsampling}

A typical technique to speed up computer vision algorithms is to compute results on a
lower scale and upsample the result to the full resolution. This upsampling
step may use the original resolution image as a guidance image. A joint bilateral upsampling approach
for this problem setting was developed in~\cite{kopf2007joint}. We describe the procedure
for the example of upsampling a color image. Given a high resolution gray scale image (the guidance image) and
the same image on a lower resolution but in colors, the task is to upsample the color image to the
same resolution as the guidance image. Using the permutohedral lattice, joint bilateral upsampling proceeds by
splatting the color image into the lattice, using 2D position and 1D intensity as features and the 3D RGB values
as the signal. A convolution is applied in the lattice and the result is read out at the pixels of the high
resolution image, that is using the 2D position and intensity of the guidance image. The possibility of
reading out (slicing) points that are not necessarily the input points is an appealing feature of the permutohedral lattice
convolution.
%

\begin{table}[t!]
  \scriptsize
  \centering
    \begin{tabular}{l c c c c c}
      \toprule
     & \textbf{Upsampling factor} & \textbf{Bicubic} & \textbf{Gaussian} & \textbf{Learned} \\ [0.1cm]
      \midrule
     \multicolumn{2}{l}\textbf{Color Upsampling (PSNR)} & & &\\
     & \textbf{2x} & 24.19 / 30.59 & 33.46 / 37.93  & \textbf{34.05 / 38.74}  \\
     & \textbf{4x} & 20.34 / 25.28 & 31.87 / 35.66  & \textbf{32.28 / 36.38}  \\
     & \textbf{8x} & 17.99 / 22.12 & 30.51 / 33.92  & \textbf{30.81 / 34.41}  \\
     & \textbf{16x} & 16.10 / 19.80 & 29.19 / 32.24  & \textbf{29.52 / 32.75}  \\
      \midrule
     \multicolumn{2}{l}\textbf{Depth Upsampling (RMSE)} & & &\\
     & \textbf{8x} & 0.753 & 0.753 & \textbf{0.748}  \\
      \bottomrule
      \\
    \end{tabular}
      \vspace{-0.2cm}
    \mycaption{Joint bilateral upsampling} {(top) PSNR values corresponding to various upsampling factors and
    upsampling strategies on the test images of the Pascal VOC12 segmentation / high-resolution 2MP dataset;
    (bottom) RMSE error values corresponding to upsampling depth images estimated using~\cite{eigen2014depth}
    computed on the test images from the NYU depth dataset~\cite{silberman2012indoor}.}
\label{tbl:upsample}
\vspace{-0.4cm}
\end{table}

\vspace{-0.2cm}
\subsubsection{Color Upsampling}

For the task of color upsampling, we compare the Gaussian bilateral filter~\cite{kopf2007joint} against
a learned generalized filter. We experimented with two different datasets: One is Pascal VOC2012 segmentation dataset~\cite{voc2012segmentation}
using the pre-defined train, validation and test splits, and other is 200 high-resolution (2MP) image dataset collected using
Google image search~\cite{google_images} with 100 train, 50 validation and 50 test images. For training we use the mean
squared error (MSE) criterion and perform stochastic gradient descent with a
momentum term of $0.9$, and weight decay of $0.0005$. The learning rate and feature scales $D$  have
been cross-validated using the validation split. The Gaussian and the learned filter have the same support.
In Table~\ref{tbl:upsample} we report result in terms of PSNR for the upsampling
factors $2\times, 4\times, 8\times$ and $16\times$.
We compare a standard bicubic interpolation, that does not use a guidance image, the Gaussian
bilateral filter case (with feature scales optimized on the validation set), and the learned filter. For all upsampling factors,
we observe that joint bilateral Gaussian upsampling outperforms bicubic interpolation. Learning a general convolution
further improves over the Gaussian filter case. A result of the upsampling is shown in Fig.~\ref{fig:upsample_visuals} and
more results are included in Sec.~\ref{sec:color_upsampling}. The learned filter recovers finer details in the images
%

\vspace{-0.2cm}
\subsubsection{Depth  Upsampling}
We use the work on recovering depth estimates from single RGB images~\cite{eigen2014depth} to
validate the learned filter on another domain for the task of joint upsampling. The dataset
of~\cite{silberman2012indoor} provides a benchmark for this task and comes with pre-defined
train, validation and test splits that we re-use. The approach of~\cite{eigen2014depth} is a CNN
model that produces a result at 1/4th of the input resolution due to down-sampling operations in
max-pooling layers. Furthermore, the authors downsample the $640\times 480$ images to $320\times 240$
as a pre-processing step before CNN convolutions. The final depth result is bicubic interpolated to the
original resolution. It is this interpolation that we replace with a Gaussian and learned joint
bilateral upsampling. The features are five-dimensional position
and color information from the high resolution input image. The filter is learned using the same protocol
as for color upsampling minimizing MSE prediction error. The quantitative results
are shown in Table~\ref{tbl:upsample}, the Gaussian filter performs equal to the bicubic interpolation ($p$-value 0.311),
the learned filter is better ($p$-value 0.015). Qualitative results are shown in Fig~\ref{fig:upsample_visuals},
both joint bilateral upsampling respect image edges in the result. For this~\cite{ferstl2015b} and other tasks
specialized interpolation algorithms exist, \eg, deconvolution networks~\cite{zeiler2010deconvolutional}. Part of future work
is to equip these approaches with bilateral filters.
%
%

\vspace{-0.1cm}
\section{Learning Pairwise Potentials in Dense CRFs}\label{sec:densecrf}
The bilateral convolution from Sec.~\ref{sec:learning} generalizes
the class of dense CRF models for which the mean-field inference from~\cite{krahenbuhl2012efficient}
applies. The dense CRF models have found wide-spread use in various
computer vision applications~\cite{sun2013fully, bell2014intrinsic, zheng2015conditional, vineet12eccv, vineet12emmcvpr, bell2015minc}.
Let us briefly review the dense CRF and then discuss its generalization.

 \subsection{Dense CRF Review}
Consider a fully pairwise connected CRF with discrete variables over each pixel in the image.
For an image with $n$ pixels, we have random variables $\mathbf{X} = \{X_1,X_2,\dots,X_n\}$ with discrete values
$\{x_1,\dots,x_n\}$ in the label space $x_i\in\{1,\ldots,\mathcal{L}\}$. The dense CRF model has
unary potentials $\psi_u(x)\in\mathbb{R}^{\mathcal{L}}$, \eg, these can be the output of CNNs.
The pairwise potentials in~\cite{krahenbuhl2012efficient} are of the form $\psi_p(x_i,x_j) = \mu(x_i,x_j) k(\f_i,\f_j)$
where $\mu$ is a label compatibility matrix, $k$ is a Gaussian kernel $k(\f_i,\f_j)=\exp(-(\f_i-\f_j)^{\top}\Sigma^{-1}(\f_i-\f_j))$ and
the vectors $\f_i$ are feature vectors, just alike the ones used for the bilateral filtering, \eg, $(x,y,r,g,b)$.
The Gibbs distribution for an image $v$ thus reads $p(x|v)\propto\exp(-\sum_i\psi_u(x_i)-\sum_{i> j}\psi_p(x_i,x_j))$.
Because of dense connectivity, exact MAP or marginal inference is intractable. The main result
of~\cite{krahenbuhl2012efficient} is to derive the mean-field approximation of this model and to relate it
to bilateral filtering which enables tractable approximate inference. 
Mean-field approximates the model $p$ with a fully factorized distribution $q=\prod_i q_i(x_i)$
and solves for $q$ by minimizing their KL divergence $KL(q||p)$.
This results in a fixed point equation which
can be solved iteratively $t=0,1,\ldots$ to update the marginal distributions $q_i$, 
\begin{equation}
q^{t+1}_i(x_i) = \frac{1}{Z_i} \exp\{-\psi_u(x_i) -\underbrace{\sum_{j \ne i} \sum_{x_j \in \mathcal{L}} \psi_p(x_i,x_j) q^{t}_j(x_j)}_\text{bilateral filtering}\}.
\label{eq:mfupdate}
\end{equation}
\vspace{-0.5cm}

for all $i$. Here, $Z_i$ ensures normalizations of the distribution $q_i$.


\subsection{Learning Pairwise Potentials}

The proposed bilateral convolution generalizes the class of potential functions $\psi_p$, since
they allow a richer class of kernels $k(\f_i,\f_j)$ that furthermore can be learned from data. So far, all
dense CRF models have used Gaussian potential functions $k$, we replace it with the
general bilateral convolution and learn the parameters of kernel $k$, thus in effect
learn the pairwise potentials of the dense CRF. This retains the desirable properties of this
model class -- efficient inference through mean-field and the feature dependency of the pairwise
potential. The benefit of the Gaussian pairwise potentials is to encode similarity between pixels
through distance and appearance, in other words, pixels with similar $(r,g,b)$ values are likely
to be of the same class. In order to learn the form of the pairwise potentials $k$ we make use of the
gradients for filter parameters in $k$ and
use back-propagation through the mean-field iterations~\cite{domke2013learning, li2014mean} to learn them.

The work of~\cite{krahenbuhl2013parameter} derived gradients to learn the feature scaling $D$ but not the
form of the kernel $k$, which still was Gaussian. In~\cite{campbell2013fully}, the features $f_i$ were
derived using a non-parametric embedding into a Euclidean space and again a Gaussian
kernel was used. The computation of the embedding was a pre-processing step, not integrated in a
end-to-end learning framework. Both aforementioned works are generalizations that are orthogonal
to our development and can be used in conjunction.


\subsection{Experimental Evaluation}

We evaluate the effect of learning more general forms of potential functions on two
pixel labeling tasks, semantic segmentation of VOC data~\cite{voc2012segmentation} and material
classification~\cite{bell2015minc}. We use pre-trained models from the literature
and empirically evaluate the relative change when learning the pairwise potentials. For both
the experiments, we use multinomial logistic classification loss and learn the filters via
back-propagation.

\begin{figure}[t]
  \centering
  \subfigure{%
    \includegraphics[width=.23\columnwidth]{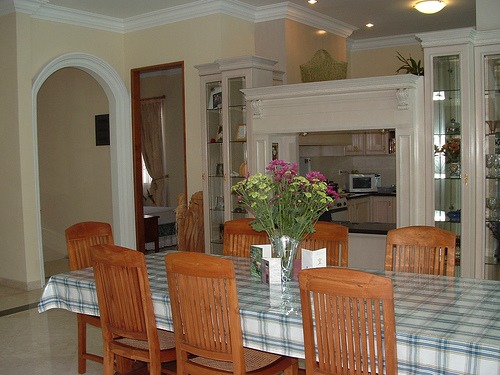} \label{fig:seg_given}
  }
  \subfigure{%
    \includegraphics[width=.23\columnwidth]{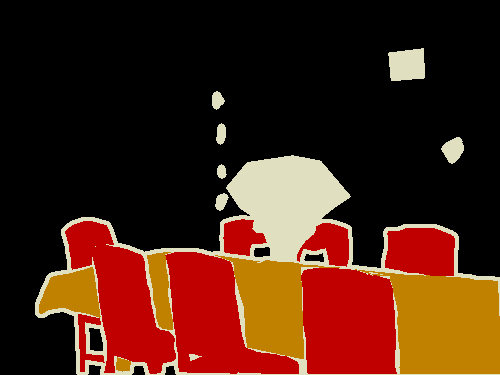} \label{fig:seg_gt}
  }
  \subfigure{%
    \includegraphics[width=.23\columnwidth]{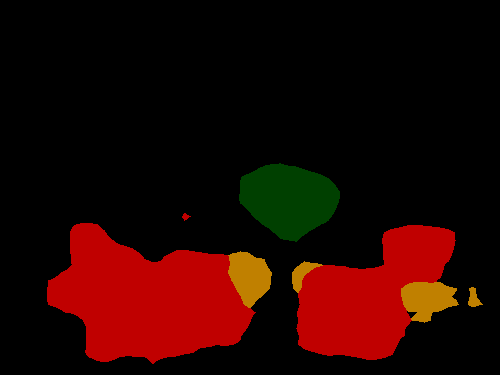} \label{fig:seg_cnn}
  }
  \subfigure{%
    \includegraphics[width=.23\columnwidth]{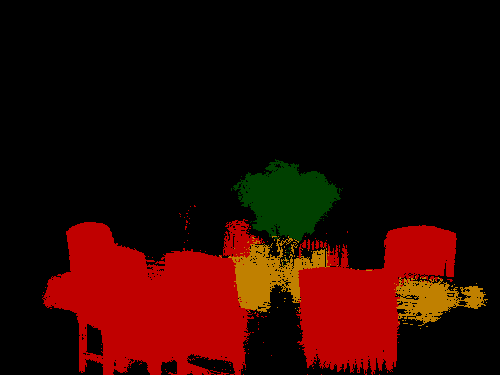} \label{fig:seg_mf2steps}
  }\\[-2ex]
  \setcounter{subfigure}{0}
  \subfigure[Input]{%
    \includegraphics[width=.23\columnwidth]{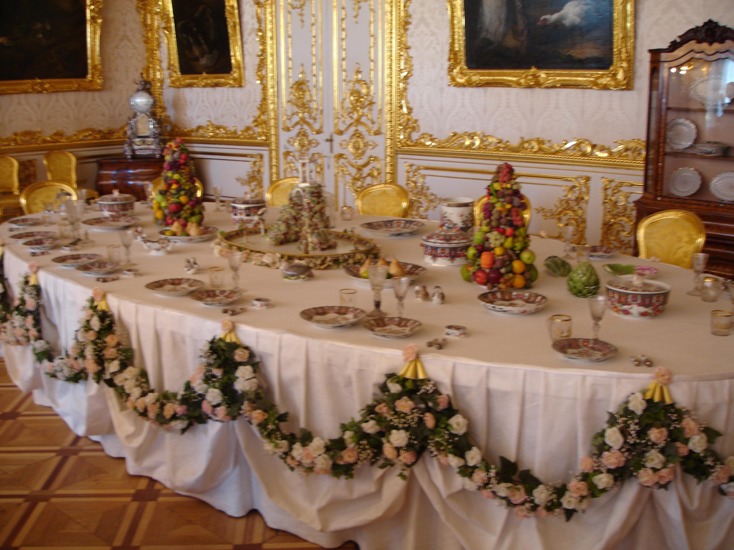} \label{fig:minc_given}
  }
  \subfigure[Ground Truth]{%
    \includegraphics[width=.23\columnwidth]{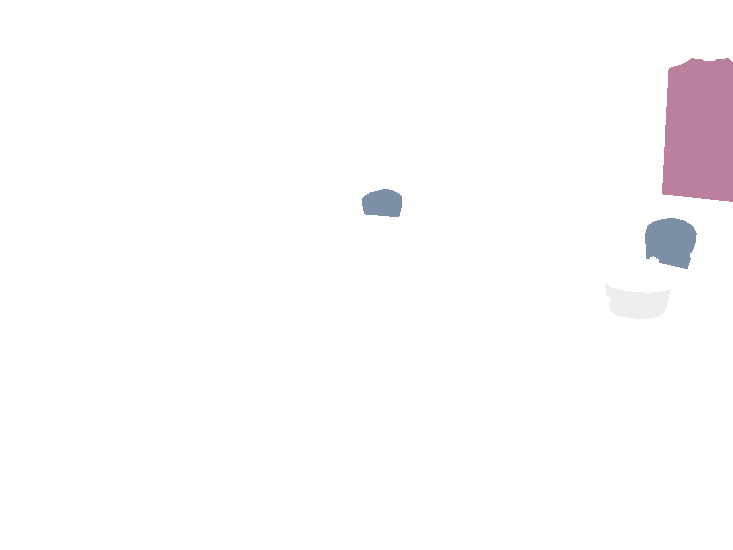} \label{fig:minc_gt}
  }
  \subfigure[CNN]{%
    \includegraphics[width=.23\columnwidth]{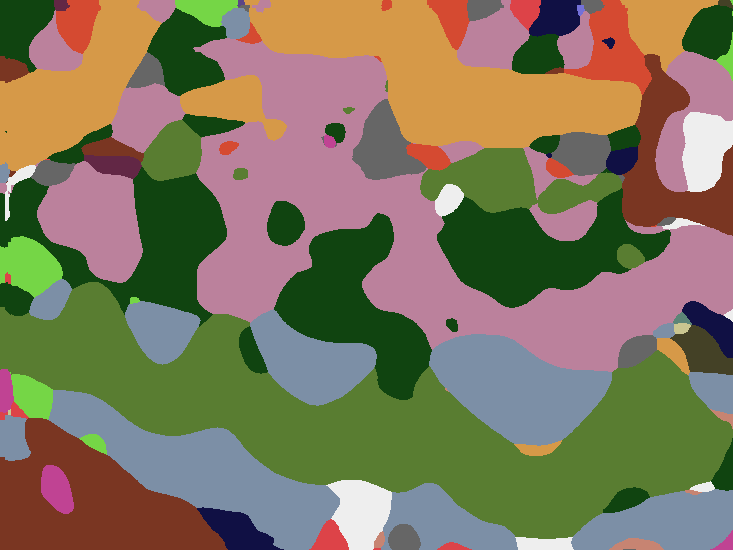} \label{fig:minc_cnn}
  }
  \subfigure[+\textit{loose}MF]{%
    \includegraphics[width=.23\columnwidth]{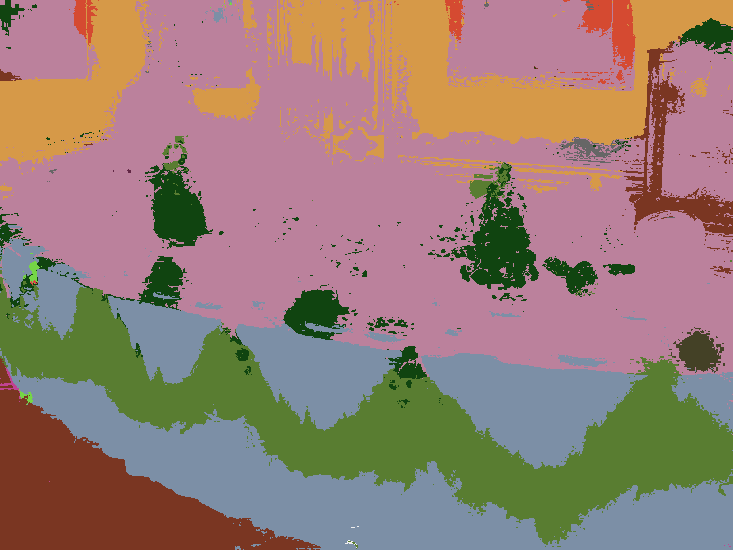} \label{fig:minc_mf2steps}
  }
  \mycaption{Segmentation results}{An example result for semantic (top) and material (bottom) segmentation.
  (c) depicts the unary results before application of MF, (d) after two steps of \textit{loose}-MF with
  a learned CRF. More examples with comparisons to Gaussian pairwise potentials are in Sec.~\ref{sec:app_semantic_segmentation} and Sec.~\ref{sec:app_material_segmentation}.}
    \label{fig:seg_visuals}
    \vspace{-0.3cm}
\end{figure}

\vspace{-0.3cm}
\subsubsection{Semantic Segmentation}
Semantic segmentation is the task of assigning semantic label (\eg, car, person
etc.) to every pixel. We choose the DeepLab network~\cite{chen2014semantic}, a
variant of the VGGnet~\cite{simonyan2014very} for obtaining unaries.
The DeepLab architecture runs a CNN
model on the input image to obtain a result that is down-sampled by a factor of
8. The result is then bilinear interpolated to the desired resolution and serves
as unaries $\psi_u(x_i)$ in a dense CRF. We use the same Pott's label
compatibility function $\mu$, and also use two kernels $k^1(\f_i,\f_j) +
k^2(p_i,p_j)$ with the same features $\f_i=(x_i,y_i,r_i,g_i)^{\top}$ and $\mathbf{p}_i=(x_i,y_i)^{\top}$
as in~\cite{chen2014semantic}. Thus, the two filters operate in parallel on
color \& position, and spatial domain respectively. We also initialize the
mean-field update equations with the CNN unaries. The only change in the model
is the type of the pairwise potential function from Gauss to a generalized form.

We evaluate the result after 1 step and 2 steps of mean-field inference and
compare the Gaussian filter versus the learned version (cf.
Tab.~\ref{tbl:seg_results}). First, as in~\cite{chen2014semantic} we observe
that one step of mean field improves the performance by 2.48\% in Intersection
over Union (IoU) score. However, a learned potential increases the score by
2.93\%. The same behaviour is observed for 2 steps: the learned result again
adds on top of the raised Gaussian mean field performance. Further, we tested a
variant of the mean-field model that learns a separate kernel for the first and
second step~\cite{li2014mean}. This ``loose'' mean-field model leads to further
improvement of the performance. It is not obvious how to take advantage
of a loose model in the case of Gaussian potentials.

\begin{table}[t]
\setlength{\tabcolsep}{2pt}
  \scriptsize
  \centering
    \begin{tabular}{l c c c c}
      \toprule
&  & + \textbf{MF-1step} & + \textbf{MF-2 step} & + \textbf{\textit{loose} MF-2 step} \\ [0.1cm]
      \midrule
      \multicolumn{4}{l}{Semantic segmentation (IoU) - CNN~\cite{chen2014semantic}: 72.08 / 66.95} & \\
      & Gauss CRF & +2.48 & +3.38  & +3.38 / +3.00   \\
      & Learned CRF & +2.93 & +3.71  &  \textbf{+3.85 / +3.37} \\
     \midrule
     \multicolumn{4}{l}{Material segmentation (Pixel Accuracy) - CNN~\cite{bell2015minc}: 67.21 / 69.23 } & \\
      & Gauss CRF & +7.91 / +6.28 & +9.68 / +7.35  &  +9.68 / +7.35  \\
      & Learned CRF & +9.48 / +6.23 & +11.89 / +6.93  &  \textbf{+11.91 / +6.93} \\
      \\
    \end{tabular}
    \vspace{-0.2cm}
    \mycaption{Improved mean-field inference with learned potentials} {(top) Average IoU score on Pascal VOC12
    validation/test data~\cite{voc2012segmentation} for semantic segmentation; (bottom) Accuracy for
    all pixels / averaged over classes on the MINC test data~\cite{bell2015minc} for material segmentation.}
  \label{tbl:seg_results}
  \vspace{-0.3cm}
\end{table}

\vspace{-0.2cm}
\subsubsection{Material Segmentation}

We adopt the method and dataset from~\cite{bell2015minc} for material segmentation task.
Their approach
proposes the same architecture as in the previous section; a CNN to predict the material labels (\eg, wool, glass, sky, etc.)
followed by a densely connected CRF using Gaussian potentials and mean-field inference.
We re-use their pre-trained CNN that is publicly available and further choose the
CRF parameters and Lab color/position features as in~\cite{bell2015minc}. We compare the Gaussian pairwise
potentials versus learned potentials with the same training procedure as for semantic segmentation.
Results for pixel accuracy and class-averaged pixel accuracy are shown in Table~\ref{tbl:seg_results}.
Following the CRF validation in~\cite{bell2015minc}, we ignored `other' label for both the training and
evaluation. For this dataset, the availability of training data is small, since there are 928 images that
are only sparsely annotated with segments. While that is enough to cross-validate parameters
for Gaussian CRFs, we would expect the general bilateral convolution to benefit from
more training data. Especially for class-averaged results we anticipate a higher increase in performance, which we
believe is hindered through size of the training set. Visual results with
annotations are shown in Fig.~\ref{fig:seg_visuals} and more are included in
Sec.~\ref{sec:app_material_segmentation}.

\begin{figure}[t!]
  \centering
  \subfigure[Sample tile images]{%
    \includegraphics[width=\columnwidth]{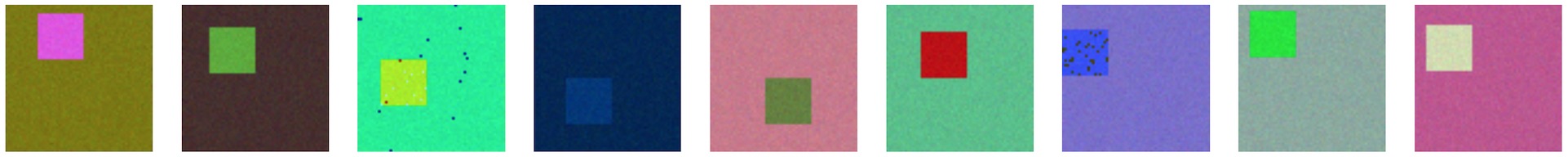}\label{fig:slate_sample}
  }\\[-1ex]
  \subfigure[NN architecture]{%
    \includegraphics[width=.48\columnwidth]{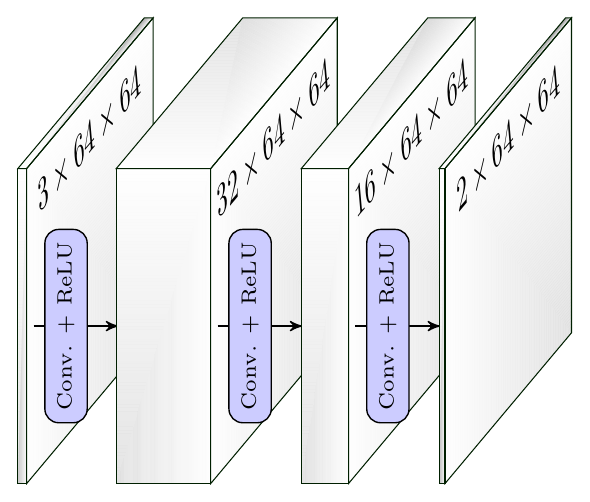}\label{fig:slate_network}
  }
  \subfigure[IoU versus Epochs]{%
    \includegraphics[width=.48\columnwidth]{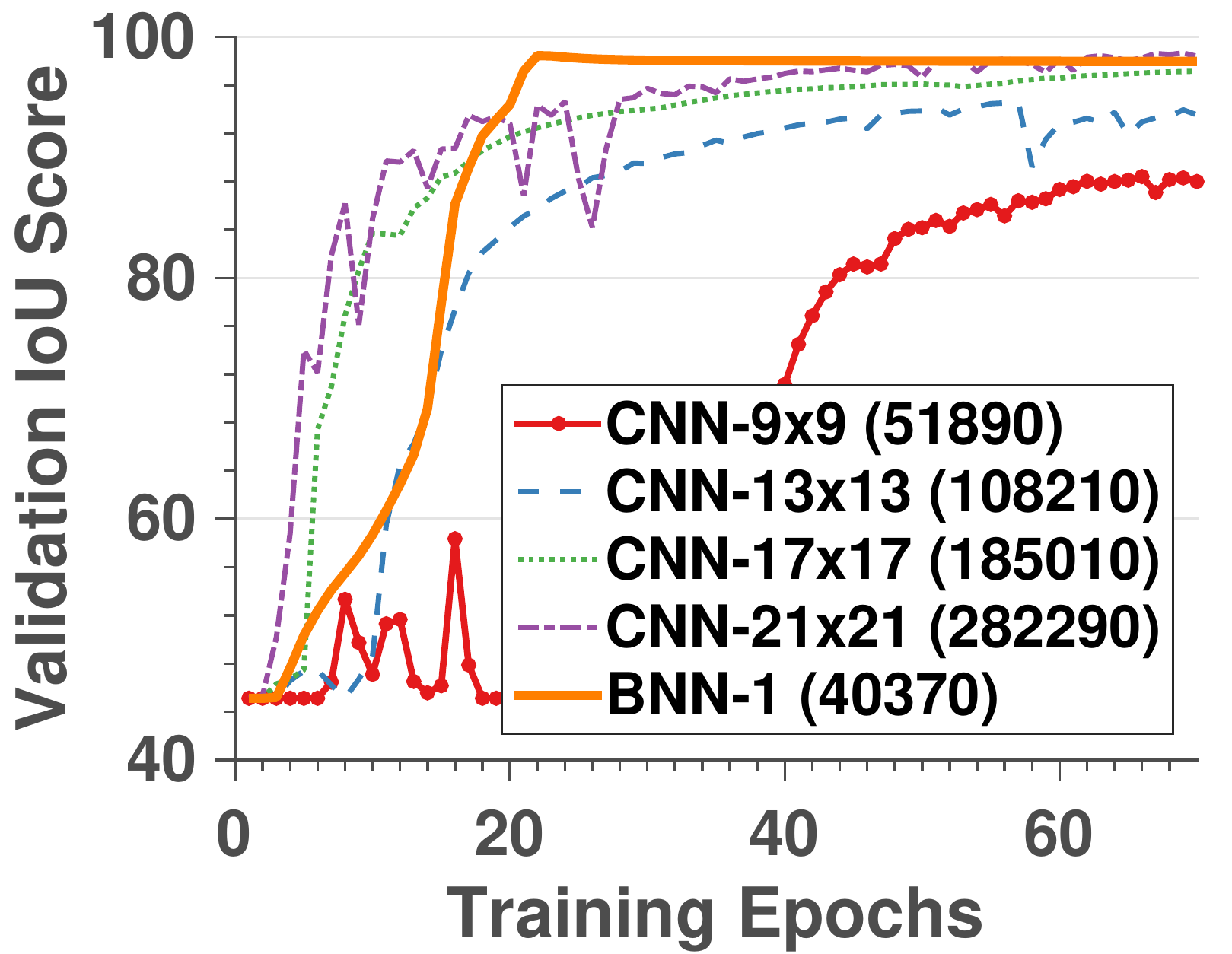}\label{fig:slate_plots}
  }
  \mycaption{Segmenting Tiles}{(a) Example tile input images; (b) the 3-layer NN architecture used
  in experiments. ``Conv'' stands for spatial convolutions, resp. bilateral convolutions;
  (c) Training progress in terms of Validation IoU versus training epochs.}
  \label{fig:slate_experiment}
  \vspace{-0.3cm}
\end{figure}

\section{Bilateral Neural Networks}\label{sec:bnn}

Probably the most promising opportunity for the generalized bilateral filter is its
use in Convolutional Neural Networks. Since we are not restricted to the Gaussian case anymore,
we can run several filters sequentially in the same way as filters are ordered in
layers in typical spatial CNN architectures. Having the gradients available allows for end-to-end training
with backpropagation, without the need for any change in CNN training protocols. We refer to the
layers of bilateral filters as ``bilateral convolution layers'' (BCL). As discussed in the introduction,
these can be understood as either linear filters in a high dimensional space or a filter with an
image adaptive receptive field. In the remainder we will refer to CNNs that include at least one bilateral
convolutional layer as a bilateral neural network (BNN).

What are the possibilities of a BCL compared to a standard spatial layer? First, we can define a
feature space $\f_i\in\mathbb{R}^d$ to define proximity between elements to perform
the convolution. This can include color or intensity as in the previous example.
Second, since the permutohedral lattice convolution takes advantage of hashing
it is advantageous for data in a sparse format. We performed a runtime comparison between
our implementation of a BCL\footnote{We have identified
several possible speed-ups that are not taken advantage of yet.}
and the caffe~\cite{jia2014caffe} implementation of a $d$-dimensional convolution. For 2D positional features (first row), the
standard layer is faster since the permutohedral algorithm comes with an overhead. For higher
dimensions $d>2$, the runtime depends on the sparsity; but ignoring the sparsity is quickly
leading to intractable runtimes. The permutohedral lattice convolution was designed especially for this purpose,
although with Gaussian filtering in mind.

\setlength{\tabcolsep}{4pt}
\begin{table}[h]
  \scriptsize
  \centering
    \begin{tabular}{l c c}
      \toprule
\textbf{Dim.-Features} & \textbf{d-dim caffe} & \textbf{BCL} \\ [0.1cm]
      \midrule
      2D-$(x, y)$ & \textbf{3.3 $\pm$ 0.3 / 0.5$ \pm$ 0.1} & 4.8 $\pm$ 0.5 / 2.8 $\pm$ 0.4  \\
      3D-$(r,g,b)$ & 364.5 $\pm$ 43.2 / 12.1 $\pm$ 0.4 & \textbf{5.1 $\pm$ 0.7 / 3.2 $\pm$ 0.4}  \\
      4D-$(x,r,g,b)$ & 30741.8 $\pm$ 9170.9 / 1446.2 $\pm$ 304.7 & \textbf{6.2 $\pm$ 0.7 / 3.8 $\pm$ 0.5}  \\
      5D-$(x,y,r,g,b)$ & out of memory & \textbf{7.6 $\pm$ 0.4 / 4.5 $\pm$ 0.4}  \\
      \\
    \end{tabular}
    \vspace{-0.1cm}
    \mycaption{Runtime comparison: BCL vs. spatial convolution} {Average CPU/GPU runtime (in ms)
    of 50 1-neighborhood filters averaged over 1000 images from Pascal VOC. All scaled
    features $(x,y,r,g,b)\in[0,50)$. BCL includes splatting and splicing operations which in layered networks
    can be re-used.}
\label{tbl:time}
  \vspace{-0.2cm}
\end{table}

Next we illustrate two use cases of BNNs and compare against spatial CNNs. Section~\ref{sec:app_mnist}
contains further explanatory experiments with examples on MNIST digit recognition.

\subsection{An Illustrative Example: Segmenting Tiles}

In order to highlight the model possibilities of using higher dimensional sparse feature
spaces for convolutions through BCLs, we constructed the following illustrative problem. A randomly colored
foreground tile with size $20\times 20$ is placed on a random colored background
of size $64 \times 64$ with additional Gaussian noise with std. dev. $0.02$ added with color values
normalized to $[0,1]$. The task is to segment out the smaller tile. Some example images
are shown in Fig.~\ref{fig:slate_sample}. A pixel classifier can not distinguish foreground
from background since the color is random. We use a three layer CNN with
convolutions interleaved with ReLU functions. The schematic of the architecture is shown
in Fig~\ref{fig:slate_network} ($32,16,2$ filters). We train standard CNNs, with filters of size $n \times n, n \in \{9, 13, 17, 21\}$.
We create 10k training, 1k validation and 1k test images and, use the
validation set to choose learning rates. In
Fig.~\ref{fig:slate_plots} we plot the validation IoU as a function of training epochs. All CNNs (and BNNs)
converge to almost perfect test predictions.

Now, we replace all spatial convolutions with bilateral convolutions for a full BNN.
The features are $\f_i=(x_i,y_i,r_i,g_i,b_i)^{\top}$ and the filter has a neighborhood of $1$.
The total number of
parameters in this network is around $40k$ compared to $52k$ for $9\times 9$ up to $282k$ for a $21\times 21$
CNN. With the same training protocol
and optimizer, the convergence rate of BNN is much faster. In this example
as in semantic segmentation discussed in the last section, color is a discriminative feature for
the label. The bilateral convolutions ``see'' the color difference, the points are already
pre-grouped in the permutohedral lattice and the task remains to assign a label to the two groups.

\subsection{Character Recognition}
The results for tile, semantic, and material segmentations when using general bilateral
filters mainly improved because the feature space was used to encode useful prior information about the
problem (similar RGB of close-by pixels have the same label). Such prior knowledge is often available when
structured predictions are to be made but the input signal may also be in a sparse format to begin with.
Let us consider handwritten character recognition, one of the prime cases for CNN use.

\begin{figure}[t]
  \centering
  \subfigure[Sample Assamese character images (9 classes, 2 samples each)]{%
    \includegraphics[width=\columnwidth]{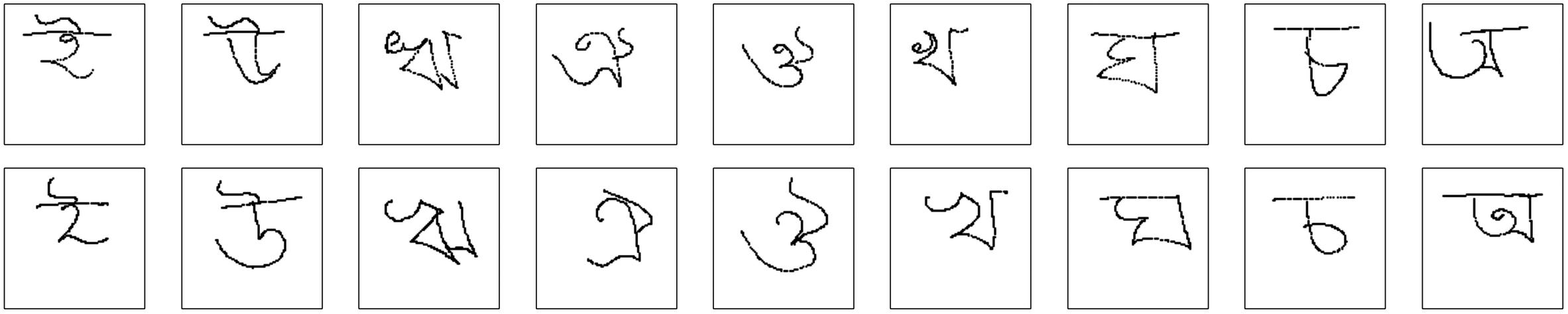}\label{fig:assamese_sample}
  }\\[-1ex]
  \subfigure[LeNet training]{%
    \includegraphics[width=.48\columnwidth]{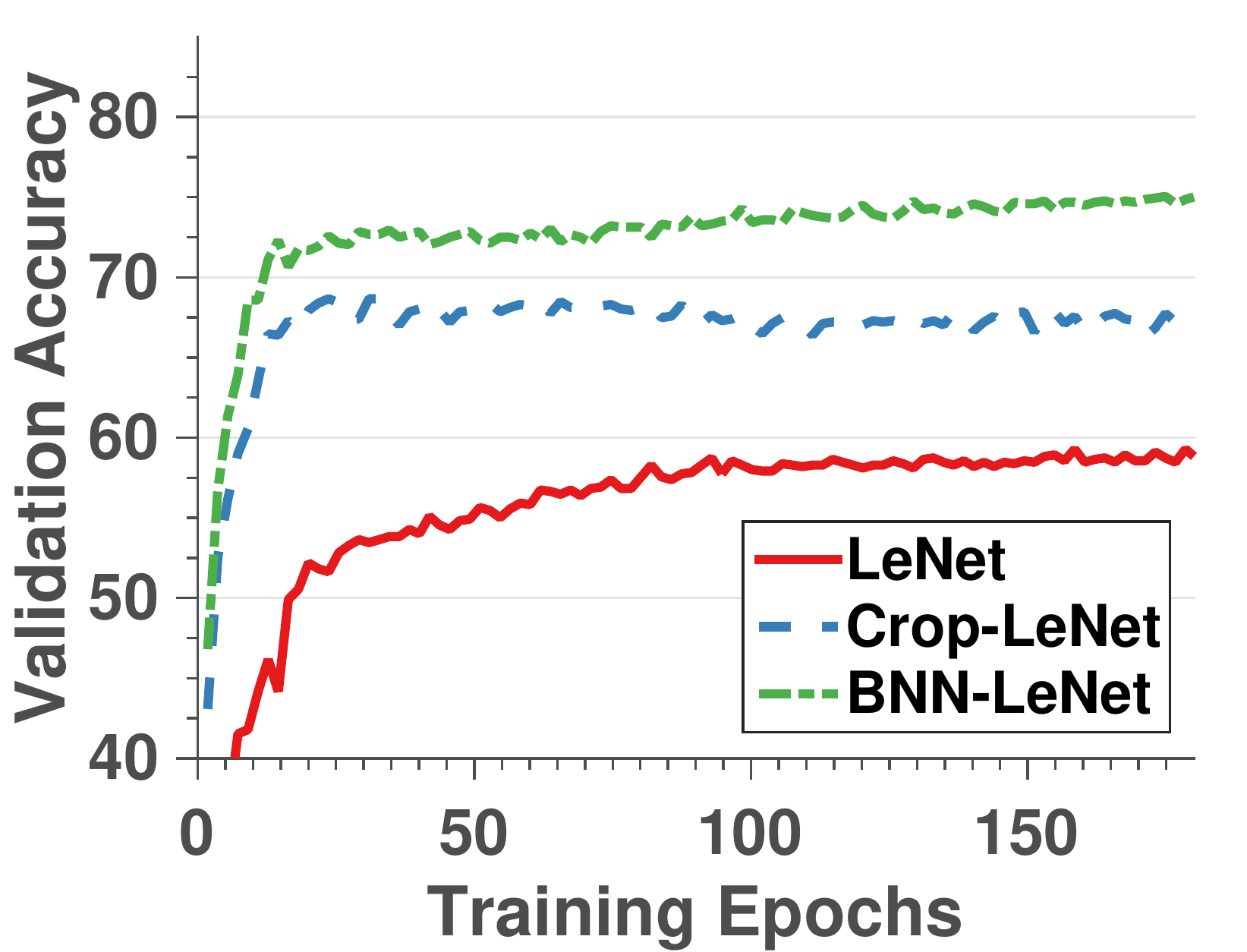}\label{fig:assamese_lenet}
  }
  \subfigure[DeepCNet training]{%
    \includegraphics[width=.48\columnwidth]{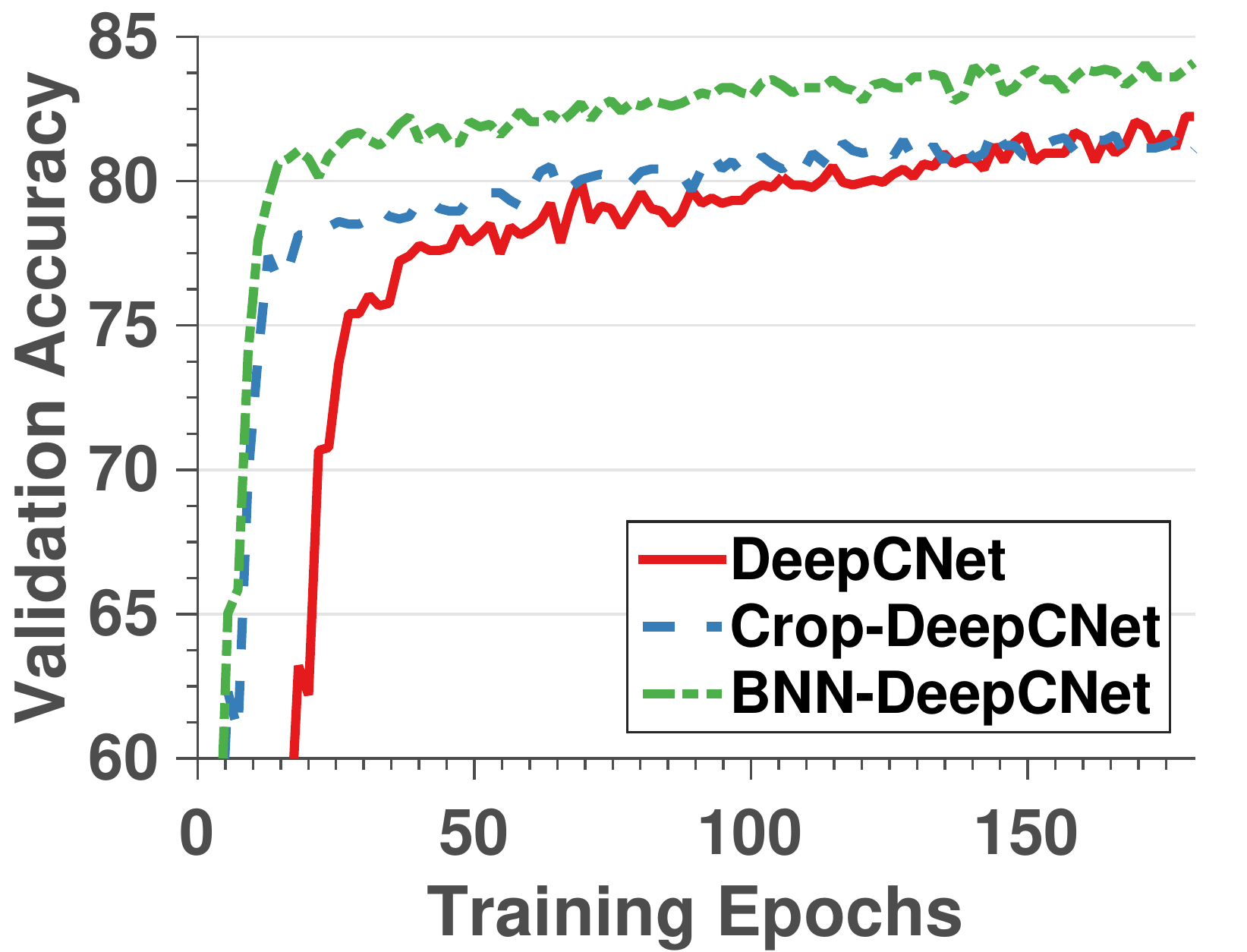}\label{fig:assamese_deepcnet}
  }
  \mycaption{Character recognition}{ (a) Sample Assamese character images~\cite{bache2013uci};
  and training progression of various models with (b) LeNet and (c) DeepCNet base networks.}
\label{fig:assamese_experiment}
  \vspace{-0.3cm}
\end{figure}

The Assamese character dataset~\cite{bache2013uci} contains 183 different Indo-Aryan
symbols with 45 writing samples per class. Some sample character images are shown in
Fig.~\ref{fig:assamese_sample}. This dataset has been collected on a tablet PC using a pen
input device and has been pre-processed to binary images of size $96 \times 96$.
Only about $3\%$ of the pixels contain a pen stroke, which we will denote by $v_i=1$.
%

A CNN is a natural choice to approach this classification task. We experiment with two CNN
architectures for this experiment that have been used for this task, LeNet-7 from~\cite{lecun1998gradient}
and DeepCNet~\cite{ciresan2012multi, graham2014spatially}. The LeNet is a shallower network with
bigger filter sizes whereas DeepCNet is deeper with smaller convolutions.
Both networks are fully specified in
Sec.~\ref{sec:app_character}. In order to simplify the task for the networks we cropped the
characters by placing a tight bounding box around them and providing the bounding boxes as input to the
networks. We will call these networks
``Crop-''LeNet, resp. DeepCNet. For training, we randomly divided the data into 30 writers for training,
6 for validation and the remaining 9 for
test. Fig.\ref{fig:assamese_lenet} and Fig.~\ref{fig:assamese_deepcnet} show the training progress for various
LeNet and DeepCNet models respectively. DeepCNet is a better choice for this problem and for both cases
pre-processing the data by cropping improves convergence.

The input is spatially sparse and the BCL provides a natural way to take advantage of this. For
both networks we create a BNN variant (BNN-LeNet and BNN-DeepCNet) by replacing the
first layer with a bilateral convolutions using the features $\f_i = (x_i, y_i)^{\top}$ and we \emph{only}
consider the foreground points $v_i=1$. The values $(x_i,y_i)$ denote the position of the pixel with respect
to the top-left corner of the bounding box around the character. In effect the lattice is very
sparse which reduces runtime because the convolutions are only performed on $3\%$ of the
points that are actually observed. A bilateral filter has $7$ parameters compared to a receptive field of $3\times 3$ for the first
DeepCNet layer and $5\times 5$ for the first LeNet layer. Thus, a BCL with the same number of filters has fewer parameters.
The result of the BCL convolution is then splatted at all points $(x_i,y_i)$ and
passed on to the remaining spatial layers. The convergence behaviour is shown in Fig.\ref{fig:assamese_experiment}
and again we find faster convergence and also better validation accuracy. The empirical results of this
experiment for all tested architectures are summarized in Table~\ref{tbl:assamese}, with BNN variants
clearly outperforming their spatial counterparts.

The absolute results can be vastly improved by making use of virtual examples, \eg,
by affine transformations~\cite{graham2014spatially}. The purpose of these experiments is to compare the networks on equal grounds while we
believe that additional data will be beneficial for both networks. We have no reason to believe that a particular
network benefits more.

\setlength{\tabcolsep}{2pt}
\newcolumntype{L}[1]{>{\raggedright\let\newline\\\arraybackslash\hspace{0pt}}b{#1}}
\newcolumntype{C}[1]{>{\centering\let\newline\\\arraybackslash\hspace{0pt}}b{#1}}
\newcolumntype{R}[1]{>{\raggedleft\let\newline\\\arraybackslash\hspace{0pt}}b{#1}}
\begin{table}[t]
  \scriptsize
  \centering
    \begin{tabular}{C{1.05cm} C{1.05cm} C{1.05cm} C{1.05cm} | C{1.05cm} C{1.05cm} C{1.05cm}}
      \toprule
 & \textbf{LeNet} & \textbf{Crop-LeNet} & \textbf{BNN-LeNet}  & \textbf{DeepCNet} & \textbf{Crop-DeepCNet} & \textbf{BNN-DeepCNet} \\ [0.2cm]
      \midrule
      Validation & 59.29 & 68.67 & \textbf{75.05}  & 82.24 & 81.88 & \textbf{84.15} \\
      Test & 55.74 & 69.10 & \textbf{74.98}  & 79.78 & 80.02 & \textbf{84.21}  \\
      \\
    \end{tabular}
    \vspace{-0.2cm}
    \mycaption{Results on Assamese character images} {Total recognition accuracy for the
    different models.}
  \label{tbl:assamese}
    \vspace{-0.3cm}
\end{table}
\vspace{-0.1cm}

\section{Conclusion}

We proposed to learn bilateral filters from data. In
hindsight, it may appear obvious that this leads to performance improvements compared
to a particular chosen parametric form, \eg, the Gaussian. To understand the algorithms that
facilitate fast approximative computation of Eq.~\eqref{eq:bilateral} as a parameterized
implementation of a bilateral filter that has free parameters is the key insight and
enables gradient descent based learning. We relaxed the non-separability in the
algorithm from~\cite{adams2010fast} to allow for more general filter functions.
There is a wide range of possible applications for learned bilateral filters~\cite{paris2009bilateral}. In this paper
we discussed some examples that generalize previous work. These include joint bilateral
upsampling, inference in dense CRFs, and the use of bilateral
convolutions in neural networks. On several and diverse experiments, we observed empirical improvements when learning
bilateral filters.

The bilateral convolutional layer allows for filters whose
receptive field change given the input image. The feature space view provides a
canonical way to encode similarity between any kind of objects, not only pixels, but \eg, bounding boxes,
segmentations, surfaces, etc. The proposed filtering operation is then a natural candidate to
define a filter convolutions on these objects, it takes advantage of sparsity and scales to higher dimensions.
Therefore, we believe that this view will be useful for several problems where CNNs
can be applied. An open research problem is whether the sprase higher dimensional
structure also allows for efficient or compact representations for intermediate layers inside
CNN architectures.

\vspace{-0.3cm}
{\small \paragraph{Acknowledgements} We thank Jonas Wulff, Laura Sevilla, Abhilash Srikantha,
Christoph Lassner, Andreas Lehrmann, Thomas Nestmeyer, Andreas Geiger,
Fatma G{\"u}ney and Gerard Pons-Moll for their valuable feedback.
}

\vspace{-0.1cm}
{\small
\bibliographystyle{ieee}
\bibliography{references}

\begin{thebibliography}{10}\itemsep=-1pt

\bibitem{google_images}
{Google Images}.
\newblock \url{https://images.google.com/}.
\newblock [Online; accessed 1-March-2015].

\bibitem{adams2010fast}
A.~Adams, J.~Baek, and M.~A. Davis.
\newblock Fast high-dimensional filtering using the permutohedral lattice.
\newblock In {\em Computer Graphics Forum}, volume~29, pages 753--762. Wiley
  Online Library, 2010.

\bibitem{adams2009gaussian}
A.~Adams, N.~Gelfand, J.~Dolson, and M.~Levoy.
\newblock Gaussian kd-trees for fast high-dimensional filtering.
\newblock In {\em ACM Transactions on Graphics (TOG)}, volume~28, page~21. ACM,
  2009.

\bibitem{arbelaezi2011bsds500}
P.~Arbel{\'a}ez, M.~Maire, C.~Fowlkes, and J.~Malik.
\newblock Contour detection and hierarchical image segmentation.
\newblock {\em IEEE Trans. Pattern Anal. Mach. Intell.}, 33(5):898--916, May
  2011.

\bibitem{aurich1995non}
V.~Aurich and J.~Weule.
\newblock Non-linear gaussian filters performing edge preserving diffusion.
\newblock In {\em DAGM}, pages 538--545. Springer, 1995.

\bibitem{awate2005higher}
S.~P. Awate and R.~T. Whitaker.
\newblock Higher-order image statistics for unsupervised,
  information-theoretic, adaptive, image filtering.
\newblock In {\em Computer Vision and Pattern Recognition, 2005. IEEE Computer
  Society Conference on}, volume~2, pages 44--51.

\bibitem{bache2013uci}
K.~Bache and M.~Lichman.
\newblock {UCI} machine learning repository, 2013.

\bibitem{barron2015defocus}
J.~T. Barron, A.~Adams, Y.~Shih, and C.~Hern\'andez.
\newblock Fast bilateral-space stereo for synthetic defocus.
\newblock {\em CVPR}, 2015.

\bibitem{barron2015bilateral}
J.~T. {Barron} and B.~{Poole}.
\newblock {The Fast Bilateral Solver}.
\newblock {\em ArXiv e-prints}, Nov. 2015.

\bibitem{bell2014intrinsic}
S.~Bell, K.~Bala, and N.~Snavely.
\newblock Intrinsic images in the wild.
\newblock {\em ACM Transactions on Graphics (TOG)}, 33(4):159, 2014.

\bibitem{bell2015minc}
S.~Bell, P.~Upchurch, N.~Snavely, and K.~Bala.
\newblock Material recognition in the wild with the materials in context
  database.
\newblock {\em Computer Vision and Pattern Recognition (CVPR)}, 2015.

\bibitem{bruna2013spectral}
J.~Bruna, W.~Zaremba, A.~Szlam, and Y.~LeCun.
\newblock Spectral networks and locally connected networks on graphs.
\newblock {\em arXiv preprint arXiv:1312.6203}, 2013.

\bibitem{buades2005non}
A.~Buades, B.~Coll, and J.-M. Morel.
\newblock A non-local algorithm for image denoising.
\newblock In {\em Computer Vision and Pattern Recognition, 2005. IEEE Computer
  Society Conference on}, volume~2, pages 60--65.

\bibitem{burger12cvpr}
H.~C. Burger, C.~J. Schuler, and S.~Harmeling.
\newblock Image denoising: Can plain neural networks compete with {BM3D}?
\newblock In {\em Computer Vision and Pattern Recognition (CVPR)}, 2012.

\bibitem{campbell2013fully}
N.~D. Campbell, K.~Subr, and J.~Kautz.
\newblock Fully-connected {CRF}s with non-parametric pairwise potential.
\newblock In {\em Computer Vision and Pattern Recognition (CVPR), 2013 IEEE
  Conference on}, pages 1658--1665.

\bibitem{chen2014semantic}
L.-C. Chen, G.~Papandreou, I.~Kokkinos, K.~Murphy, and A.~L. Yuille.
\newblock Semantic image segmentation with deep convolutional nets and fully
  connected {CRF}s.
\newblock {\em arXiv preprint arXiv:1412.7062}, 2014.

\bibitem{ciresan2012multi}
D.~Ciresan, U.~Meier, and J.~Schmidhuber.
\newblock Multi-column deep neural networks for image classification.
\newblock In {\em Computer Vision and Pattern Recognition (CVPR), 2012 IEEE
  Conference on}, pages 3642--3649.

\bibitem{domke2013learning}
J.~Domke.
\newblock Learning graphical model parameters with approximate marginal
  inference.
\newblock {\em Pattern Analysis and Machine Intelligence, IEEE Transactions
  on}, 35(10):2454--2467, 2013.

\bibitem{eigen2014depth}
D.~Eigen, C.~Puhrsch, and R.~Fergus.
\newblock Depth map prediction from a single image using a multi-scale deep
  network.
\newblock In {\em Advances in Neural Information Processing Systems}, pages
  2366--2374, 2014.

\bibitem{voc2012segmentation}
M.~Everingham, L.~V. Gool, C.~Williams, J.~Winn, and A.~Zisserman.
\newblock The {PASCAL} {VOC2012} challenge results.
\newblock 2012.

\bibitem{ferstl2015b}
D.~Ferstl, M.~Ruether, and H.~Bischof.
\newblock Variational depth superresolution using example-based edge
  representations.
\newblock In {\em Proceedings International Conference on Computer Vision
  (ICCV), IEEE}, December 2015.

\bibitem{gastal2012adaptive}
E.~S. Gastal and M.~M. Oliveira.
\newblock Adaptive manifolds for real-time high-dimensional filtering.
\newblock {\em ACM Transactions on Graphics (TOG)}, 31(4):33, 2012.

\bibitem{graham2014spatially}
B.~Graham.
\newblock Spatially-sparse convolutional neural networks.
\newblock {\em arXiv preprint arXiv:1409.6070}, 2014.

\bibitem{graham2015sparse}
B.~Graham.
\newblock Sparse {3D} convolutional neural networks.
\newblock {\em arXiv preprint arXiv:1505.02890}, 2015.

\bibitem{hariharan2011moredata}
B.~Hariharan, P.~Arbel{\'a}ez, L.~Bourdev, S.~Maji, and J.~Malik.
\newblock Semantic contours from inverse detectors.
\newblock In {\em International Conference on Computer Vision (ICCV)}, 2011.

\bibitem{he2013guided}
K.~He, J.~Sun, and X.~Tang.
\newblock Guided image filtering.
\newblock {\em Pattern Analysis and Machine Intelligence, IEEE Transactions
  on}, 35(6):1397--1409, 2013.

\bibitem{hu2007trained}
H.~Hu and G.~de~Haan.
\newblock Trained bilateral filters and applications to coding artifacts
  reduction.
\newblock In {\em Image Processing, 2007. IEEE International Conference on},
  volume~1, pages I--325.

\bibitem{ionescu2015matrix}
C.~Ionescu, O.~Vantzos, and C.~Sminchisescu.
\newblock Matrix backpropagation for deep networks with structured layers.
\newblock {\em ICCV}, 2015.

\bibitem{jaderberg2015spatial}
M.~Jaderberg, K.~Simonyan, A.~Zisserman, and K.~Kavukcuoglu.
\newblock Spatial transformer networks.
\newblock {\em arXiv preprint arXiv:1506.02025}, 2015.

\bibitem{jia2014caffe}
Y.~Jia, E.~Shelhamer, J.~Donahue, S.~Karayev, J.~Long, R.~Girshick,
  S.~Guadarrama, and T.~Darrell.
\newblock Caffe: Convolutional architecture for fast feature embedding.
\newblock In {\em Proceedings of the ACM International Conference on
  Multimedia}, pages 675--678. ACM, 2014.

\bibitem{kiefel2014human}
M.~Kiefel and P.~V. Gehler.
\newblock Human pose estimation with fields of parts.
\newblock In {\em Computer Vision--ECCV 2014}, pages 331--346. Springer, 2014.

\bibitem{kopf2007joint}
J.~Kopf, M.~F. Cohen, D.~Lischinski, and M.~Uyttendaele.
\newblock Joint bilateral upsampling.
\newblock {\em ACM Transactions on Graphics (TOG)}, 26(3):96, 2007.

\bibitem{krahenbuhl2012efficient}
P.~Kr{\"a}henb{\"u}hl and V.~Koltun.
\newblock Efficient inference in fully connected {CRF}s with {G}aussian edge
  potentials.
\newblock {\em arXiv preprint arXiv:1210.5644}, 2012.

\bibitem{krahenbuhl2013parameter}
P.~Kr{\"a}henb{\"u}hl and V.~Koltun.
\newblock Parameter learning and convergent inference for dense random fields.
\newblock In {\em Proceedings of the 30th International Conference on Machine
  Learning (ICML-13)}, pages 513--521, 2013.

\bibitem{lecun1998gradient}
Y.~LeCun, L.~Bottou, Y.~Bengio, and P.~Haffner.
\newblock Gradient-based learning applied to document recognition.
\newblock {\em Proceedings of the IEEE}, 86(11):2278--2324, 1998.

\bibitem{lecun1998mnist}
Y.~LeCun, C.~Cortes, and C.~J. Burges.
\newblock The mnist database of handwritten digits, 1998.

\bibitem{li2014mean}
Y.~Li and R.~Zemel.
\newblock Mean-field networks.
\newblock {\em arXiv preprint arXiv:1410.5884}, 2014.

\bibitem{SMPL:2015}
M.~Loper, N.~Mahmood, J.~Romero, G.~Pons-Moll, and M.~J. Black.
\newblock {SMPL}: A skinned multi-person linear model.
\newblock {\em ACM Trans. Graphics (Proc. SIGGRAPH Asia)}, 34(6):248:1--248:16,
  Oct. 2015.

\bibitem{milanfar2011tour}
P.~Milanfar.
\newblock A tour of modern image filtering.
\newblock {\em IEEE Signal Processing Magazine}, 2, 2011.

\bibitem{paris2006fast}
S.~Paris and F.~Durand.
\newblock A fast approximation of the bilateral filter using a signal
  processing approach.
\newblock In {\em Computer Vision--ECCV 2006}, pages 568--580. Springer, 2006.

\bibitem{paris2009bilateral}
S.~Paris, P.~Kornprobst, J.~Tumblin, and F.~Durand.
\newblock {\em Bilateral filtering: Theory and applications}.
\newblock Now Publishers Inc, 2009.

\bibitem{rick2015propagated}
J.-H. Rick~Chang and Y.-C. Frank~Wang.
\newblock Propagated image filtering.
\newblock In {\em Proceedings of the IEEE Conference on Computer Vision and
  Pattern Recognition}, pages 10--18, 2015.

\bibitem{silberman2012indoor}
N.~Silberman, D.~Hoiem, P.~Kohli, and R.~Fergus.
\newblock Indoor segmentation and support inference from rgbd images.
\newblock In {\em Computer Vision--ECCV 2012}, pages 746--760. Springer, 2012.

\bibitem{simonyan2014very}
K.~Simonyan and A.~Zisserman.
\newblock Very deep convolutional networks for large-scale image recognition.
\newblock {\em arXiv preprint arXiv:1409.1556}, 2014.

\bibitem{smith97ijcv}
S.~M. Smith and J.~M. Brady.
\newblock {SUSAN} -- a new approach to low level image processing.
\newblock {\em Int. J. Comput. Vision}, 23(1):45--78, May 1997.

\bibitem{sun2013fully}
D.~Sun, J.~Wulff, E.~B. Sudderth, H.~Pfister, and M.~J. Black.
\newblock A fully-connected layered model of foreground and background flow.
\newblock In {\em Computer Vision and Pattern Recognition (CVPR), 2013 IEEE
  Conference on}, pages 2451--2458.

\bibitem{takeda2007kernel}
H.~Takeda, S.~Farsiu, and P.~Milanfar.
\newblock Kernel regression for image processing and reconstruction.
\newblock {\em Image Processing, IEEE Transactions on}, 16(2):349--366, 2007.

\bibitem{isomap_code}
J.~B. Tenenbaum.
\newblock {A Global Geometric Framework for Nonlinear Dimensionality
  Reduction}.
\newblock \url{http://isomap.stanford.edu/}, 2000.
\newblock [Online; accessed 12-October-2015].

\bibitem{tenenbaum2000global}
J.~B. Tenenbaum, V.~De~Silva, and J.~C. Langford.
\newblock A global geometric framework for nonlinear dimensionality reduction.
\newblock {\em Science}, 290(5500):2319--2323, 2000.

\bibitem{tomasi1998bilateral}
C.~Tomasi and R.~Manduchi.
\newblock Bilateral filtering for gray and color images.
\newblock In {\em Computer Vision, 1998. Sixth International Conference on},
  pages 839--846. IEEE, 1998.

\bibitem{vineet12emmcvpr}
V.~Vineet, G.~Sheasby, J.~Warrell, and P.~H. Torr.
\newblock Posefield: An efficient mean-field based method for joint estimation
  of human pose, segmentation and depth.
\newblock In {\em EMMCVPR}, 2013.

\bibitem{vineet12eccv}
V.~Vineet, J.~Warrell, and P.~H. Torr.
\newblock Filter-based mean-field inference for random fields with higher order
  terms and product label-spaces.
\newblock In {\em ECCV}, 2012.

\bibitem{yagou2002mesh}
H.~Yagou, Y.~Ohtake, and A.~Belyaev.
\newblock Mesh smoothing via mean and median filtering applied to face normals.
\newblock In {\em Geometric Modeling and Processing, 2002. Proceedings}, pages
  124--131. IEEE, 2002.

\bibitem{zeiler2010deconvolutional}
M.~D. Zeiler, D.~Krishnan, G.~W. Taylor, and R.~Fergus.
\newblock Deconvolutional networks.
\newblock In {\em Computer Vision and Pattern Recognition (CVPR), 2010 IEEE
  Conference on}, pages 2528--2535.

\bibitem{zheng2015conditional}
S.~Zheng, S.~Jayasumana, B.~Romera-Paredes, V.~Vineet, Z.~Su, D.~Du, C.~Huang,
  and P.~H. Torr.
\newblock Conditional random fields as recurrent neural networks.
\newblock {\em IEEE International Conference on Computer Vision}, 2015.

\end{thebibliography}
}

\appendix
\section*{Appendix}
The appendix contains a more detailed overview of the permutohedral
lattice convolution in Section~\ref{sec:permconv}, more experiments in
Section~\ref{sec:addexps} and additional results with experiment protocols for
the experiments presented before in Section~\ref{sec:addresults}.

\section{General Permutohedral Convolutions}
\label{sec:permconv}

A core technical contribution of this work is the generalization of the Gaussian permutohedral lattice
convolution proposed in~\cite{adams2010fast} to the full non-separable case with the
ability to perform backpropagation. Although, conceptually, there are minor
difference between non-Gaussian and general parameterized filters, there are non-trivial practical
differences in terms of the algorithmic implementation. The Gauss filters belong to
the separable class and can thus be decomposed into multiple
sequential one dimensional convolutions. We are interested in the general filter
convolutions, which can not be decomposed. Thus, performing a general permutohedral
convolution at a lattice point requires the computation of the inner product with the
neighboring elements in all the directions in the high-dimensional space.

Here, we give more details of the implementation differences of separable
and non-separable filters. In the following we will explain the scalar case first.
Recall, that the forward pass of general permutohedral convolution
involves 3 steps: \textit{splatting}, \textit{convolving} and \textit{slicing}.
We follow the same splatting and slicing strategies as in~\cite{adams2010fast}
since these operations do not depend on the filter kernel. The main difference
between our work and the existing implementation of~\cite{adams2010fast} is
the way that the convolution operation is executed. This proceeds by constructing
a \emph{blur neighbor} matrix $K$ that stores for every lattice point all
values of the lattice neighbors that are needed to compute the filter output.

\begin{figure}[t!]
  \centering
    \includegraphics[width=\columnwidth]{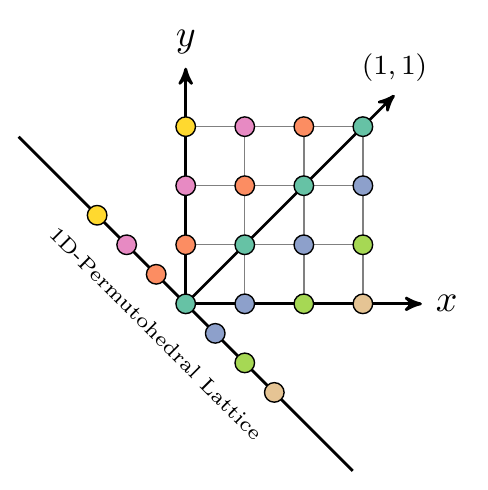}
  \mycaption{Illustration of 1D permutohedral lattice construction}
  {A $4\times 4$ $(x,y)$ grid lattice is projected onto the plane defined by the normal
  vector $(1,1)^{\top}$. This grid has $s+1=4$ and $d=2$ $(s+1)^{d}=4^2=16$ elements.
  In the projection, all points of the same color are projected onto the same points in the plane.
  The number of elements of the projected lattice is $t=(s+1)^d-s^d=4^2-3^2=7$, that is
  the $(4\times 4)$ grid minus the size of lattice that is $1$ smaller at each size, in this
  case a $(3\times 3)$ lattice (the upper right $(3\times 3)$ elements).
  }
\label{fig:latticeconstruction}
\end{figure}

The blur neighbor matrix is constructed by traversing through all the populated
lattice points and their neighboring elements.
This is done recursively to share computations. For any lattice point, the neighbors that are
$n$ hops away are the direct neighbors of the points that are $n-1$ hops away.
The size of a $d$ dimensional spatial filter with width $s+1$ is $(s+1)^{d}$ (\eg, a
$3\times 3$ filter, $s=2$ in $d=2$ has $3^2=9$ elements) and this size grows
exponentially in the number of dimensions $d$. The permutohedral lattice is constructed by
projecting a regular grid onto the plane spanned by the $d$ dimensional normal vector ${(1,\ldots,1)}^{\top}$. See
Fig.~\ref{fig:latticeconstruction} for an illustration of 1D lattice construction.
Many corners of a grid filter are projected onto the same point, in total $t = {(s+1)}^{d} -
s^{d}$ elements remain in the permutohedral filter with $s$ neighborhood in $d-1$ dimensions.
If the lattice has $m$ populated elements, the
matrix $K$ has size $t\times m$. Note that, since the input signal is typically
sparse, only a few lattice corners are being populated in the \textit{slicing} step.
We use a hash-table to keep track of these points and traverse only through
the populated lattice points for this neighborhood matrix construction.

Once the blur neighbor matrix $K$ is constructed, we can perform the convolution
by the matrix vector multiplication
\begin{equation}
l' = BK,
\label{eq:conv}
\end{equation}
where $B$ is the $1 \times t$ filter kernel (whose values we will learn) and $l'\in\mathbb{R}^{1\times m}$
is the result of the filtering at the $m$ lattice points. In practice, we found that the
matrix $K$ is sometimes too large to fit into GPU memory and we divided the matrix $K$
into smaller pieces to compute Eq.~\ref{eq:conv} sequentially.

In the general multi-dimensional case, the signal $l$ is of $c$ dimensions. Then
the kernel is of size $B\times t$ and $K$ stores the $c$ dimensional vectors
accordingly. When the input and output points are different, we slice only the
input points and splat only at the output points.

\section{Additional Experiments}
\label{sec:addexps}
In this section we discuss more use-cases for the learned bilateral filters, one
use-case of BNNs and two single filter applications for image and 3D mesh denoising.

\subsection{Recognition of subsampled MNIST}\label{sec:app_mnist}

One of the strengths of the proposed filter convolution is that it does not
require the input to lie on a regular grid. The only requirement is to define a distance
between features of the input signal.
We highlight this feature with the following experiment using the
classical MNIST ten class classification problem~\cite{lecun1998mnist}. We sample a
sparse set of $N$ points $(x,y)\in [0,1]\times [0,1]$
uniformly at random in the input image, use their interpolated values
as signal and the \emph{continuous} $(x,y)$ positions as features. This mimics
sub-sampling of a high-dimensional signal. To compare against a spatial convolution,
we interpolate the sparse set of values at the grid positions.

We take a reference implementation of LeNet~\cite{lecun1998gradient} that
is part of the Caffe project~\cite{jia2014caffe} and compare it
against the same architecture but replacing the first convolutional
layer with a bilateral convolution layer (BCL). The filter size
and numbers are adjusted to get a comparable number of parameters
($5\times 5$ for LeNet, $2$-neighborhood for BCL).

The results are shown in Table~\ref{tab:all-results}. We see that training
on the original MNIST data (column Original, LeNet vs. BNN) leads to a slight
decrease in performance of the BNN (99.03\%) compared to LeNet
(99.19\%). The BNN can be trained and evaluated on sparse
signals, and we resample the image as described above for $N=$ 100\%, 60\% and
20\% of the total number of pixels. The methods are also evaluated
on test images that are subsampled in the same way. Note that we can
train and test with different subsampling rates. We introduce an additional
bilinear interpolation layer for the LeNet architecture to train on the same
data. In essence, both models perform a spatial interpolation and thus we
expect them to yield a similar classification accuracy. Once the data is of
higher dimensions the permutohedral convolution will be faster due to hashing
the sparse input points, as well as less memory demanding in comparison to
naive application of a spatial convolution with interpolated values.

\begin{table}[t]
  \begin{center}
    \footnotesize
    \centering
    \begin{tabular}[t]{lllll}
      \toprule
              &     & \multicolumn{3}{c}{Test Subsampling} \\
       Method  & Original & 100\% & 60\% & 20\%\\
      \midrule
       LeNet &  \textbf{0.9919} & 0.9660 & 0.9348 & \textbf{0.6434} \\
       BNN &  0.9903 & \textbf{0.9844} & \textbf{0.9534} & 0.5767 \\
      \hline
       LeNet 100\% & 0.9856 & 0.9809 & 0.9678 & \textbf{0.7386} \\
       BNN 100\% & \textbf{0.9900} & \textbf{0.9863} & \textbf{0.9699} & 0.6910 \\
      \hline
       LeNet 60\% & 0.9848 & 0.9821 & 0.9740 & 0.8151 \\
       BNN 60\% & \textbf{0.9885} & \textbf{0.9864} & \textbf{0.9771} & \textbf{0.8214}\\
      \hline
       LeNet 20\% & \textbf{0.9763} & \textbf{0.9754} & 0.9695 & 0.8928 \\
       BNN 20\% & 0.9728 & 0.9735 & \textbf{0.9701} & \textbf{0.9042}\\
      \bottomrule
    \end{tabular}
  \end{center}
\vspace{-.2cm}
\caption{Classification accuracy on MNIST. We compare the
    LeNet~\cite{lecun1998gradient} implementation that is part of
    Caffe~\cite{jia2014caffe} to the network with the first layer
    replaced by a bilateral convolution layer (BCL). Both are trained
    on the original image resolution (first two rows). Three more BNN
    and CNN models are trained with randomly subsampled images (100\%,
    60\% and 20\% of the pixels). An additional bilinear interpolation
    layer samples the input signal on a spatial grid for the CNN model.
  }
  \label{tab:all-results}
\vspace{-.3cm}
\end{table}

\begin{figure}[t!]
  \centering
    \includegraphics[width=\columnwidth]{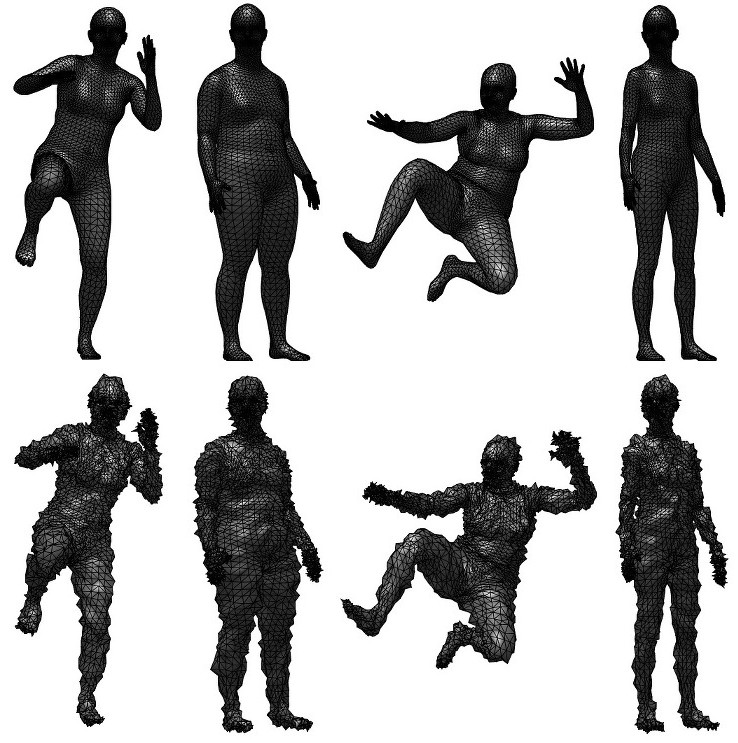}
  \mycaption{Sample data for 3D mesh denoising}
  {(top) Some 3D body meshes sampled from~\cite{SMPL:2015} and (bottom) the corresponding noisy meshes used in denoising experiments.}
\label{fig:samplebody}
\end{figure}

\subsection{Image Denoising}

The main application that inspired the development of the bilateral
filtering operation is image denoising~\cite{aurich1995non}, there
using a single Gaussian kernel. Our development allows to learn this
kernel function from data and we explore how to improve using a \emph{single}
but more general bilateral filter.

We use the Berkeley segmentation dataset
(BSDS500)~\cite{arbelaezi2011bsds500} as a test bed. The color
images in the dataset are converted to gray-scale,
and corrupted with Gaussian noise with a standard deviation of
$\frac {25} {255}$.

We compare the performance of four different filter models on a
denoising task.
The first baseline model (``Spatial'' in Table \ref{tab:denoising}, $25$
weights) uses a single spatial filter with a kernel size of
$5$ and predicts the scalar gray-scale value at the center pixel. The next model
(``Gauss Bilateral'') applies a bilateral \emph{Gaussian}
filter to the noisy input, using position and intensity features $\f=(x,y,v)^\top$.
The third setup (``Learned Bilateral'', $65$ weights)
takes a Gauss kernel as initialization and
fits all filter weights on the ``train'' image set to minimize the
mean squared error with respect to the clean images.
We run a combination
of spatial and permutohedral convolutions on spatial and bilateral
features (``Spatial + Bilateral (Learned)'') to check for a complementary
performance of the two convolutions.

\label{sec:exp:denoising}
\begin{table}[!h]
\begin{center}
  \footnotesize
  \begin{tabular}[t]{lr}
    \toprule
    Method & PSNR \\
    \midrule
    Noisy Input & $20.17$ \\
    Spatial & $26.27$ \\
    Gauss Bilateral & $26.51$ \\
    Learned Bilateral & $26.58$ \\
    Spatial + Bilateral (Learned) & $26.65$ \\
    \bottomrule
  \end{tabular}
\end{center}
\vspace{-0.5em}
\caption{PSNR results of a denoising task using the BSDS500
  dataset~\cite{arbelaezi2011bsds500}}
\vspace{-0.5em}
\label{tab:denoising}
\end{table}

The PSNR scores evaluated on full images of the ``test'' image set are
shown in Table \ref{tab:denoising}. We find that an untrained bilateral
filter already performs better than a trained spatial convolution
($26.27$ to $26.51$). A learned convolution further improve the
performance slightly. We chose this simple one-kernel setup to
validate an advantage of the generalized bilateral filter. A competitive
denoising system would employ RGB color information and also
needs to be properly adjusted in network size. Multi-layer perceptrons
have obtained state-of-the-art denoising results~\cite{burger12cvpr}
and the permutohedral lattice layer can readily be used in such an
architecture, which is intended future work.

\subsection{3D Mesh Denoising}\label{sec:mesh_denoising}

Permutohedral convolutions can naturally be extended to higher ($>2$)
dimensional data. To highlight this, we use the proposed convolution for the task
of denoising 3D meshes.

\begin{figure}[t!]
  \centering
    \includegraphics[width=\columnwidth]{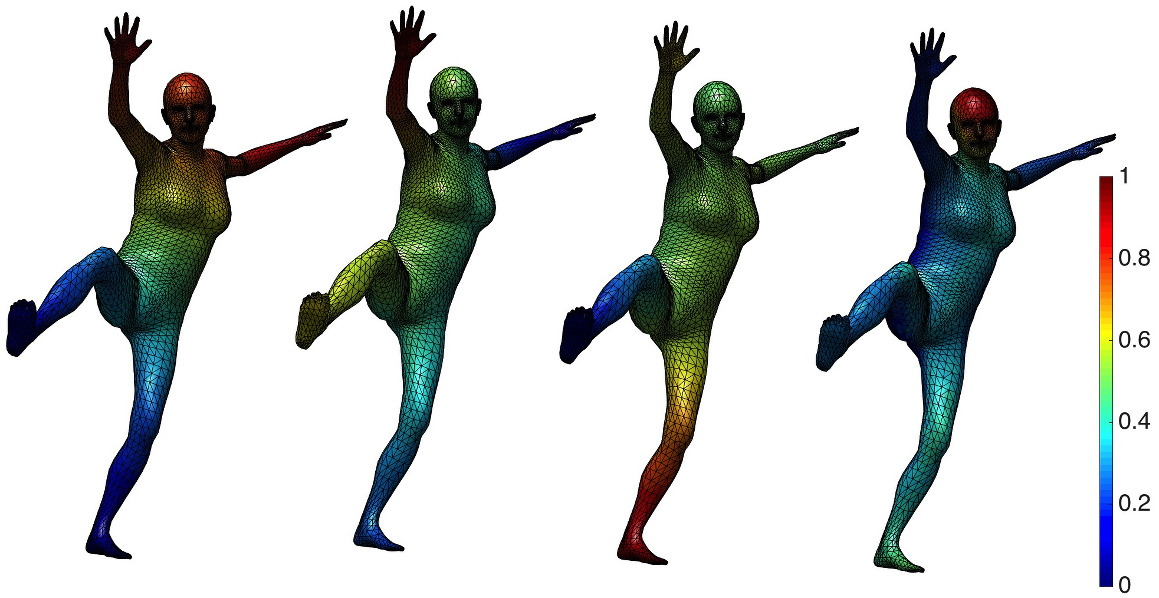}
  \mycaption{4D isomap features for 3D human bodies}
  {Visualization of 4D isomap features for a sample 3D mesh. Isomap feature values are overlaid onto mesh vertices.}
  \label{fig:isomap}
\end{figure}

We sample 3D human body meshes using a
generative 3D body model from~\cite{SMPL:2015}. To the clean meshes, we add Gaussian random
noise displacements along the surface normal at each vertex location. Figure~\ref{fig:samplebody} shows
some sample 3D meshes sampled
from~\cite{SMPL:2015} and corresponding noisy meshes. The task is to take the noisy meshes as inputs and
recover the original 3D body meshes. We create 1000 training, 200 validation and another 500
testing examples for the experiments.

\paragraph{Mesh Representation:} The 3D human body meshes from~\cite{SMPL:2015} are represented with
3D vertex locations and the edge connections between the vertices. We found that this signal representation
using global 3D coordinates is not suitable for denoising with bilateral filtering. Therefore, we first
smooth the noisy mesh using mean smoothing applied to the face normals~\cite{yagou2002mesh} and represent the noisy
mesh vertices as 3D vector displacements with respect to the corresponding smoothed mesh. Thus, the
task becomes denoising the 3D vector displacements with respect to the smoothed mesh.

\paragraph{Isomap Features:} To apply permutohedral convolution, we need to define features at each input
vertex point. We use 4 dimensional isomap embedding~\cite{tenenbaum2000global} of the given 3D mesh graph as features. The
given 3D mesh is converted into weighted edge graph with edge weights set to Euclidean distance between the
connected vertices and to infinity between the non-connected vertices. Then 4 dimensional isomap embedding is
computed for this weighted edge graph using a publicly available implementation~\cite{isomap_code}. Fig.~\ref{fig:isomap}
shows the visualization of isomap features on a sample 3D mesh.

\setlength{\tabcolsep}{2pt}
\newcolumntype{L}[1]{>{\raggedright\let\newline\\\arraybackslash\hspace{0pt}}b{#1}}
\newcolumntype{C}[1]{>{\centering\let\newline\\\arraybackslash\hspace{0pt}}b{#1}}
\newcolumntype{R}[1]{>{\raggedleft\let\newline\\\arraybackslash\hspace{0pt}}b{#1}}
\begin{table}[t]
  \scriptsize
  \centering
    \begin{tabular}{C{2.0cm} C{1.5cm} C{1.5cm} C{1.5cm} C{1.5cm}}
      \toprule
& \textbf{Noisy Mesh} & \textbf{Normal Smoothing} & \textbf{Gauss Bilateral} & \textbf{Learned Bilateral} \\ [0.1cm]
      \midrule
\textbf{Vertex Distance (RMSE)} &  5.774 & 3.183 & 2.872 & \textbf{2.825}  \\
\textbf{Normal Angle Error} & 19.680  & 19.707 & 19.357 & \textbf{19.207}  \\
      \bottomrule
      \\
    \end{tabular}
      \vspace{-0.2cm}
    \mycaption{Body Denoising} {Vertex distance RMSE values and normal angle error (in degrees)
    corresponding to different denoising strategies averaged over 500 test meshes.}
\label{tbl:bodydenoise}
\end{table}

\begin{figure}[t!]
  \centering
    \includegraphics[width=\columnwidth]{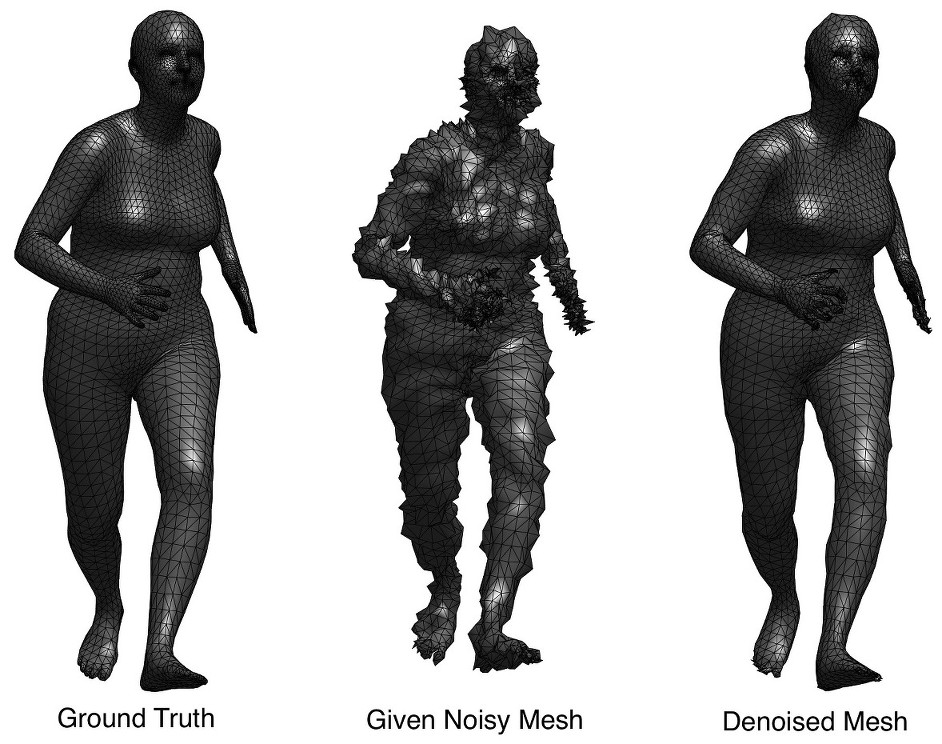}
  \mycaption{Sample Denoising Result}
  {Ground truth mesh (left), corresponding given noisy mesh (middle) and the denoised result (right) using learned bilateral filter.}
  \label{fig:bodyresult}
\end{figure}

\paragraph{Experimental Results:} Mesh denoising with a bilateral filter proceeds by
splatting the input 3D mesh vectors (displacements with respect to smoothed mesh) into
the 4D isomap feature space, filtering the signal in this 4D space and then slicing back
into original 3D input space. The Table~\ref{tbl:bodydenoise} shows quantitative results as RMSE
for different denoising strategies. The normal smoothing~\cite{yagou2002mesh} already reduces the RMSE. The Gauss bilateral filter
results in significant improvement over normal
smoothing with and learning the filter weights again improves the result. A visual result is shown in Figure \ref{fig:bodyresult}.

\section{Additional results}
\label{sec:addresults}

This section contains more qualitative results for the experiments of the main paper.

\begin{figure*}[th!]
  \centering
    \includegraphics[width=2\columnwidth,trim={5cm 2.5cm 5cm 4.5cm},clip]{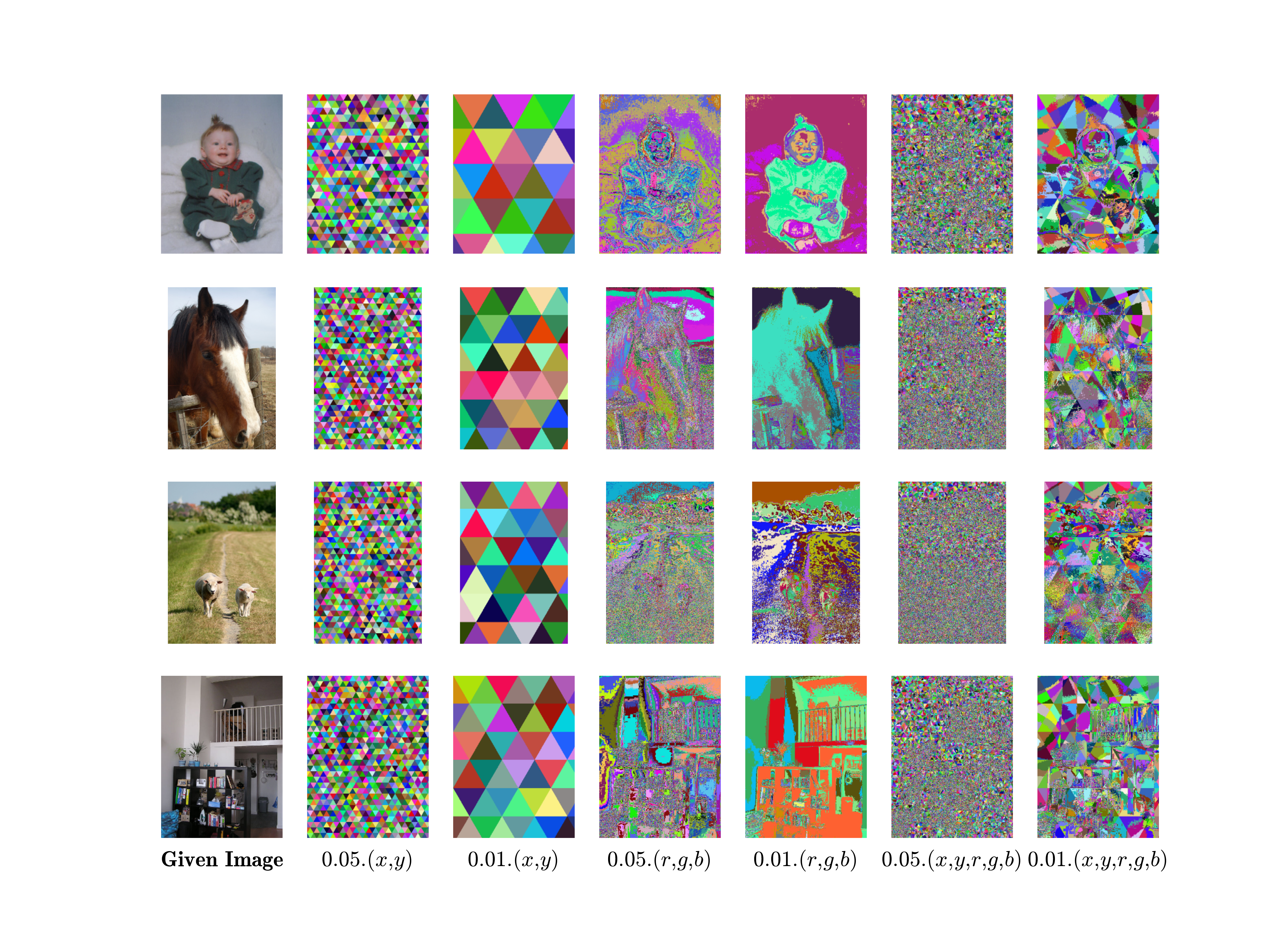}
  \mycaption{Visualization of the Permutohedral Lattice}
  {Sample lattice visualizations for different feature spaces. All pixels falling in the same simplex cell are shown with
  the same color. $(x,y)$ features correspond to image pixel positions, and $(r,g,b) \in [0,255]$ correspond
  to the red, green and blue color values.}
\label{fig:latticeviz}
\end{figure*}

\subsection{Lattice Visualization}

Figure~\ref{fig:latticeviz} shows sample lattice visualizations for different feature spaces.

\begin{table}[t]
  \scriptsize
  \centering
    \begin{tabular}{c c c c c c}
      \toprule
      & & \multicolumn{4}{c}{Test Factor} \\ [0.1cm]
     & & \textbf{2$\times$} & \textbf{4$\times$} & \textbf{8$\times$} & \textbf{16$\times$} \\ [0.15cm]
    \parbox[t]{3mm}{\multirow{4}{*}{\rotatebox[origin=c]{90}{Train Factor}}} & \textbf{2$\times$} & \textbf{38.45} & 36.12  & 34.06 & 32.43 \\ [0.1cm]
     & \textbf{4$\times$} & 38.40 & \textbf{36.16} & \textbf{34.08} & 32.47 \\ [0.1cm]
     & \textbf{8$\times$} & 38.40 & 36.15  & \textbf{34.08} & 32.47 \\ [0.1cm]
     & \textbf{16$\times$} & 38.26 & 36.13  & 34.06 & \textbf{32.49} \\
      \bottomrule
      \\
    \end{tabular}
      \vspace{-0.2cm}
    \mycaption{Color Upsampling with different train and test upsampling factors} {PSNR values corresponding to different upsampling factors used at train and test times on the 2 megapixel image dataset, using our learned bilateral filters.}
\label{tbl:crossupsample}
\vspace{-0.4cm}
\end{table}

\subsection{Color Upsampling}\label{sec:color_upsampling}

\begin{figure*}[t!]
  \centering
    \subfigure{%
   \raisebox{2.0em}{
    \includegraphics[width=.12\columnwidth]{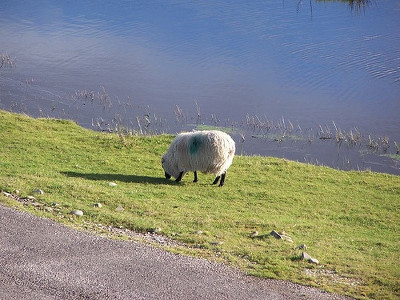}
   }
  }
  \subfigure{%
    \includegraphics[width=.36\columnwidth]{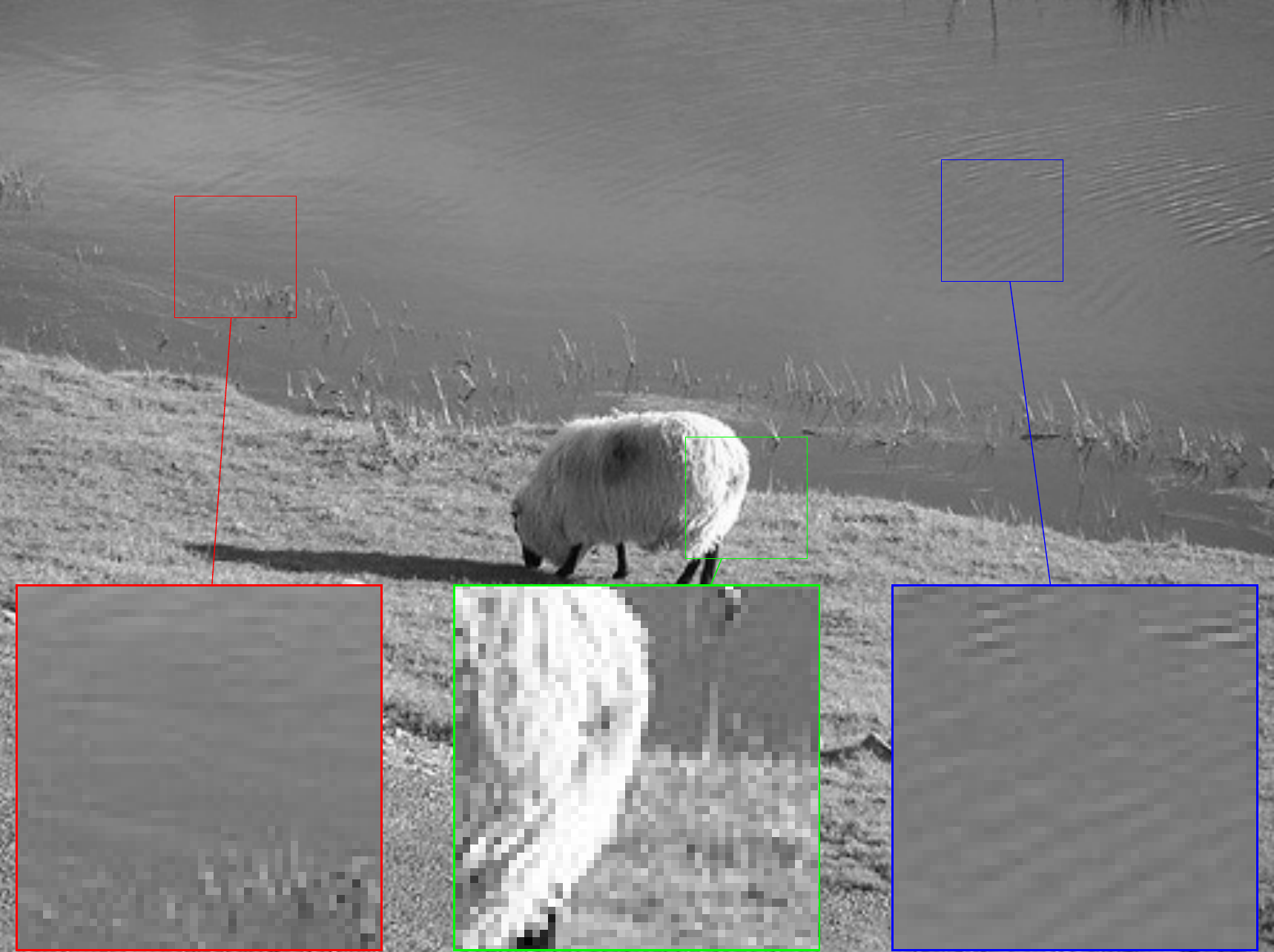}
  }
  \subfigure{%
    \includegraphics[width=.36\columnwidth]{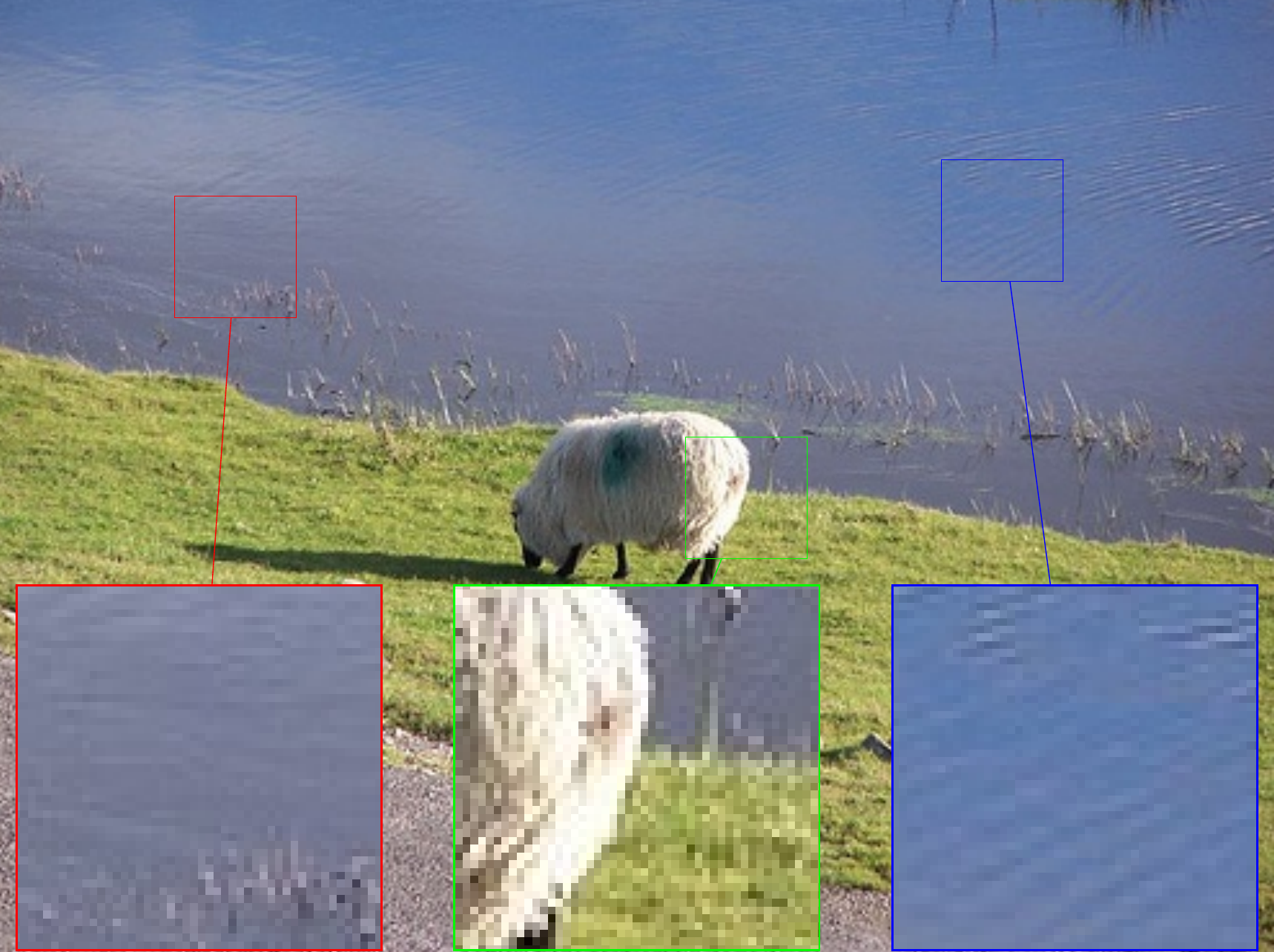}
  }
  \subfigure{%
    \includegraphics[width=.36\columnwidth]{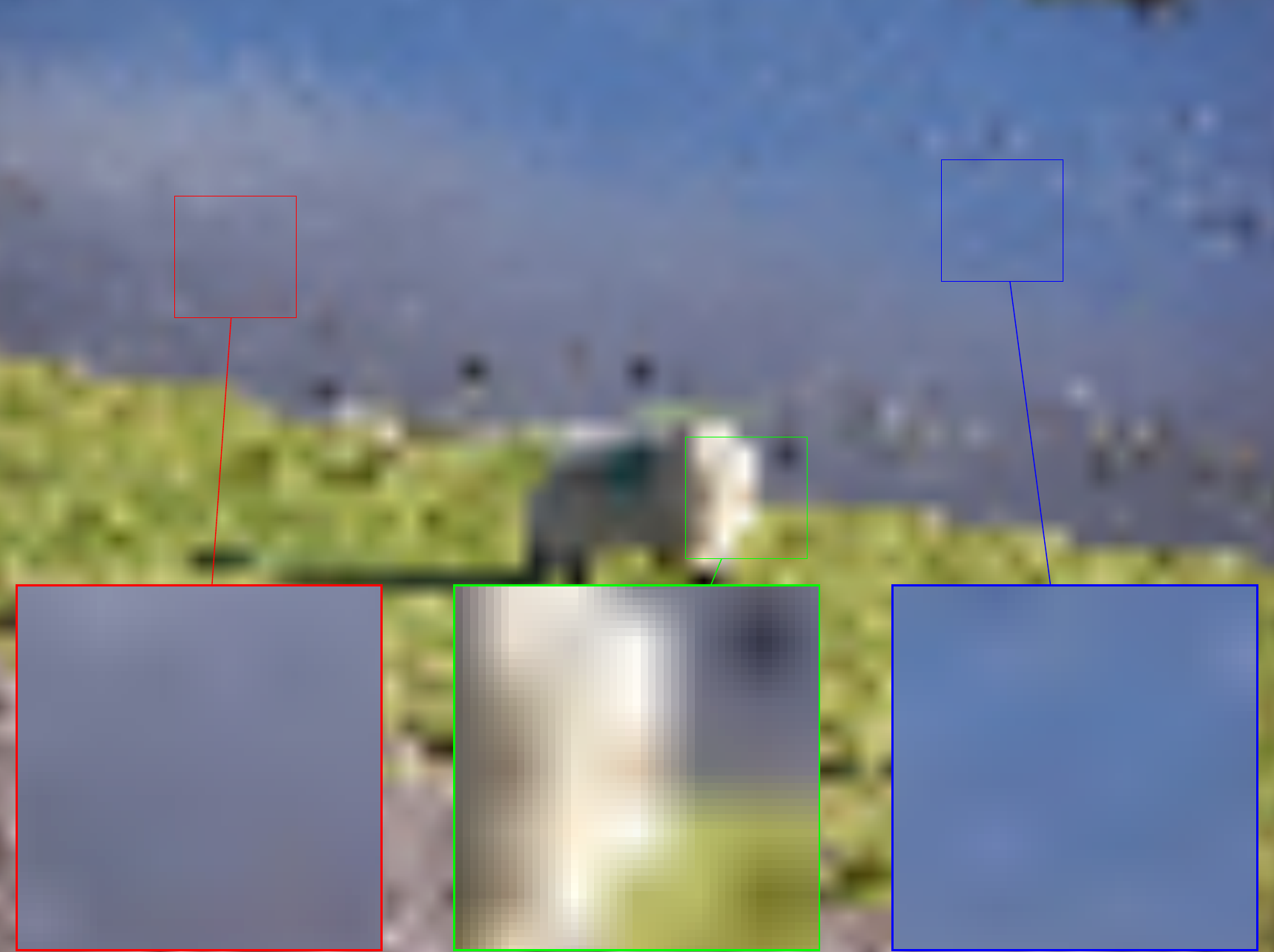}
  }
  \subfigure{%
    \includegraphics[width=.36\columnwidth]{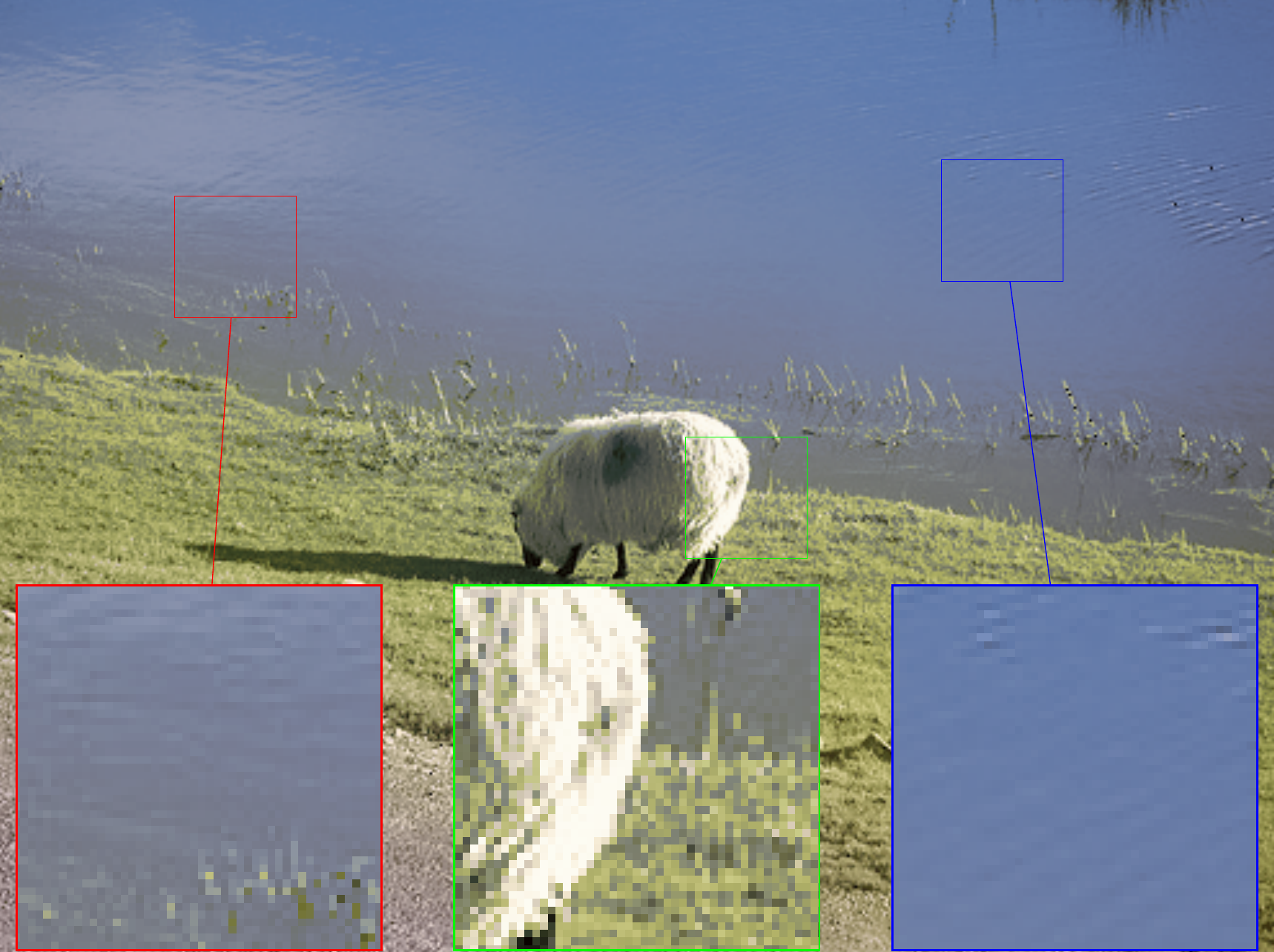}
  }
  \subfigure{%
    \includegraphics[width=.36\columnwidth]{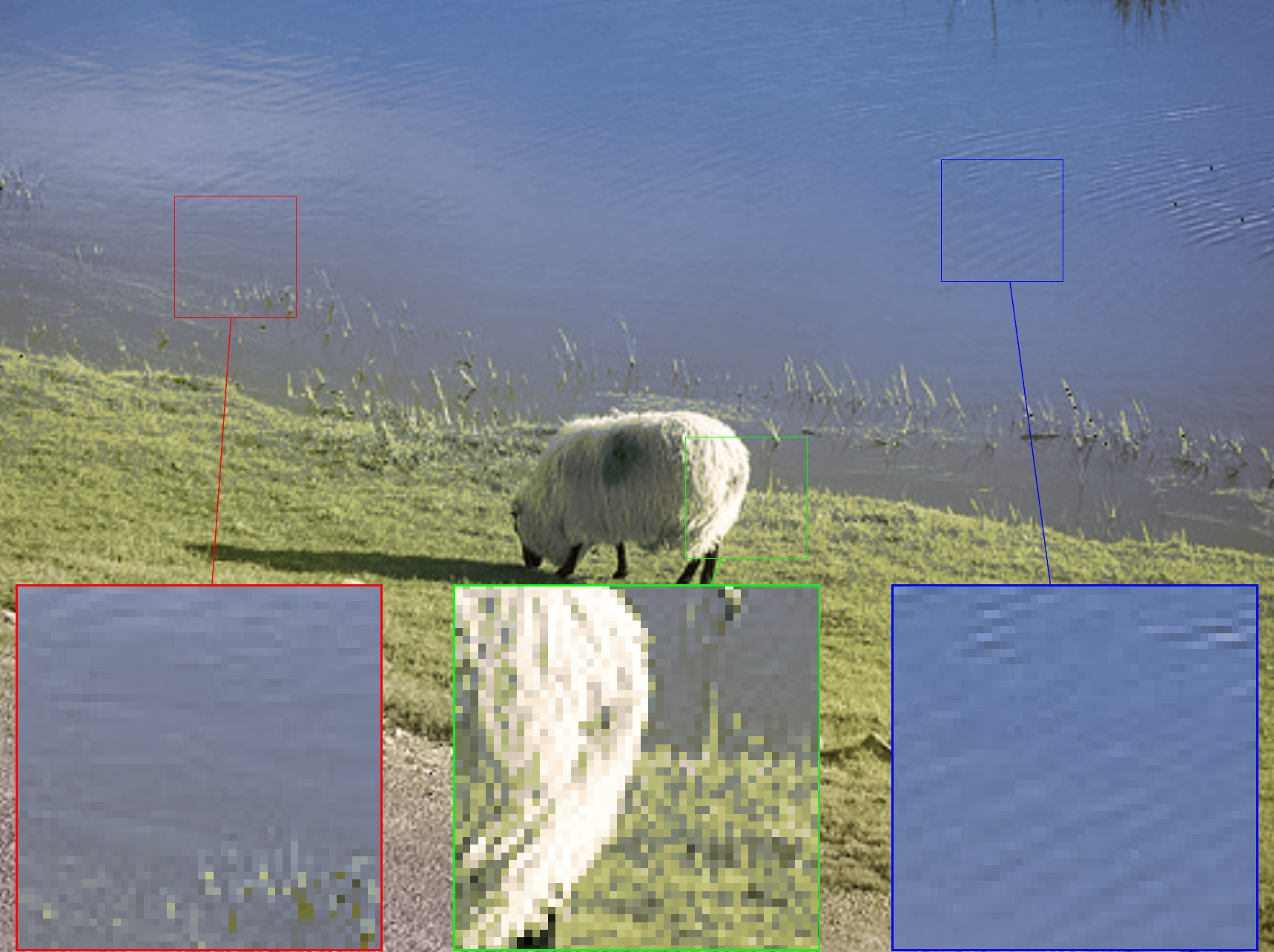}
  }\\
    \subfigure{%
   \raisebox{2.0em}{
    \includegraphics[width=.12\columnwidth]{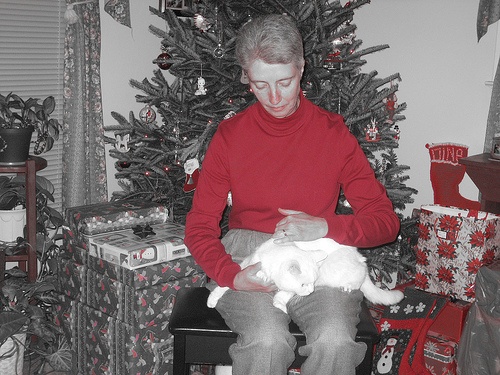}
   }
  }
  \subfigure{%
    \includegraphics[width=.36\columnwidth]{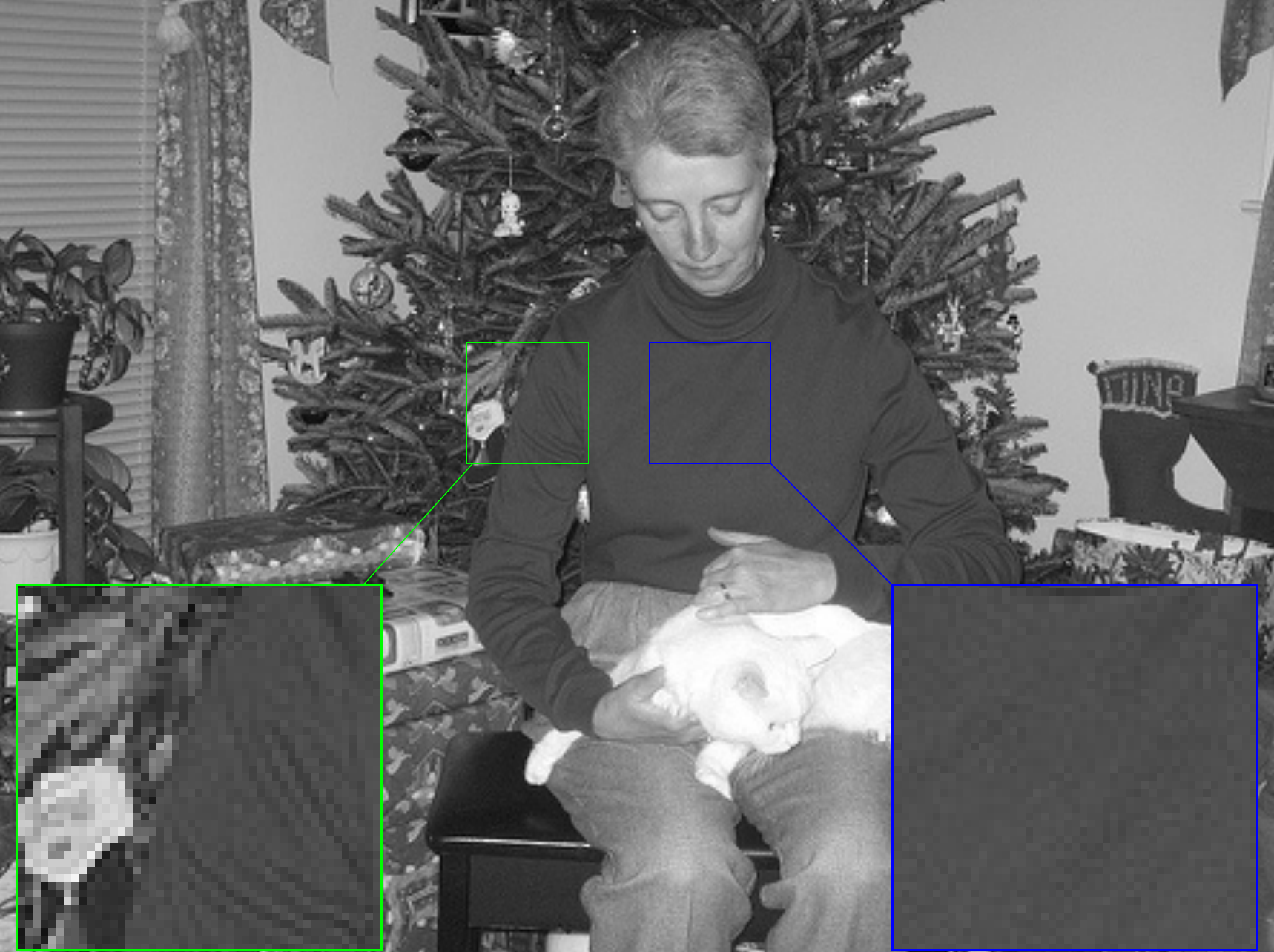}
  }
  \subfigure{%
    \includegraphics[width=.36\columnwidth]{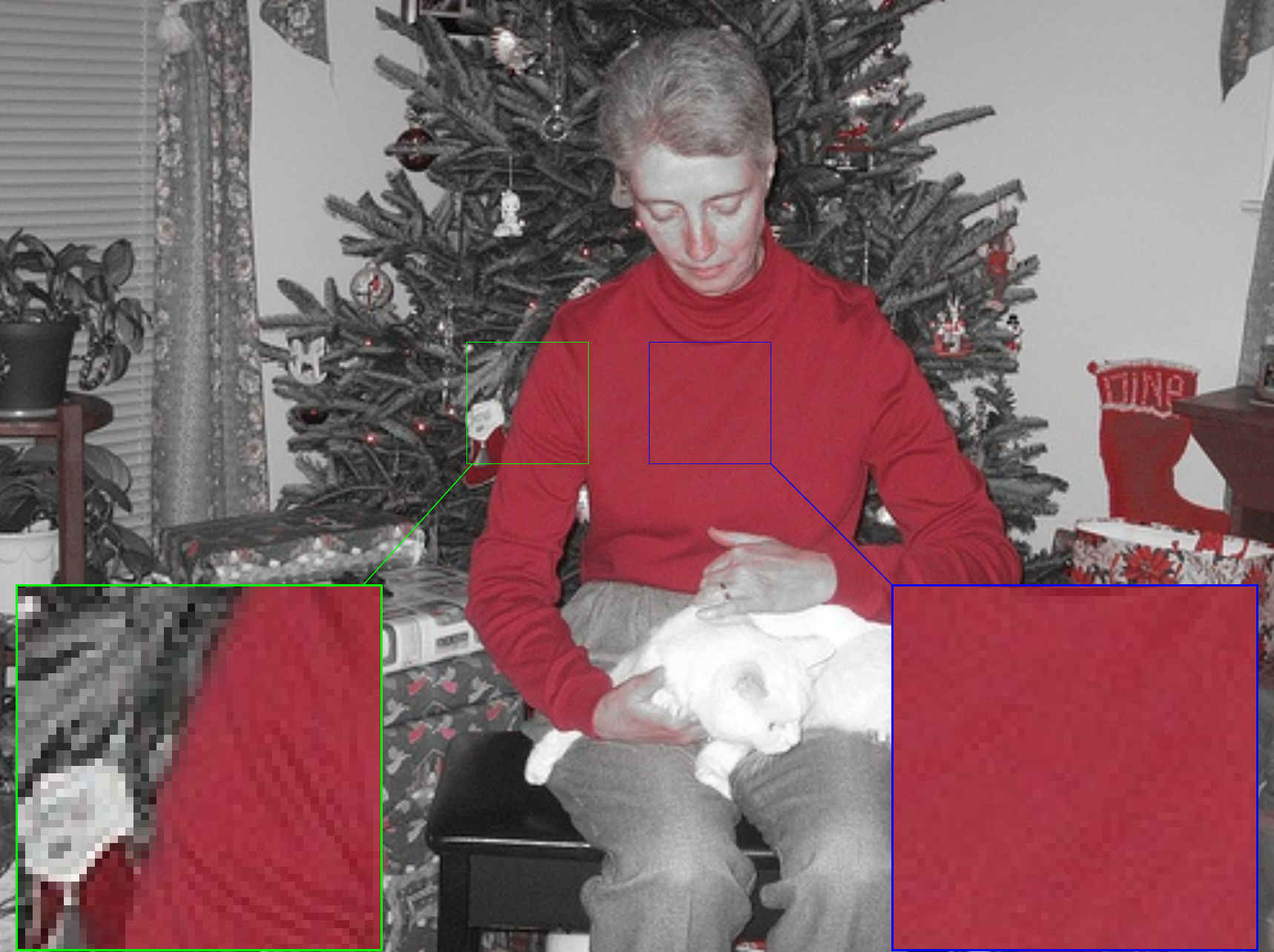}
  }
  \subfigure{%
    \includegraphics[width=.36\columnwidth]{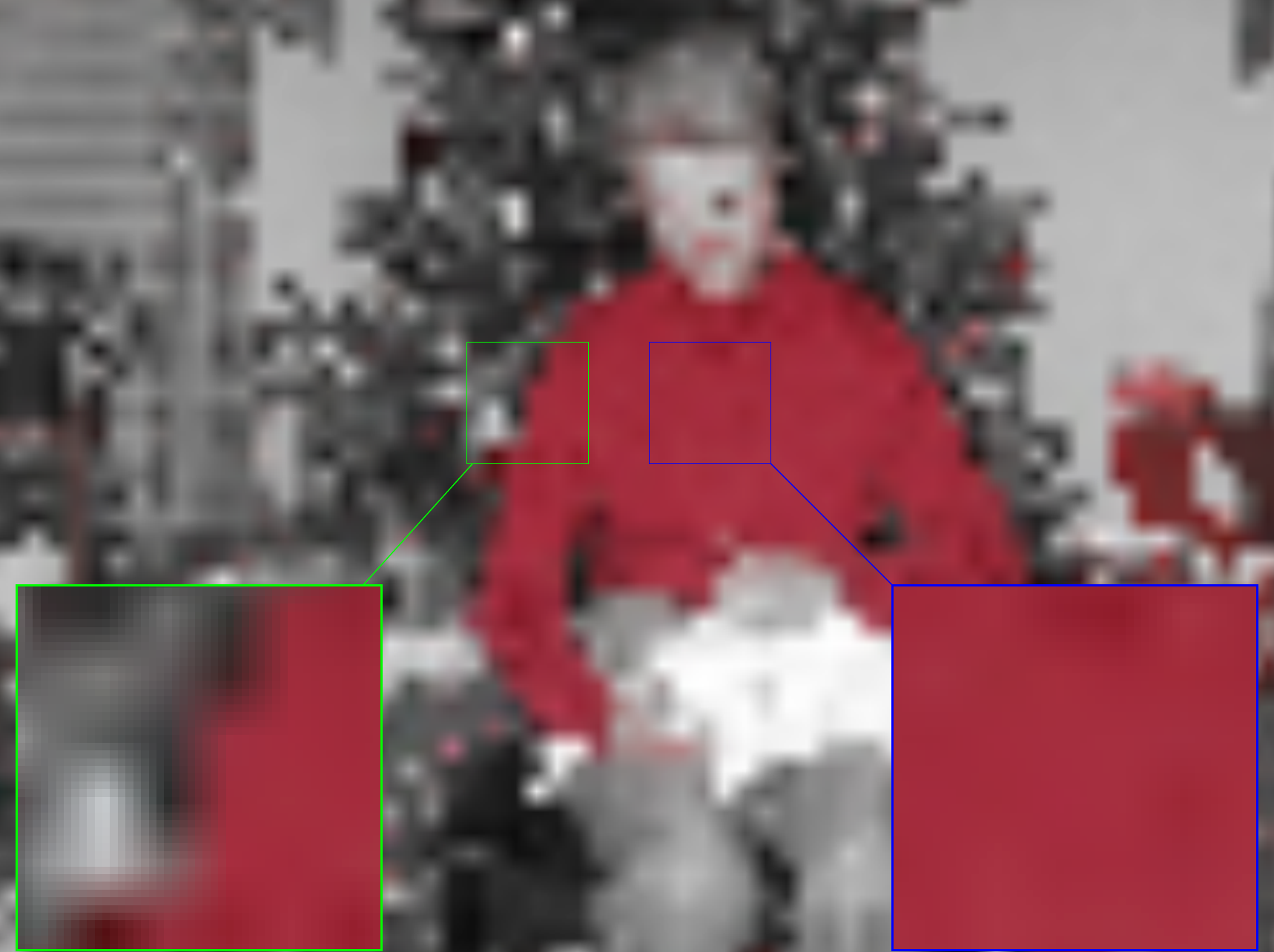}
  }
  \subfigure{%
    \includegraphics[width=.36\columnwidth]{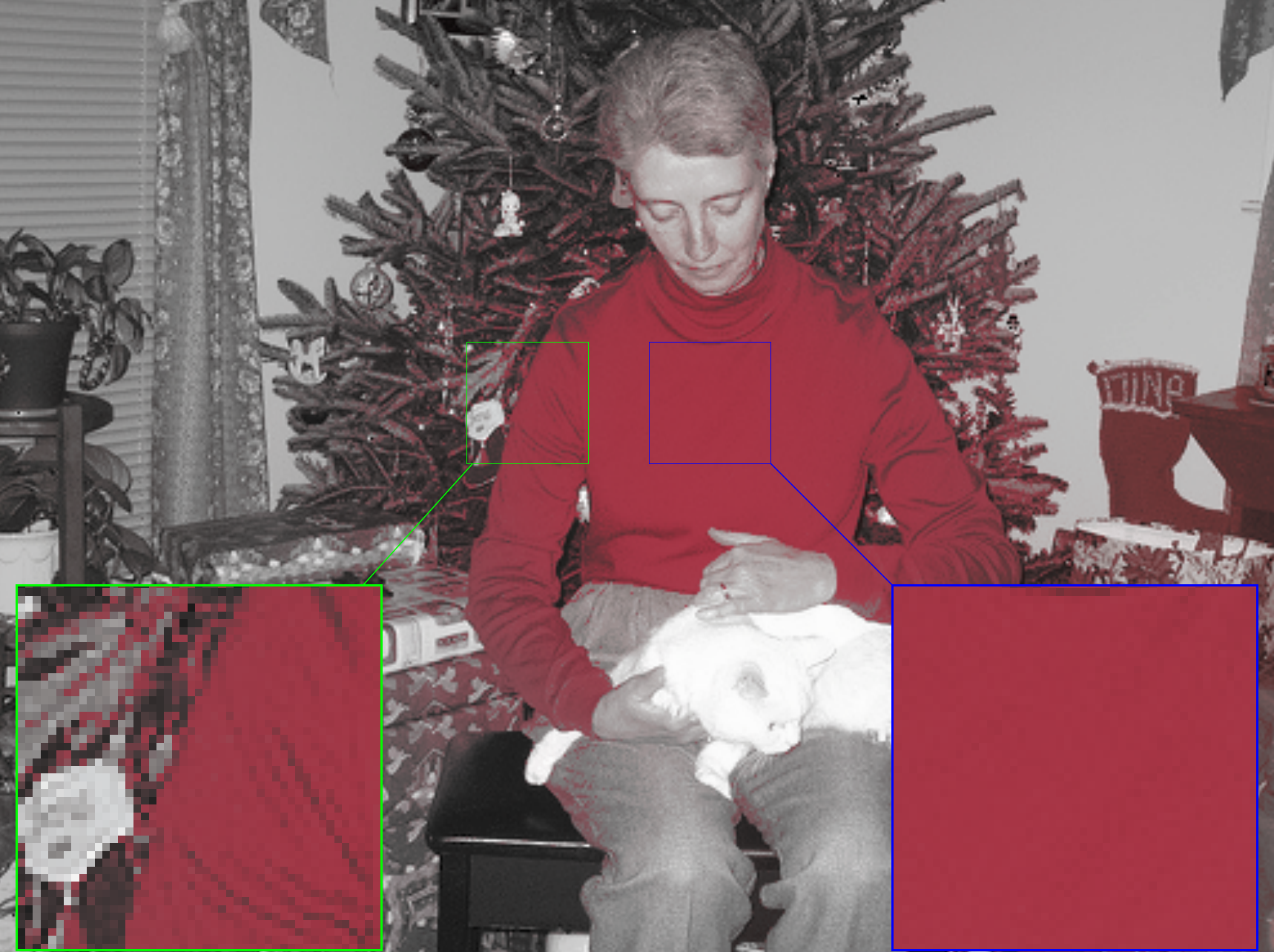}
  }
  \subfigure{%
    \includegraphics[width=.36\columnwidth]{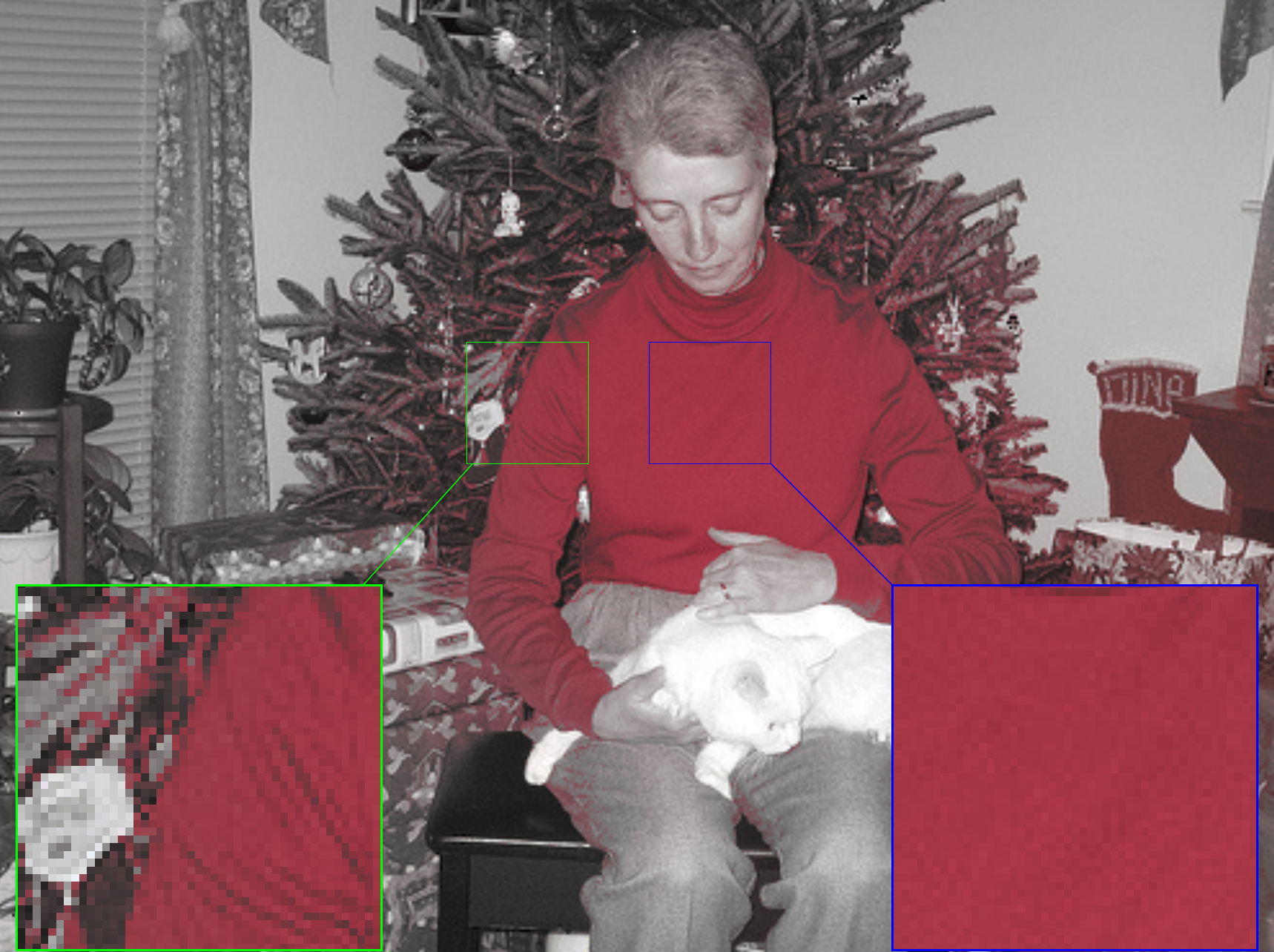}
  }\\
  \setcounter{subfigure}{0}
  \subfigure[Input]{%
  \raisebox{2.0em}{
    \includegraphics[width=.12\columnwidth]{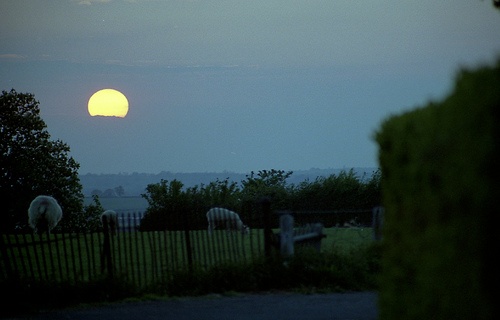}
   }
  }
  \subfigure[Gray Guidance]{%
    \includegraphics[width=.36\columnwidth]{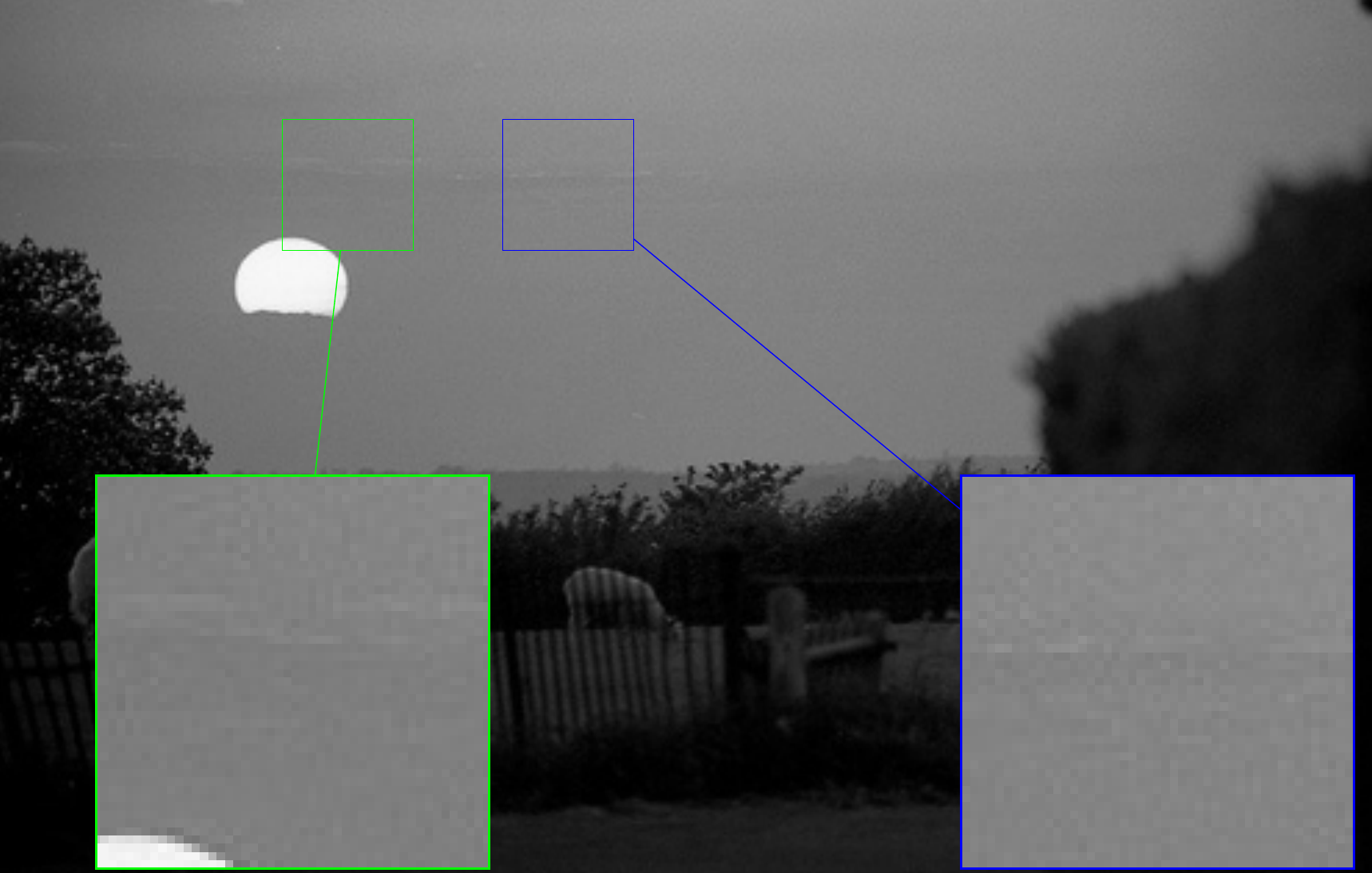}
  }
   \subfigure[Ground Truth]{%
    \includegraphics[width=.36\columnwidth]{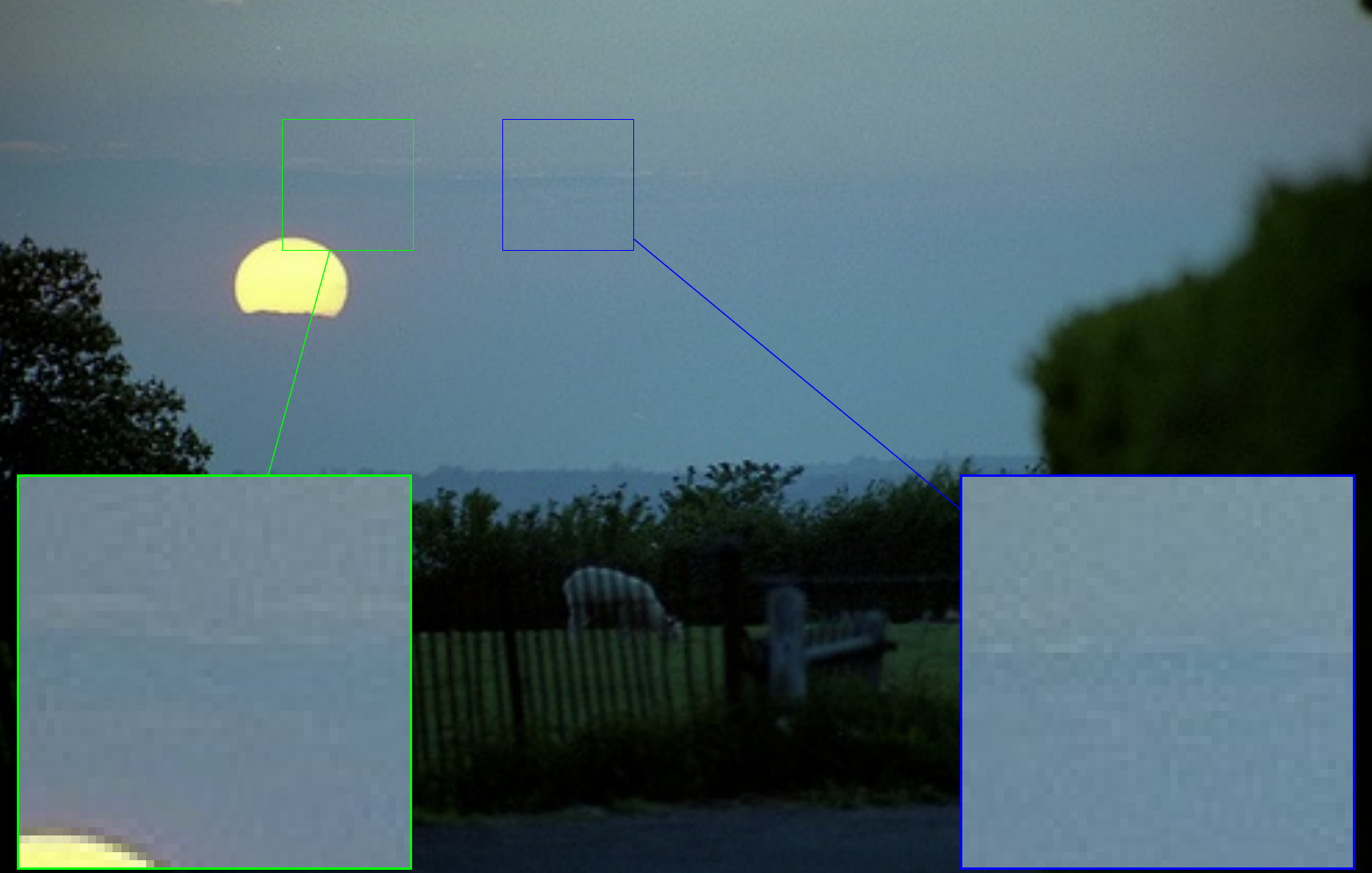}
  }
  \subfigure[Bicubic Interpolation]{%
    \includegraphics[width=.36\columnwidth]{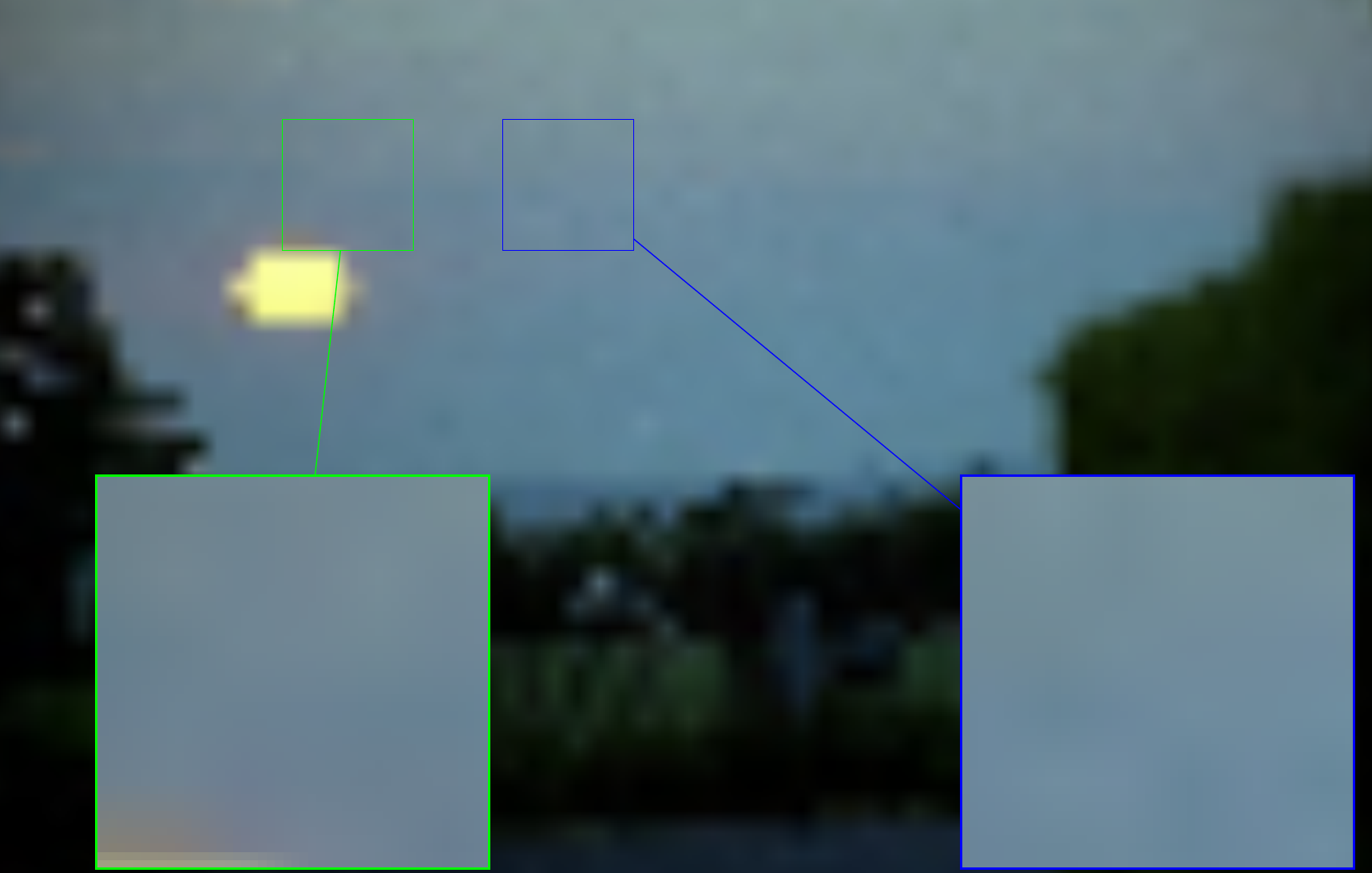}
  }
  \subfigure[Gauss Bilateral]{%
    \includegraphics[width=.36\columnwidth]{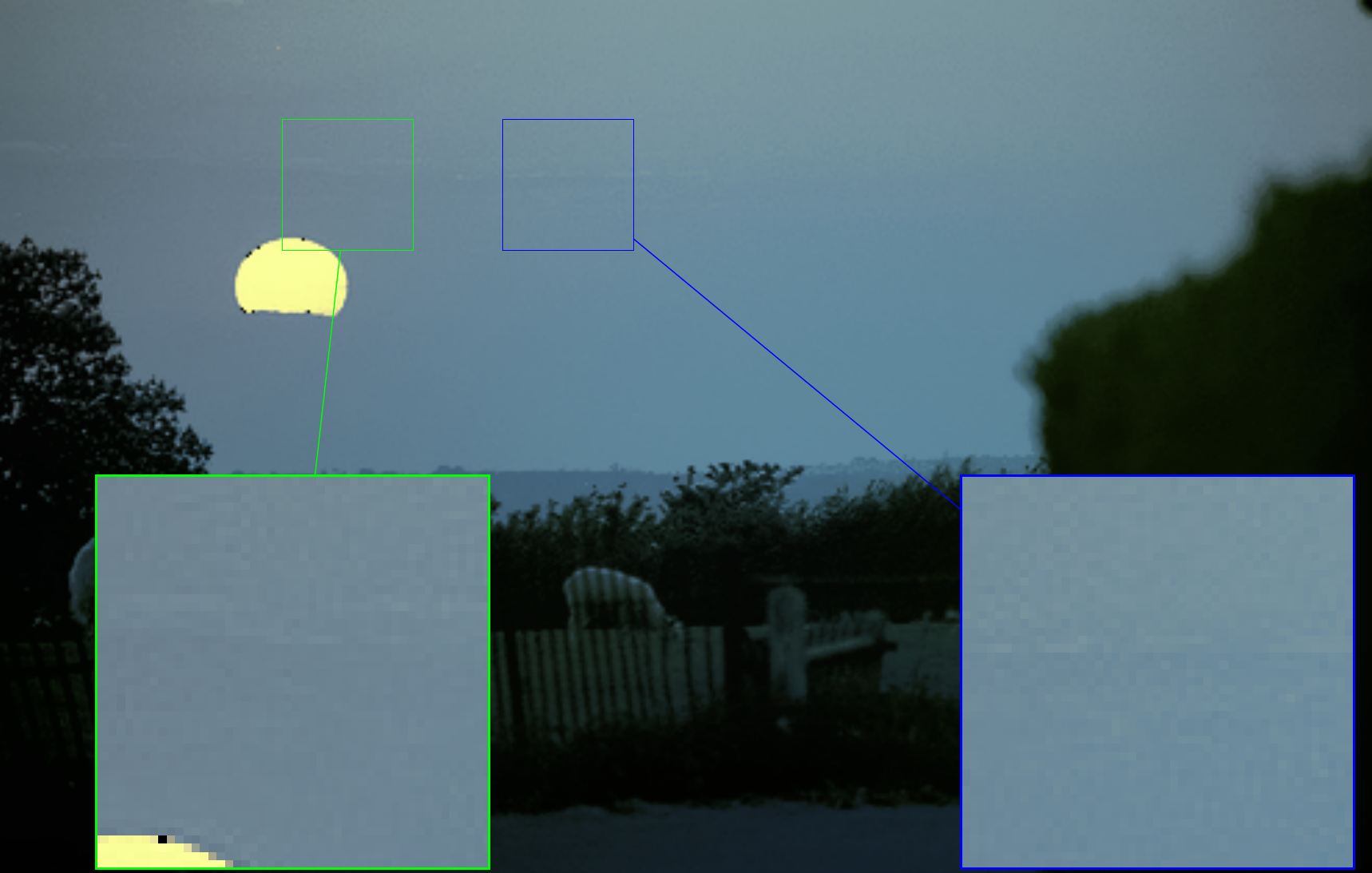}
  }
  \subfigure[Learned Bilateral]{%
    \includegraphics[width=.36\columnwidth]{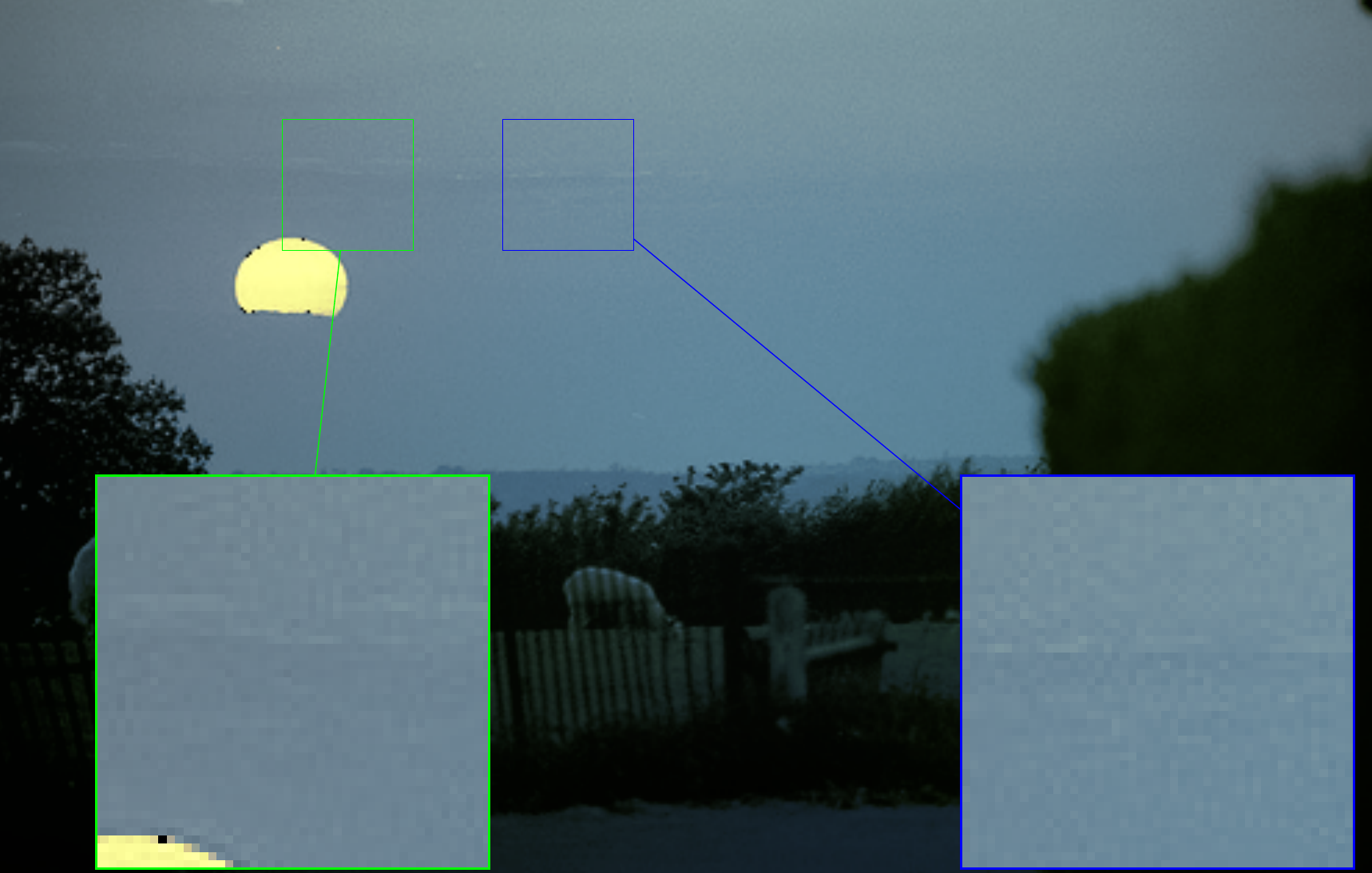}
  }
  \mycaption{Color Upsampling}{Color $8\times$ upsampling results
  using different methods (best viewed on screen).}

\label{fig:Colour_upsample_visuals}
\end{figure*}

In addition to the experiments discussed in the main paper, we performed
the cross-factor analysis of training and testing at different upsampling
factors. Table~\ref{tbl:crossupsample} shows the PSNR results for this
analysis. Although, in terms of PSNR, it is optimal to train and test at the
same upsampling factor, the differences are small when training and testing
upsampling factors are different. Some images of the upsampling for the Pascal
VOC12 dataset are shown in Fig.~\ref{fig:Colour_upsample_visuals}. It is
especially the low level image details that are better preserved with
a learned bilateral filter compared to the Gaussian case.

\subsection{Depth Upsampling}

\begin{figure*}[t!]
  \centering
    \subfigure{%
   \raisebox{2.0em}{
    \includegraphics[width=.12\columnwidth]{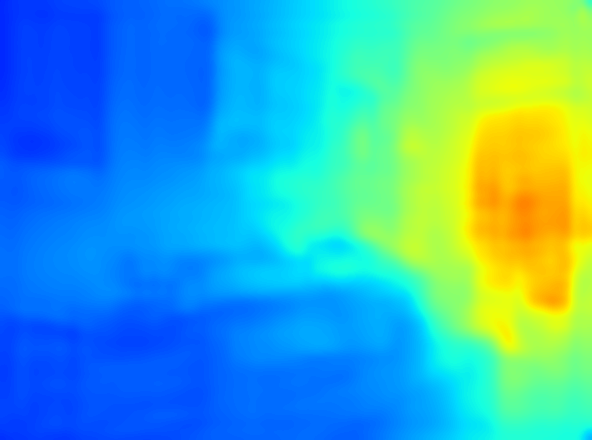}
   }
  }
  \subfigure{%
    \includegraphics[width=.36\columnwidth]{figures/supplementary/2given_image}
  }
  \subfigure{%
    \includegraphics[width=.36\columnwidth]{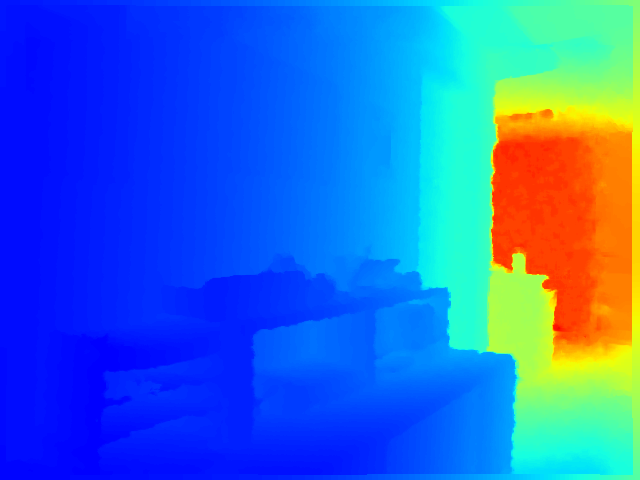}
  }
  \subfigure{%
    \includegraphics[width=.36\columnwidth]{figures/supplementary/2bicubic}
  }
  \subfigure{%
    \includegraphics[width=.36\columnwidth]{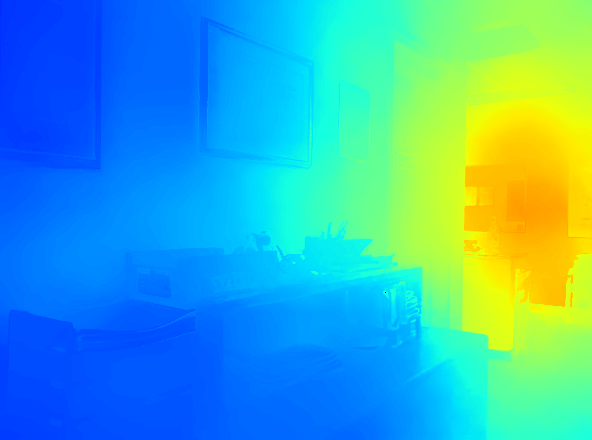}
  }
  \subfigure{%
    \includegraphics[width=.36\columnwidth]{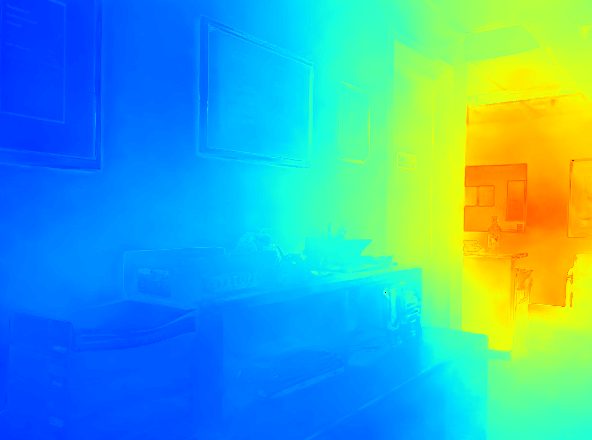}
  }\\
    \subfigure{%
   \raisebox{2.0em}{
    \includegraphics[width=.12\columnwidth]{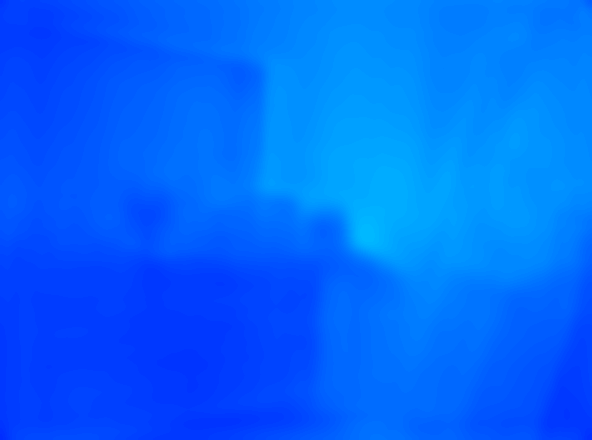}
   }
  }
  \subfigure{%
    \includegraphics[width=.36\columnwidth]{figures/supplementary/32given_image}
  }
  \subfigure{%
    \includegraphics[width=.36\columnwidth]{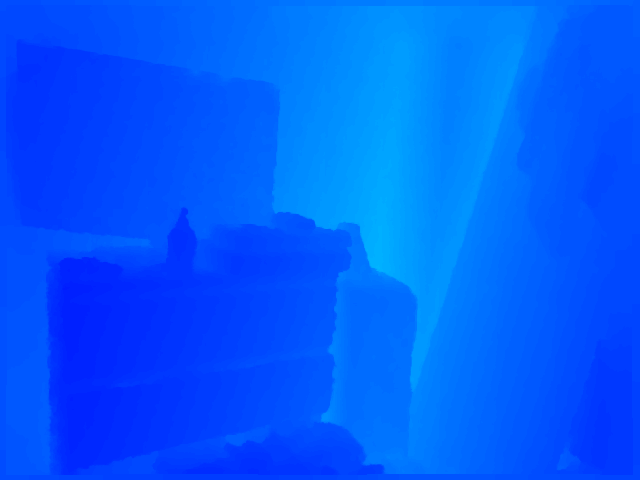}
  }
  \subfigure{%
    \includegraphics[width=.36\columnwidth]{figures/supplementary/32bicubic}
  }
  \subfigure{%
    \includegraphics[width=.36\columnwidth]{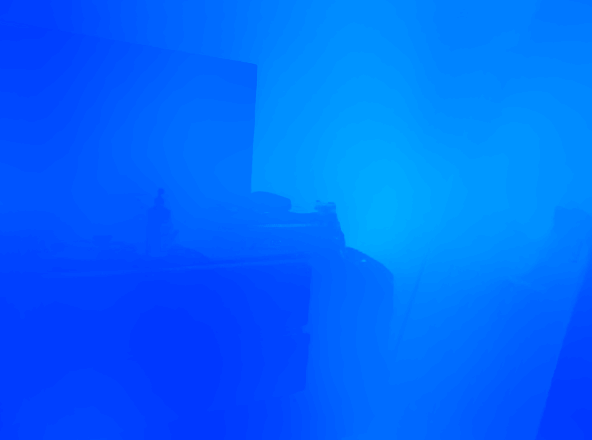}
  }
  \subfigure{%
    \includegraphics[width=.36\columnwidth]{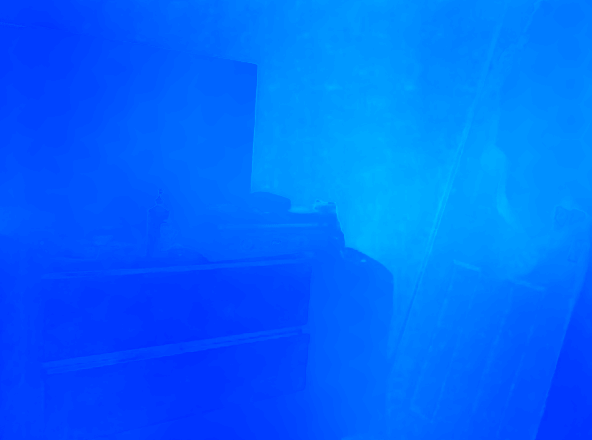}
  }\\
  \setcounter{subfigure}{0}
  \subfigure[Input]{%
  \raisebox{2.0em}{
    \includegraphics[width=.12\columnwidth]{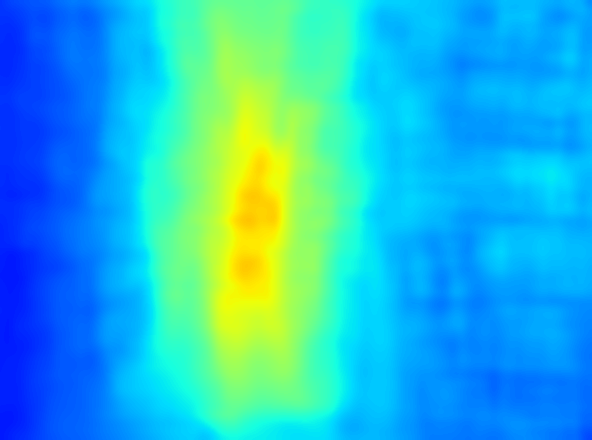}
   }
  }
  \subfigure[Guidance]{%
    \includegraphics[width=.36\columnwidth]{figures/supplementary/41given_image}
  }
   \subfigure[Ground Truth]{%
    \includegraphics[width=.36\columnwidth]{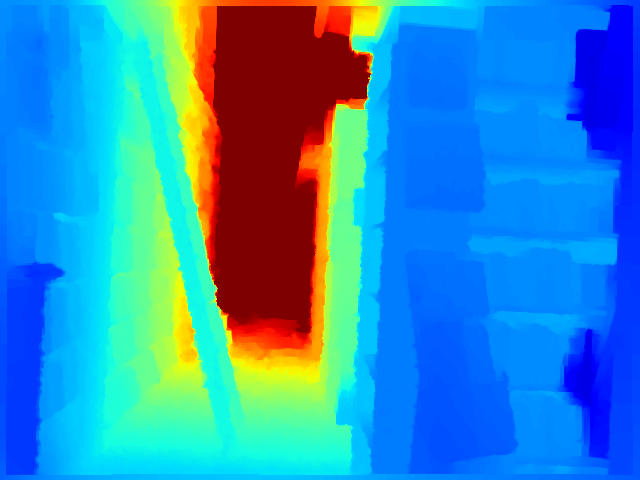}
  }
  \subfigure[Bicubic Interpolation]{%
    \includegraphics[width=.36\columnwidth]{figures/supplementary/41bicubic}
  }
  \subfigure[Gauss Bilateral]{%
    \includegraphics[width=.36\columnwidth]{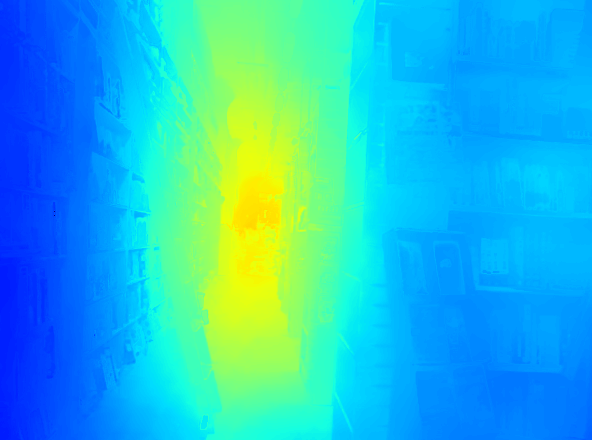}
  }
  \subfigure[Learned Bilateral]{%
    \includegraphics[width=.36\columnwidth]{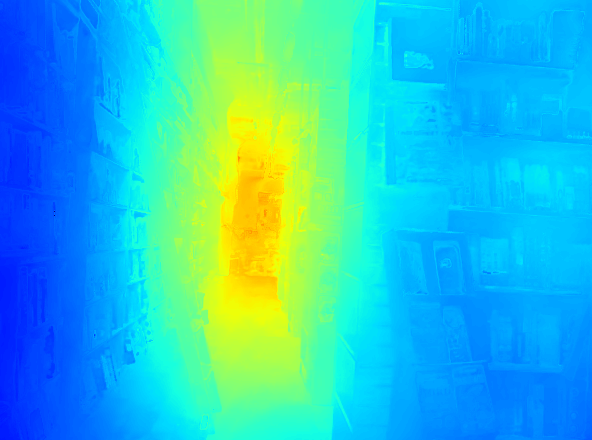}
  }
  \mycaption{Depth Upsampling}{Depth $8\times$ upsampling results
  using different upsampling strategies.}

\label{fig:depth_upsample_visuals}
\end{figure*}

Figure~\ref{fig:depth_upsample_visuals} presents some more qualitative results comparing bicubic interpolation, Gauss
bilateral and learned bilateral upsampling on NYU depth dataset image~\cite{silberman2012indoor}.

\subsection{Character Recognition}\label{sec:app_character}

\begin{figure*}[t!]
  \centering
  \subfigure[LeNet-7]{
    \includegraphics[width=1.5\columnwidth]{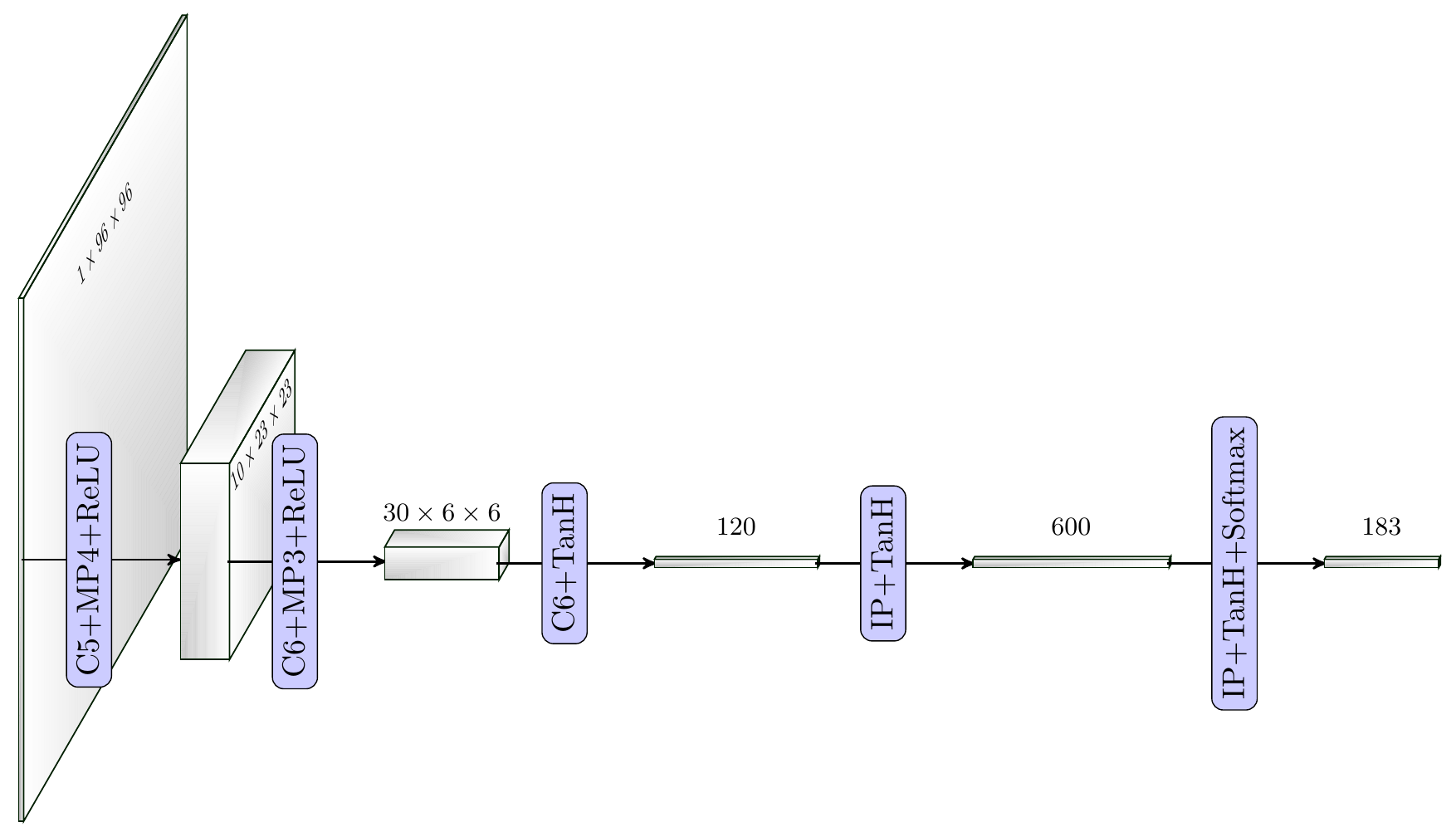}
    }\\
    \subfigure[DeepCNet]{
    \includegraphics[width=2\columnwidth]{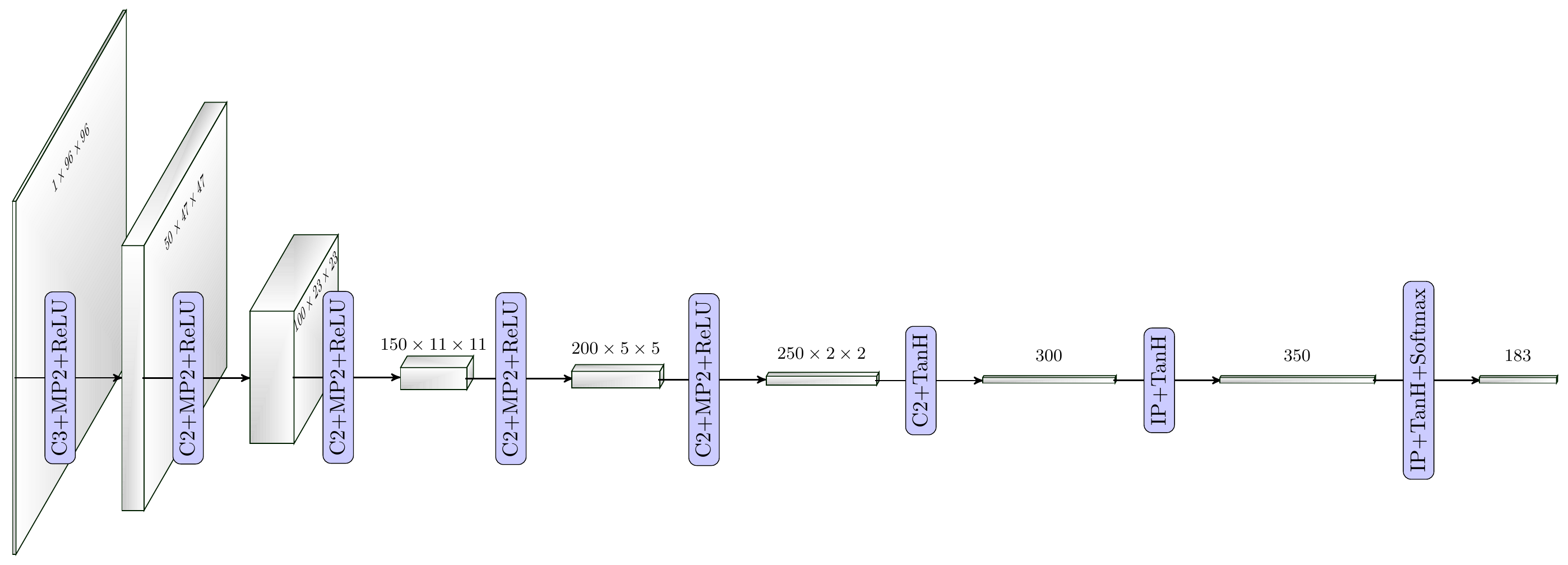}
    }
  \mycaption{CNNs for Character Recognition}
  {Schematic of (top) LeNet-7~\cite{lecun1998mnist} and (bottom) DeepCNet(5,50)~\cite{ciresan2012multi,graham2014spatially} architectures used in Assamese
  character recognition experiments.}
\label{fig:nnrecognition}
\end{figure*}

 Figure~\ref{fig:nnrecognition} shows the schematic of different layers
 of the network architecture for LeNet-7~\cite{lecun1998mnist}
 and DeepCNet(5, 50)~\cite{ciresan2012multi,graham2014spatially}. For the BNN variants, the first layer filters are replaced
 with learned bilateral filters and are learned end-to-end.

\subsection{Semantic Segmentation}\label{sec:app_semantic_segmentation}

Some more visual results for semantic segmentation are shown in Figure~\ref{fig:semantic_visuals}.
These include the underlying DeepLab CNN\cite{chen2014semantic} result (DeepLab),
the 2 step mean-field result with Gaussian edge potentials (+2stepMF-GaussCRF)
and also corresponding results with learned edge potentials (+2stepMF-LearndCRF).
In general, we observe that mean-field in learned CRF leads to slightly dilated
classification regions in comparison to using Gaussian CRF thereby filling-in the
false negative pixels and also correcting some mis-classified regions.

\subsection{Material Segmentation}\label{sec:app_material_segmentation}

In Fig.~\ref{fig:material_visuals}, we present visual results comparing 2 step
mean-field inference with Gaussian and learned pairwise CRF potentials. In
general, we observe that the pixels belonging to dominant classes in the
training data are being more accurately classified with learned CRF. This leads to
a significant improvements in overall pixel accuracy. This also results
in a slight decrease of the accuracy from less frequent class pixels thereby
slightly reducing the average class accuracy with learning. We attribute this
to the type of annotation that is available for this dataset, which is not
for the entire image but for some segments in the image. We have very few
images of the infrequent classes to combat this behaviour during training.

\subsection{Experiment Protocols}
\label{sec:protocols}

Table~\ref{tbl:parameters} shows experiment protocols of different experiments.

\definecolor{voc_1}{RGB}{0, 0, 0}
\definecolor{voc_2}{RGB}{128, 0, 0}
\definecolor{voc_3}{RGB}{0, 128, 0}
\definecolor{voc_4}{RGB}{128, 128, 0}
\definecolor{voc_5}{RGB}{0, 0, 128}
\definecolor{voc_6}{RGB}{128, 0, 128}
\definecolor{voc_7}{RGB}{0, 128, 128}
\definecolor{voc_8}{RGB}{128, 128, 128}
\definecolor{voc_9}{RGB}{64, 0, 0}
\definecolor{voc_10}{RGB}{192, 0, 0}
\definecolor{voc_11}{RGB}{64, 128, 0}
\definecolor{voc_12}{RGB}{192, 128, 0}
\definecolor{voc_13}{RGB}{64, 0, 128}
\definecolor{voc_14}{RGB}{192, 0, 128}
\definecolor{voc_15}{RGB}{64, 128, 128}
\definecolor{voc_16}{RGB}{192, 128, 128}
\definecolor{voc_17}{RGB}{0, 64, 0}
\definecolor{voc_18}{RGB}{128, 64, 0}
\definecolor{voc_19}{RGB}{0, 192, 0}
\definecolor{voc_20}{RGB}{128, 192, 0}
\definecolor{voc_21}{RGB}{0, 64, 128}
\definecolor{voc_22}{RGB}{128, 64, 128}

\begin{figure*}[t]
  \centering
  \fcolorbox{white}{voc_1}{\rule{0pt}{6pt}\rule{6pt}{0pt}} Background~~
  \fcolorbox{white}{voc_2}{\rule{0pt}{6pt}\rule{6pt}{0pt}} Aeroplane~~
  \fcolorbox{white}{voc_3}{\rule{0pt}{6pt}\rule{6pt}{0pt}} Bicycle~~
  \fcolorbox{white}{voc_4}{\rule{0pt}{6pt}\rule{6pt}{0pt}} Bird~~
  \fcolorbox{white}{voc_5}{\rule{0pt}{6pt}\rule{6pt}{0pt}} Boat~~
  \fcolorbox{white}{voc_6}{\rule{0pt}{6pt}\rule{6pt}{0pt}} Bottle~~
  \fcolorbox{white}{voc_7}{\rule{0pt}{6pt}\rule{6pt}{0pt}} Bus~~
  \fcolorbox{white}{voc_8}{\rule{0pt}{6pt}\rule{6pt}{0pt}} Car~~ \\
  \fcolorbox{white}{voc_9}{\rule{0pt}{6pt}\rule{6pt}{0pt}} Cat~~
  \fcolorbox{white}{voc_10}{\rule{0pt}{6pt}\rule{6pt}{0pt}} Chair~~
  \fcolorbox{white}{voc_11}{\rule{0pt}{6pt}\rule{6pt}{0pt}} Cow~~
  \fcolorbox{white}{voc_12}{\rule{0pt}{6pt}\rule{6pt}{0pt}} Dining Table~~
  \fcolorbox{white}{voc_13}{\rule{0pt}{6pt}\rule{6pt}{0pt}} Dog~~
  \fcolorbox{white}{voc_14}{\rule{0pt}{6pt}\rule{6pt}{0pt}} Horse~~
  \fcolorbox{white}{voc_15}{\rule{0pt}{6pt}\rule{6pt}{0pt}} Motorbike~~
  \fcolorbox{white}{voc_16}{\rule{0pt}{6pt}\rule{6pt}{0pt}} Person~~ \\
  \fcolorbox{white}{voc_17}{\rule{0pt}{6pt}\rule{6pt}{0pt}} Potted Plant~~
  \fcolorbox{white}{voc_18}{\rule{0pt}{6pt}\rule{6pt}{0pt}} Sheep~~
  \fcolorbox{white}{voc_19}{\rule{0pt}{6pt}\rule{6pt}{0pt}} Sofa~~
  \fcolorbox{white}{voc_20}{\rule{0pt}{6pt}\rule{6pt}{0pt}} Train~~
  \fcolorbox{white}{voc_21}{\rule{0pt}{6pt}\rule{6pt}{0pt}} TV monitor~~ \\
  \subfigure{%
    \includegraphics[width=.4\columnwidth]{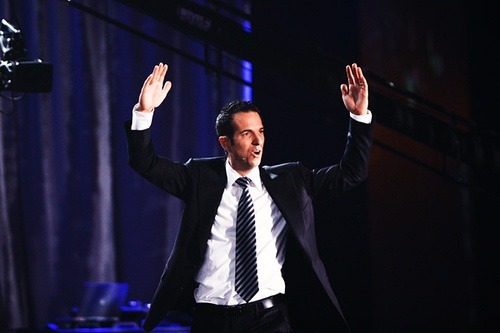}
  }
  \subfigure{%
    \includegraphics[width=.4\columnwidth]{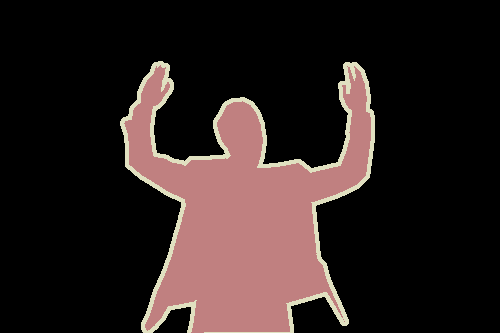}
  }
  \subfigure{%
    \includegraphics[width=.4\columnwidth]{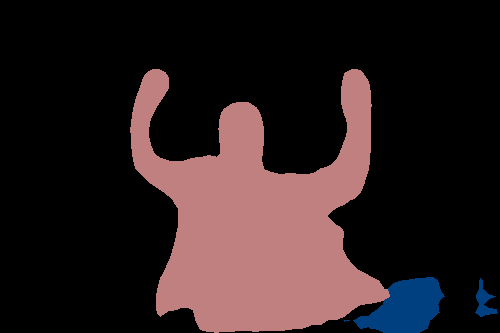}
  }
  \subfigure{%
    \includegraphics[width=.4\columnwidth]{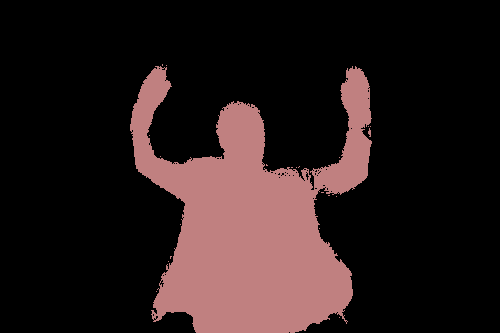}
  }
  \subfigure{%
    \includegraphics[width=.4\columnwidth]{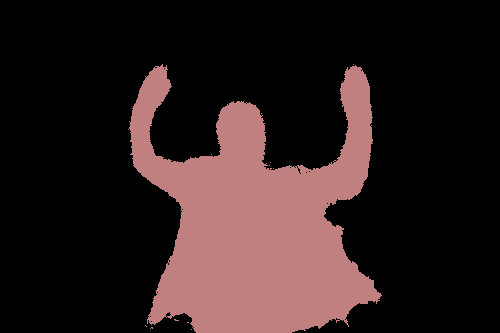}
  }\\
  \subfigure{%
    \includegraphics[width=.4\columnwidth]{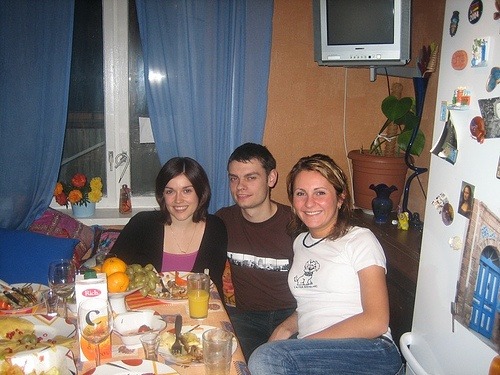}
  }
  \subfigure{%
    \includegraphics[width=.4\columnwidth]{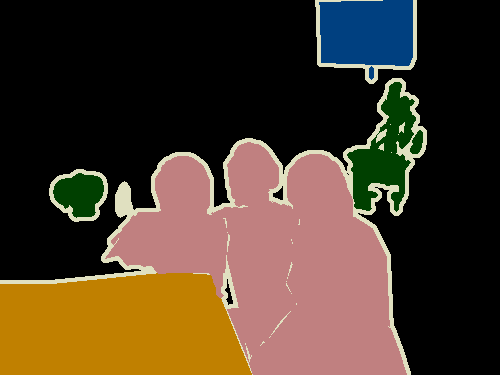}
  }
  \subfigure{%
    \includegraphics[width=.4\columnwidth]{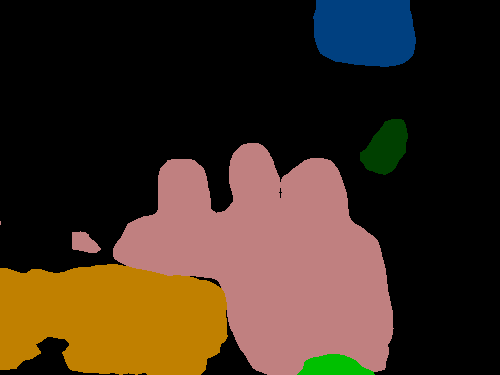}
  }
  \subfigure{%
    \includegraphics[width=.4\columnwidth]{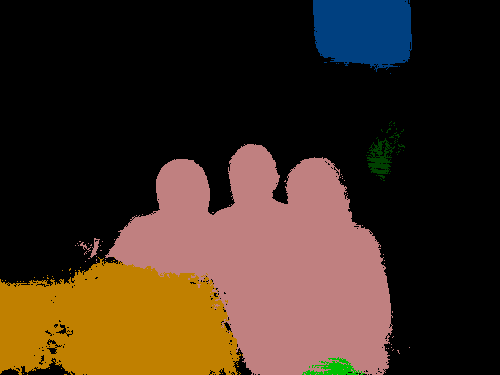}
  }
  \subfigure{%
    \includegraphics[width=.4\columnwidth]{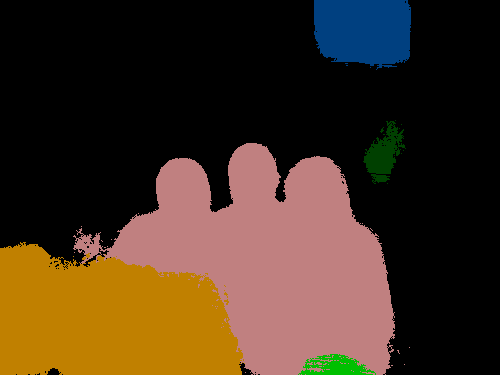}
  }\\
    \subfigure{%
    \includegraphics[width=.4\columnwidth]{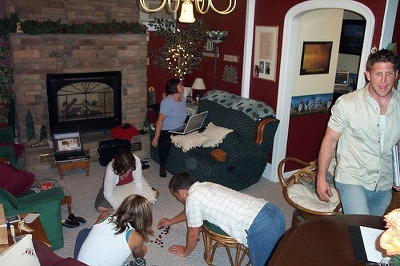}
  }
  \subfigure{%
    \includegraphics[width=.4\columnwidth]{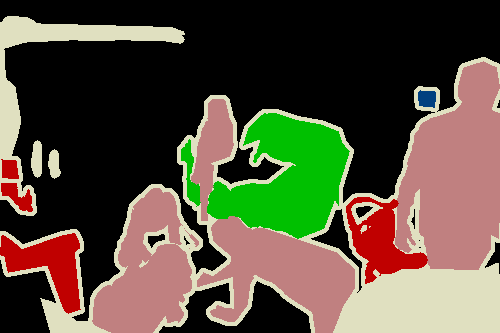}
  }
  \subfigure{%
    \includegraphics[width=.4\columnwidth]{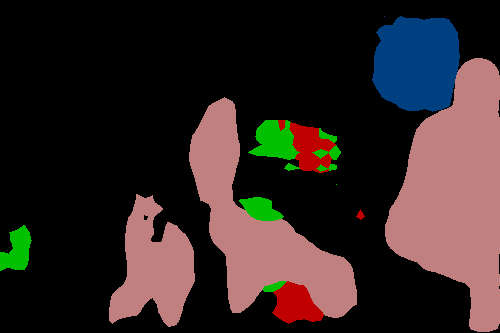}
  }
  \subfigure{%
    \includegraphics[width=.4\columnwidth]{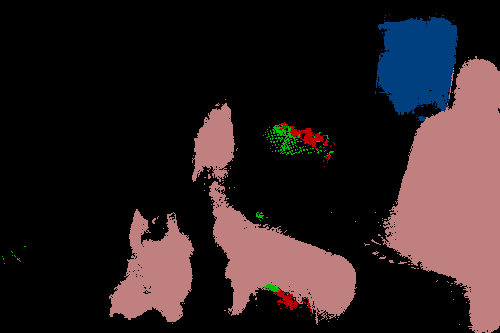}
  }
  \subfigure{%
    \includegraphics[width=.4\columnwidth]{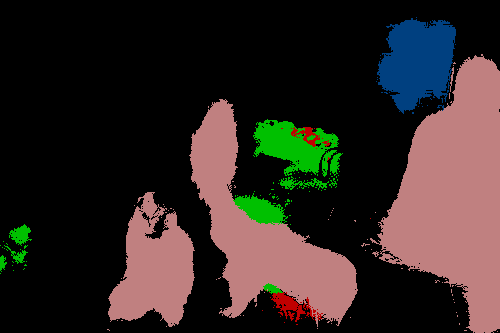}
  }\\
   \subfigure{%
    \includegraphics[width=.4\columnwidth]{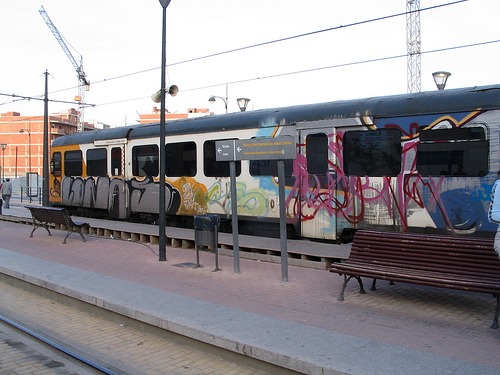}
  }
  \subfigure{%
    \includegraphics[width=.4\columnwidth]{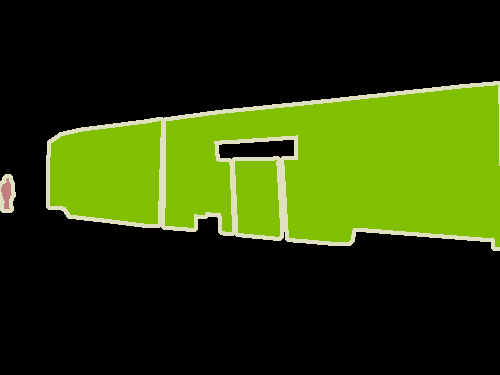}
  }
  \subfigure{%
    \includegraphics[width=.4\columnwidth]{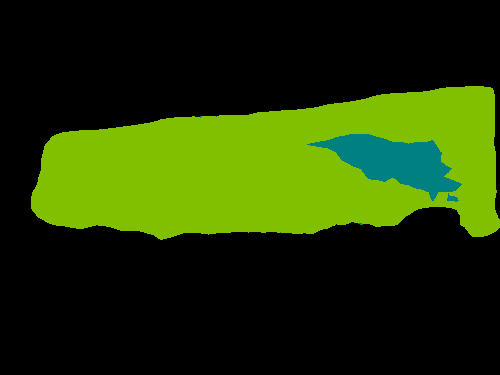}
  }
  \subfigure{%
    \includegraphics[width=.4\columnwidth]{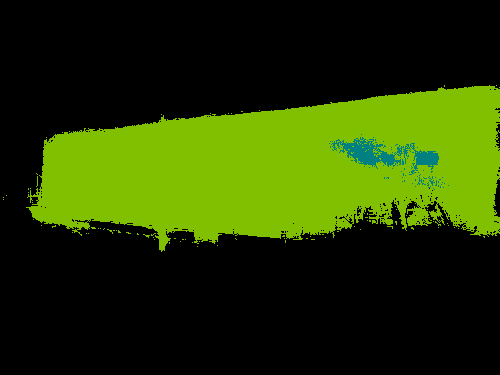}
  }
  \subfigure{%
    \includegraphics[width=.4\columnwidth]{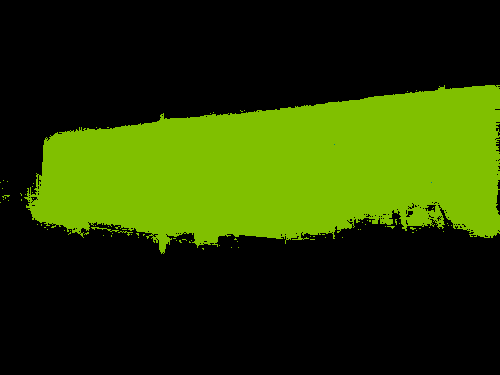}
  }\\
     \subfigure{%
    \includegraphics[width=.4\columnwidth]{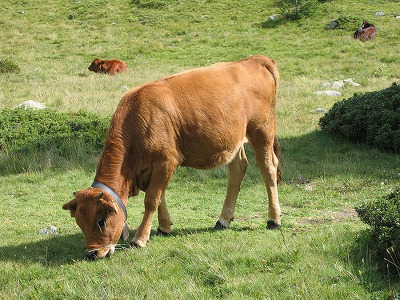}
  }
  \subfigure{%
    \includegraphics[width=.4\columnwidth]{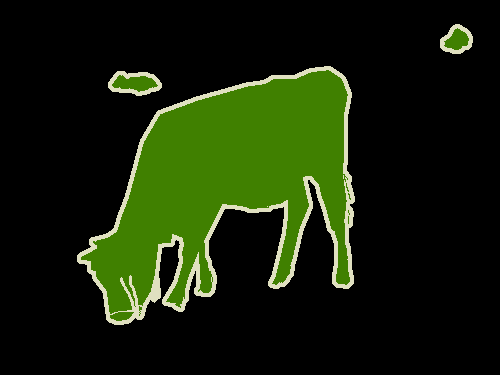}
  }
  \subfigure{%
    \includegraphics[width=.4\columnwidth]{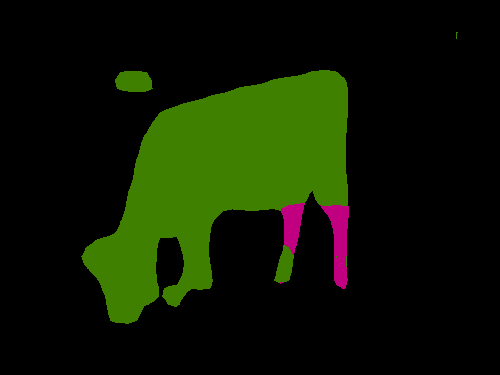}
  }
  \subfigure{%
    \includegraphics[width=.4\columnwidth]{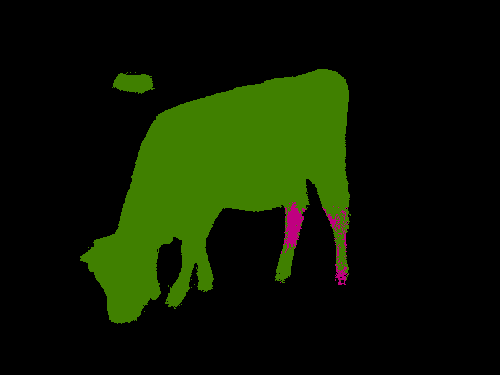}
  }
  \subfigure{%
    \includegraphics[width=.4\columnwidth]{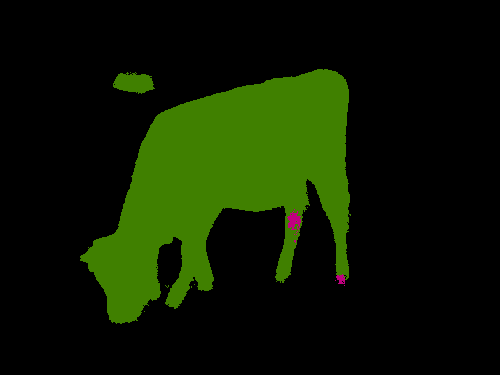}
  }\\
       \subfigure{%
    \includegraphics[width=.4\columnwidth]{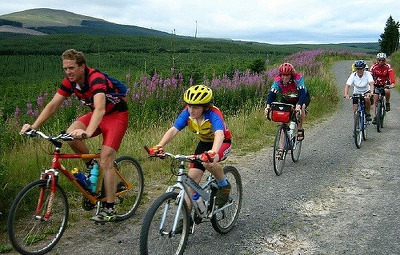}
  }
  \subfigure{%
    \includegraphics[width=.4\columnwidth]{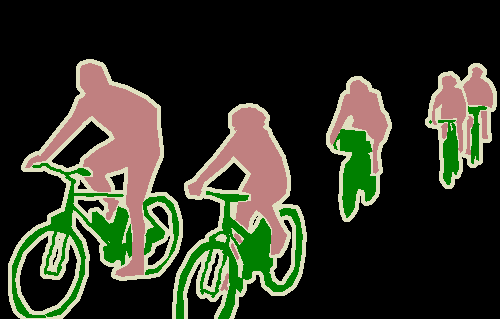}
  }
  \subfigure{%
    \includegraphics[width=.4\columnwidth]{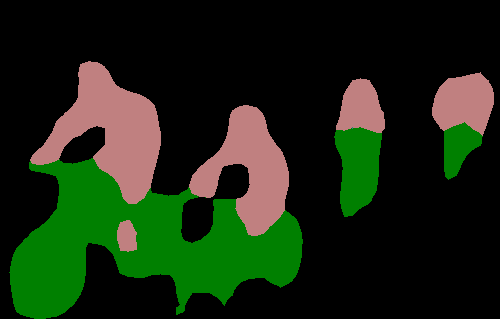}
  }
  \subfigure{%
    \includegraphics[width=.4\columnwidth]{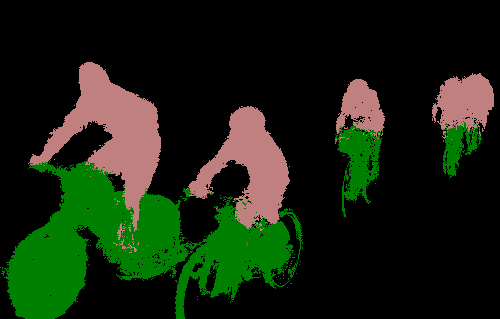}
  }
  \subfigure{%
    \includegraphics[width=.4\columnwidth]{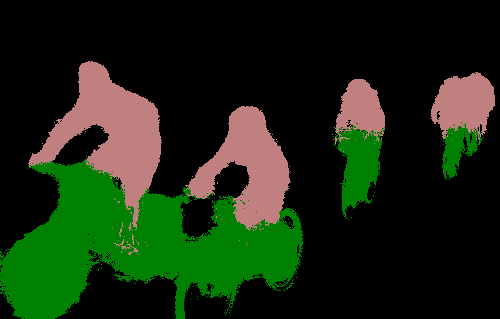}
  }\\
  \setcounter{subfigure}{0}
  \subfigure[Input]{%
    \includegraphics[width=.4\columnwidth]{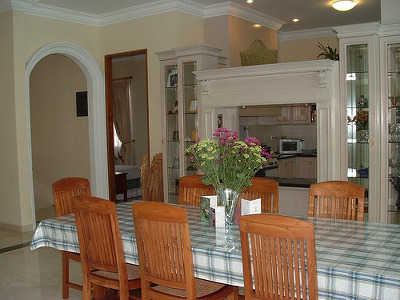}
  }
  \subfigure[Ground Truth]{%
    \includegraphics[width=.4\columnwidth]{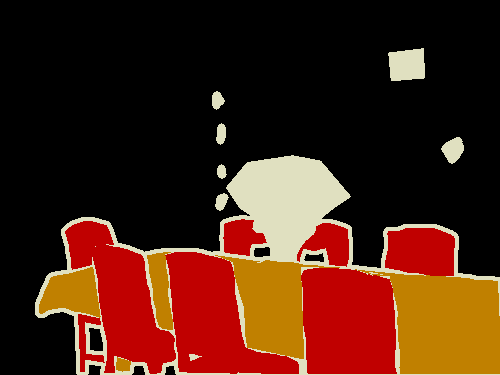}
  }
  \subfigure[DeepLab]{%
    \includegraphics[width=.4\columnwidth]{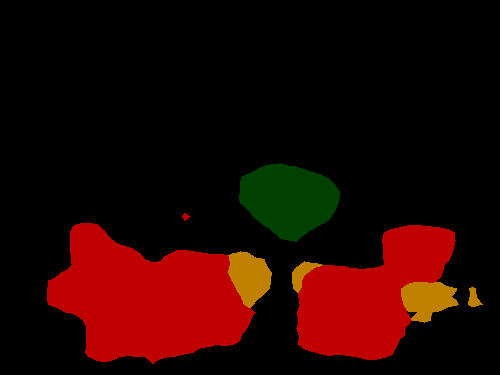}
  }
  \subfigure[+2stepMF-GaussCRF]{%
    \includegraphics[width=.4\columnwidth]{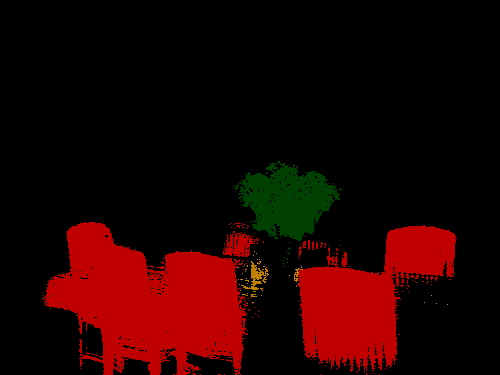}
  }
  \subfigure[+2stepMF-LearnedCRF]{%
    \includegraphics[width=.4\columnwidth]{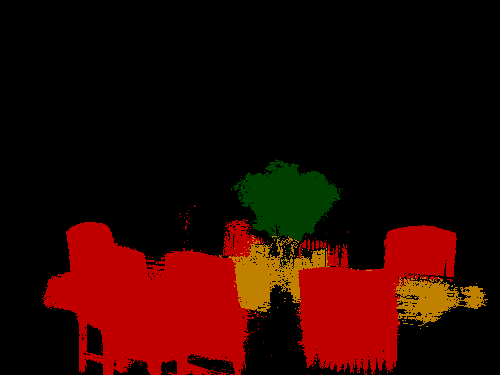}
  }
  \mycaption{Semantic Segmentation}{Example results of semantic segmentation.
  (c) depicts the unary results before application of MF, (d) after two steps of MF with Gaussian edge CRF potentials, (e) after
  two steps of MF with learned edge CRF potentials.}
    \label{fig:semantic_visuals}
    \vspace{-0.3cm}
\end{figure*}

\definecolor{minc_1}{HTML}{771111}
\definecolor{minc_2}{HTML}{CAC690}
\definecolor{minc_3}{HTML}{EEEEEE}
\definecolor{minc_4}{HTML}{7C8FA6}
\definecolor{minc_5}{HTML}{597D31}
\definecolor{minc_6}{HTML}{104410}
\definecolor{minc_7}{HTML}{BB819C}
\definecolor{minc_8}{HTML}{D0CE48}
\definecolor{minc_9}{HTML}{622745}
\definecolor{minc_10}{HTML}{666666}
\definecolor{minc_11}{HTML}{D54A31}
\definecolor{minc_12}{HTML}{101044}
\definecolor{minc_13}{HTML}{444126}
\definecolor{minc_14}{HTML}{75D646}
\definecolor{minc_15}{HTML}{DD4348}
\definecolor{minc_16}{HTML}{5C8577}
\definecolor{minc_17}{HTML}{C78472}
\definecolor{minc_18}{HTML}{75D6D0}
\definecolor{minc_19}{HTML}{5B4586}
\definecolor{minc_20}{HTML}{C04393}
\definecolor{minc_21}{HTML}{D69948}
\definecolor{minc_22}{HTML}{7370D8}
\definecolor{minc_23}{HTML}{7A3622}
\definecolor{minc_24}{HTML}{000000}

\begin{figure*}[t]
  \centering
  \fcolorbox{white}{minc_1}{\rule{0pt}{6pt}\rule{6pt}{0pt}} Brick~~
  \fcolorbox{white}{minc_2}{\rule{0pt}{6pt}\rule{6pt}{0pt}} Carpet~~
  \fcolorbox{white}{minc_3}{\rule{0pt}{6pt}\rule{6pt}{0pt}} Ceramic~~
  \fcolorbox{white}{minc_4}{\rule{0pt}{6pt}\rule{6pt}{0pt}} Fabric~~
  \fcolorbox{white}{minc_5}{\rule{0pt}{6pt}\rule{6pt}{0pt}} Foliage~~
  \fcolorbox{white}{minc_6}{\rule{0pt}{6pt}\rule{6pt}{0pt}} Food~~
  \fcolorbox{white}{minc_7}{\rule{0pt}{6pt}\rule{6pt}{0pt}} Glass~~
  \fcolorbox{white}{minc_8}{\rule{0pt}{6pt}\rule{6pt}{0pt}} Hair~~ \\
  \fcolorbox{white}{minc_9}{\rule{0pt}{6pt}\rule{6pt}{0pt}} Leather~~
  \fcolorbox{white}{minc_10}{\rule{0pt}{6pt}\rule{6pt}{0pt}} Metal~~
  \fcolorbox{white}{minc_11}{\rule{0pt}{6pt}\rule{6pt}{0pt}} Mirror~~
  \fcolorbox{white}{minc_12}{\rule{0pt}{6pt}\rule{6pt}{0pt}} Other~~
  \fcolorbox{white}{minc_13}{\rule{0pt}{6pt}\rule{6pt}{0pt}} Painted~~
  \fcolorbox{white}{minc_14}{\rule{0pt}{6pt}\rule{6pt}{0pt}} Paper~~
  \fcolorbox{white}{minc_15}{\rule{0pt}{6pt}\rule{6pt}{0pt}} Plastic~~
  \fcolorbox{white}{minc_16}{\rule{0pt}{6pt}\rule{6pt}{0pt}} Polished Stone~~ \\
  \fcolorbox{white}{minc_17}{\rule{0pt}{6pt}\rule{6pt}{0pt}} Skin~~
  \fcolorbox{white}{minc_18}{\rule{0pt}{6pt}\rule{6pt}{0pt}} Sky~~
  \fcolorbox{white}{minc_19}{\rule{0pt}{6pt}\rule{6pt}{0pt}} Stone~~
  \fcolorbox{white}{minc_20}{\rule{0pt}{6pt}\rule{6pt}{0pt}} Tile~~
  \fcolorbox{white}{minc_21}{\rule{0pt}{6pt}\rule{6pt}{0pt}} Wallpaper~~
  \fcolorbox{white}{minc_22}{\rule{0pt}{6pt}\rule{6pt}{0pt}} Water~~
  \fcolorbox{white}{minc_23}{\rule{0pt}{6pt}\rule{6pt}{0pt}} Wood~~ \\
  \subfigure{%
    \includegraphics[width=.4\columnwidth]{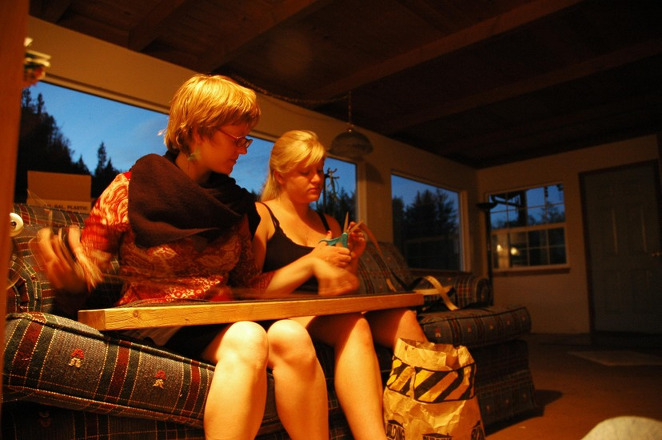}
  }
  \subfigure{%
    \includegraphics[width=.4\columnwidth]{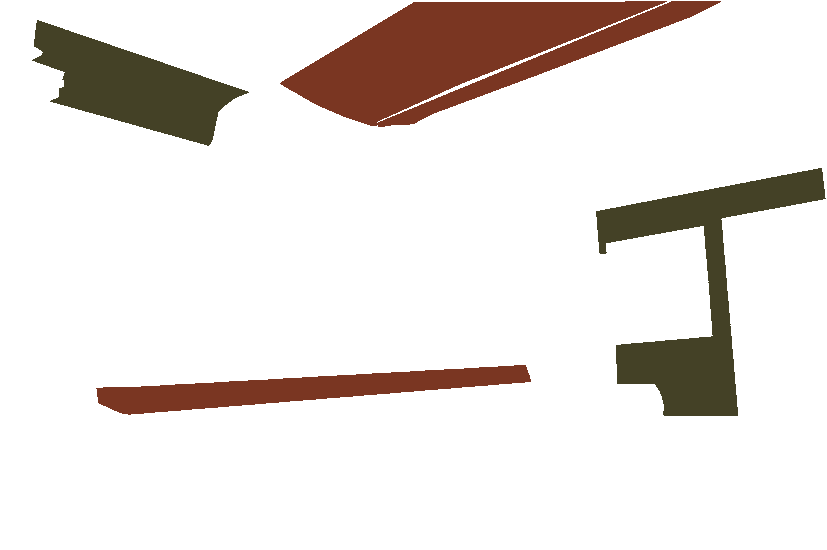}
  }
  \subfigure{%
    \includegraphics[width=.4\columnwidth]{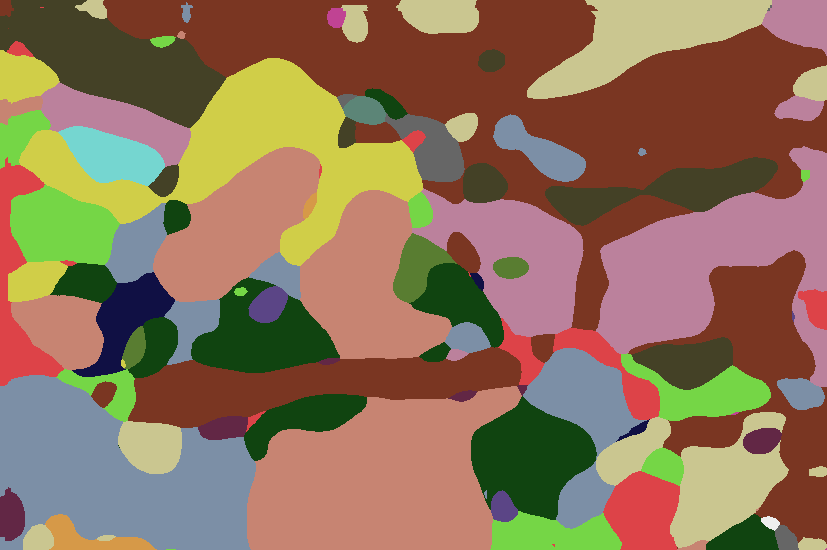}
  }
  \subfigure{%
    \includegraphics[width=.4\columnwidth]{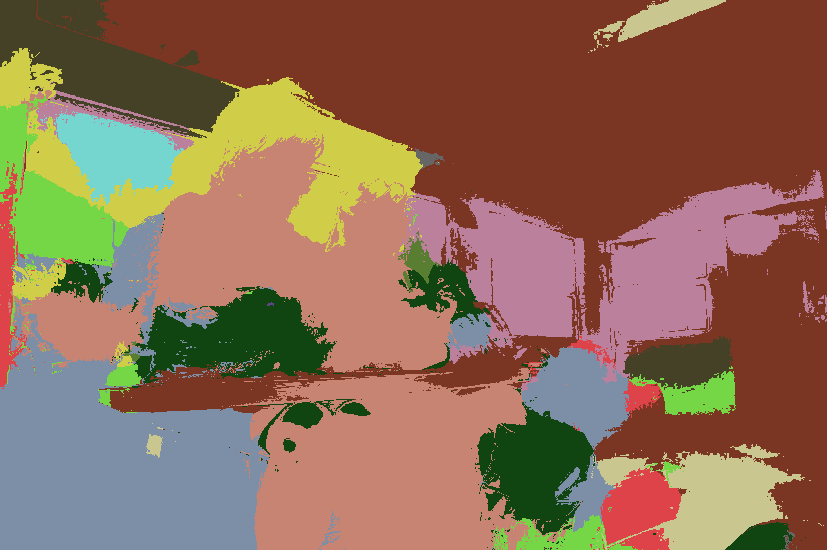}
  }
  \subfigure{%
    \includegraphics[width=.4\columnwidth]{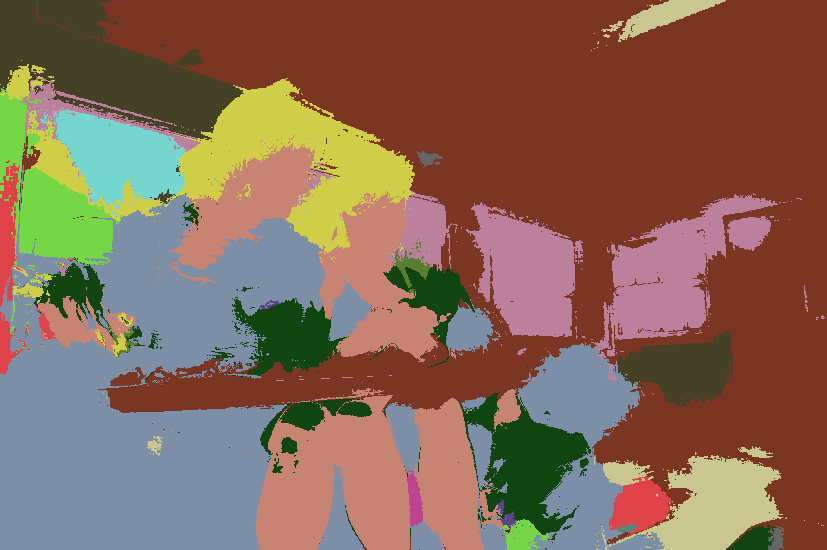}
  }\\[-2ex]
  \subfigure{%
    \includegraphics[width=.4\columnwidth]{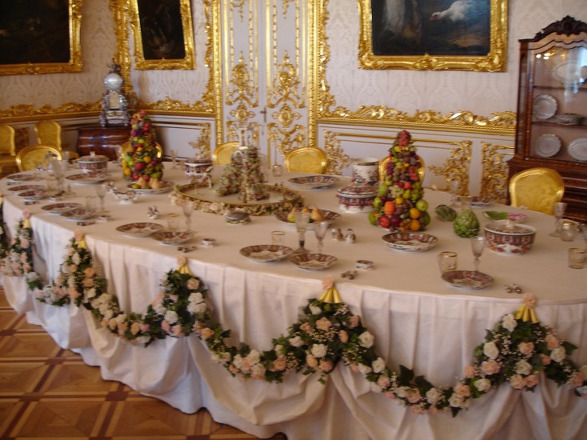}
  }
  \subfigure{%
    \includegraphics[width=.4\columnwidth]{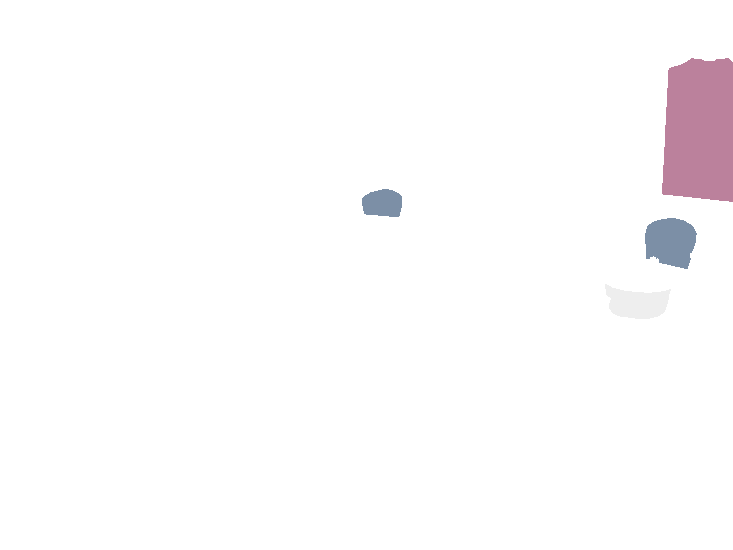}
  }
  \subfigure{%
    \includegraphics[width=.4\columnwidth]{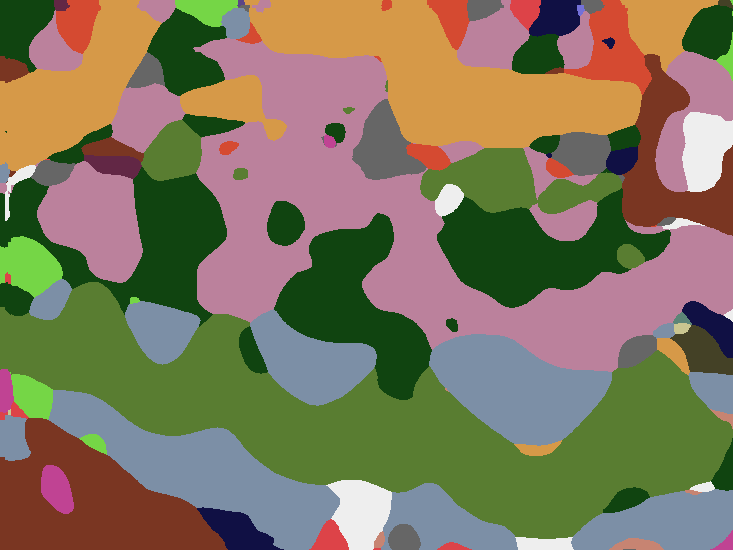}
  }
  \subfigure{%
    \includegraphics[width=.4\columnwidth]{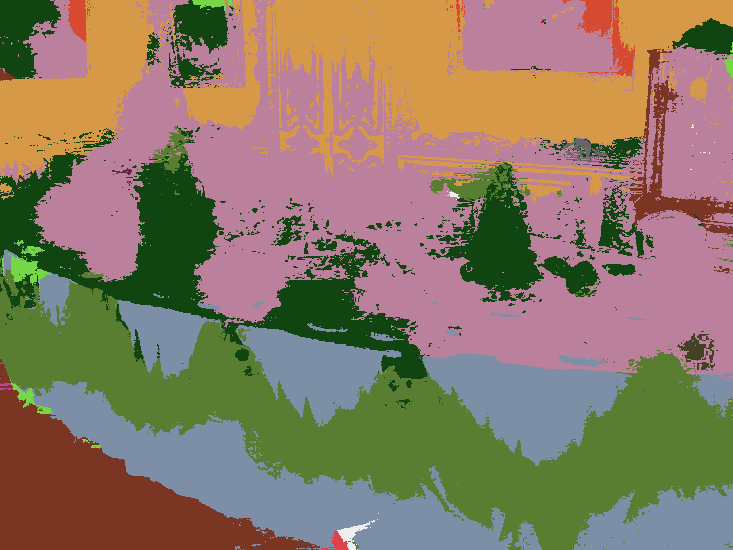}
  }
  \subfigure{%
    \includegraphics[width=.4\columnwidth]{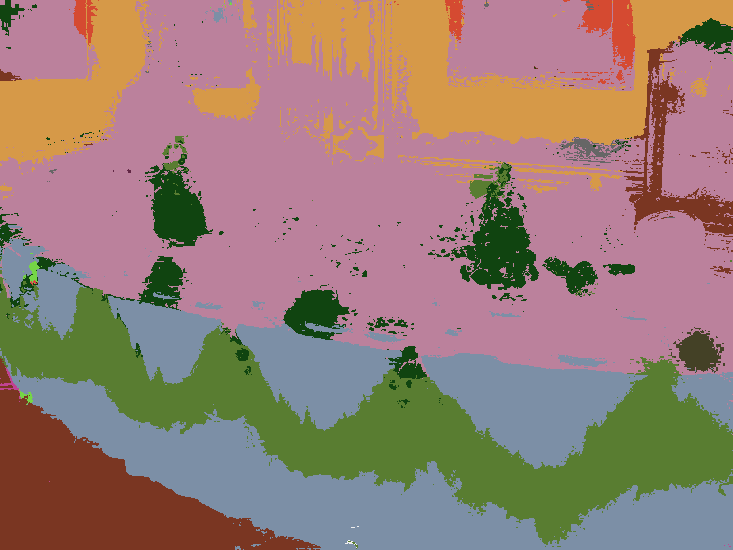}
  }\\[-2ex]
    \subfigure{%
    \includegraphics[width=.4\columnwidth]{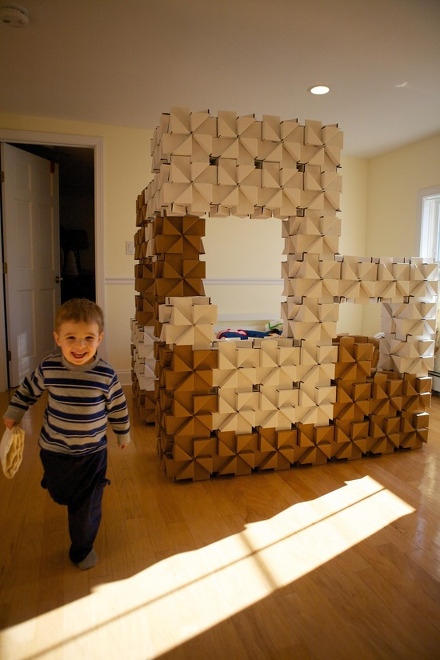}
  }
  \subfigure{%
    \includegraphics[width=.4\columnwidth]{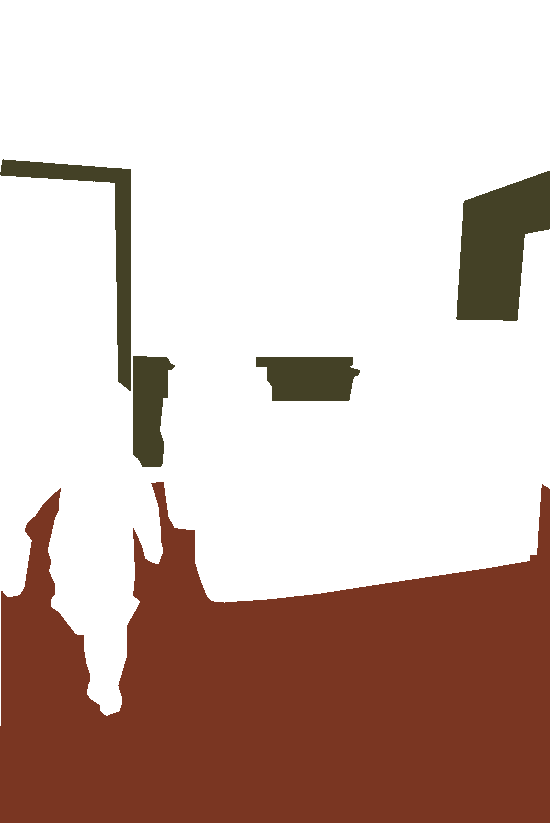}
  }
  \subfigure{%
    \includegraphics[width=.4\columnwidth]{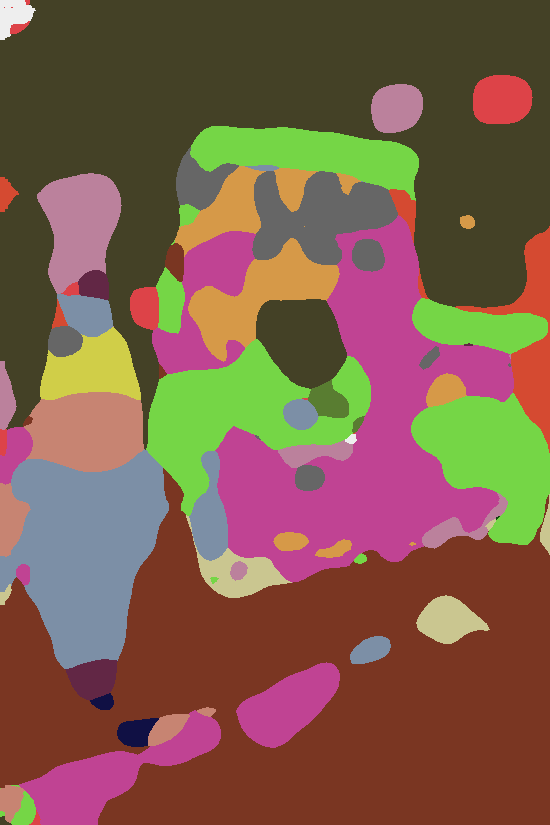}
  }
  \subfigure{%
    \includegraphics[width=.4\columnwidth]{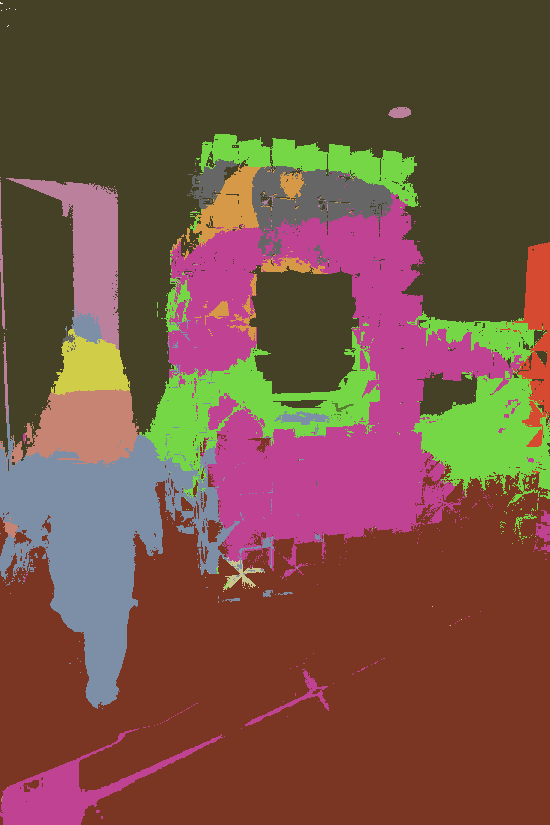}
  }
  \subfigure{%
    \includegraphics[width=.4\columnwidth]{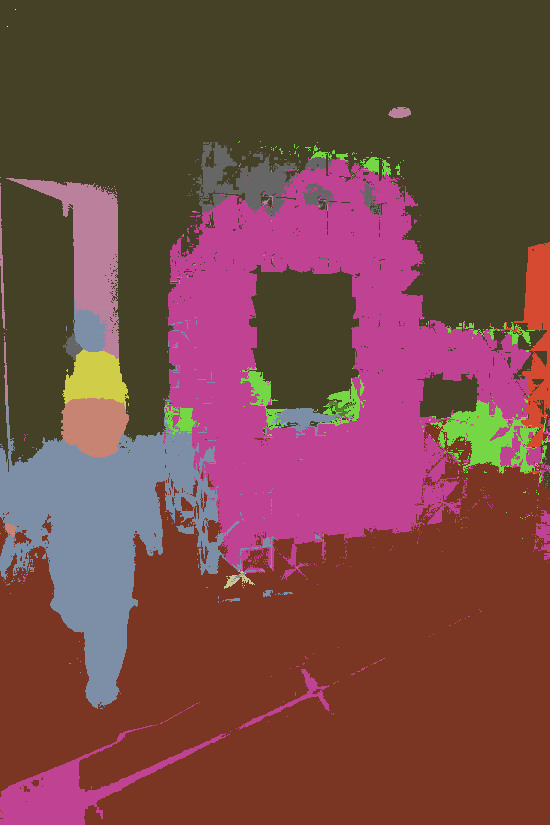}
  }\\[-2ex]
   \subfigure{%
    \includegraphics[width=.4\columnwidth]{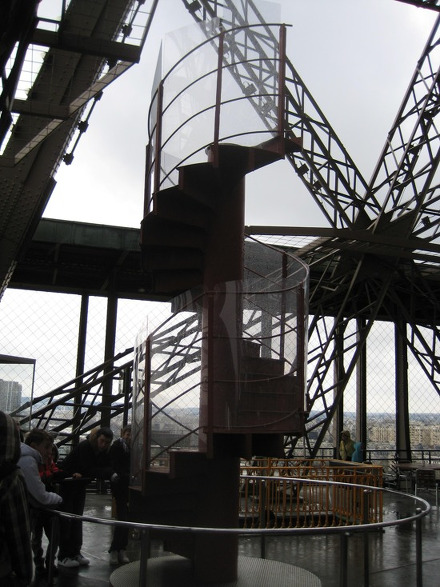}
  }
  \subfigure{%
    \includegraphics[width=.4\columnwidth]{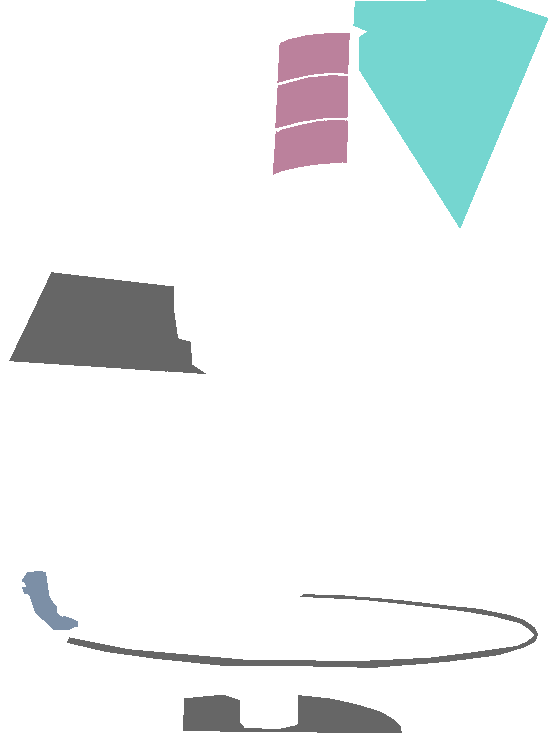}
  }
  \subfigure{%
    \includegraphics[width=.4\columnwidth]{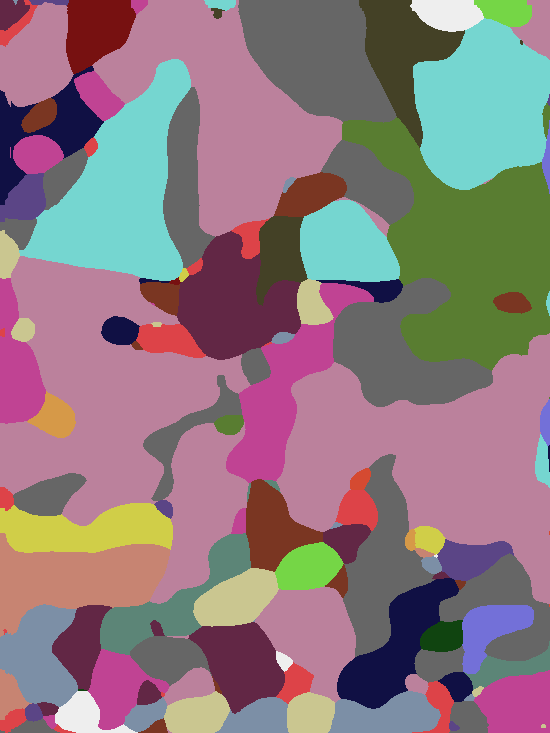}
  }
  \subfigure{%
    \includegraphics[width=.4\columnwidth]{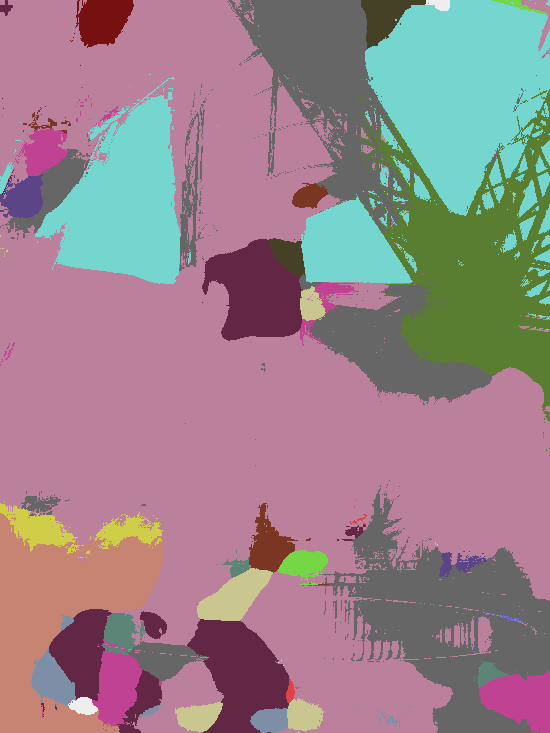}
  }
  \subfigure{%
    \includegraphics[width=.4\columnwidth]{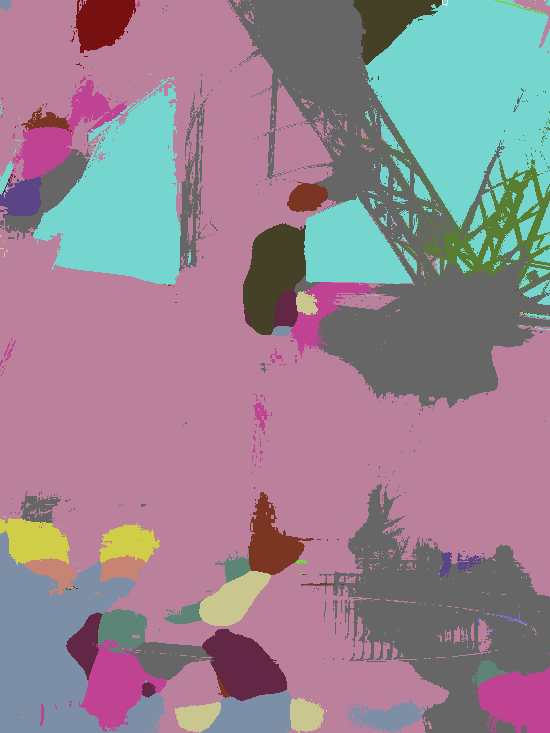}
  }\\[-2ex]
  \setcounter{subfigure}{0}
  \subfigure[Input]{%
    \includegraphics[width=.4\columnwidth]{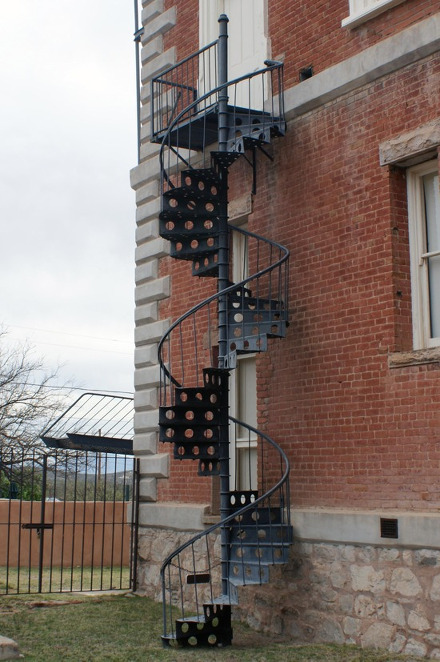}
  }
  \subfigure[Ground Truth]{%
    \includegraphics[width=.4\columnwidth]{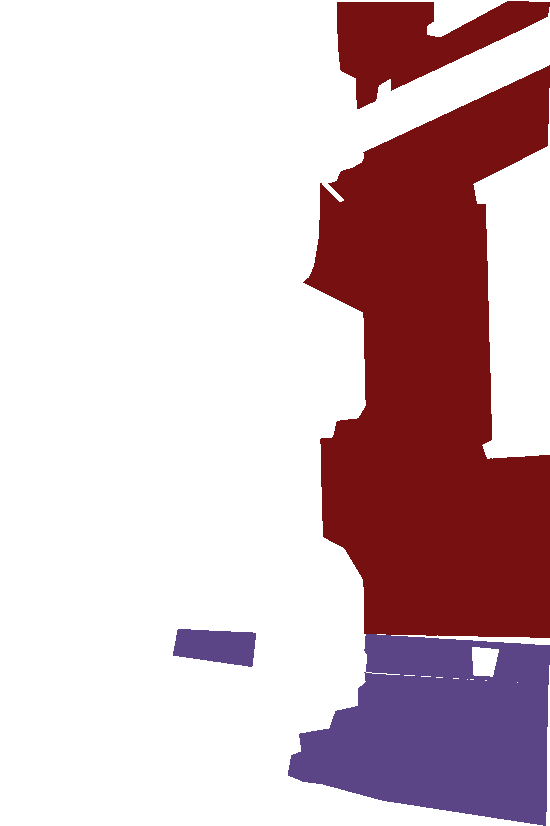}
  }
  \subfigure[DeepLab]{%
    \includegraphics[width=.4\columnwidth]{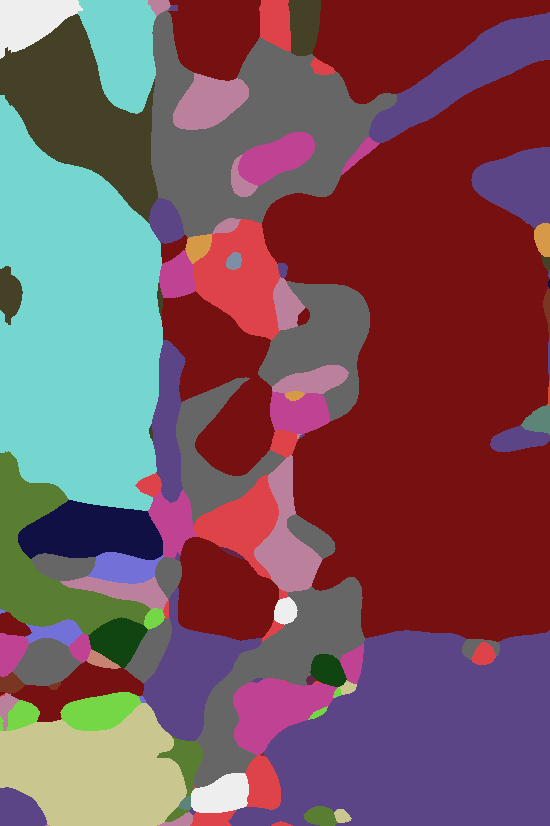}
  }
  \subfigure[+2stepMF-GaussCRF]{%
    \includegraphics[width=.4\columnwidth]{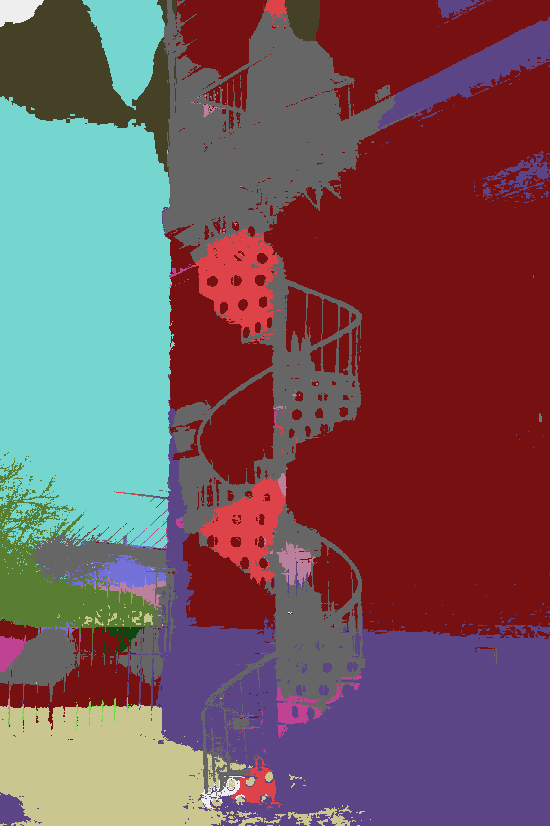}
  }
  \subfigure[+2stepMF-LearnedCRF]{%
    \includegraphics[width=.4\columnwidth]{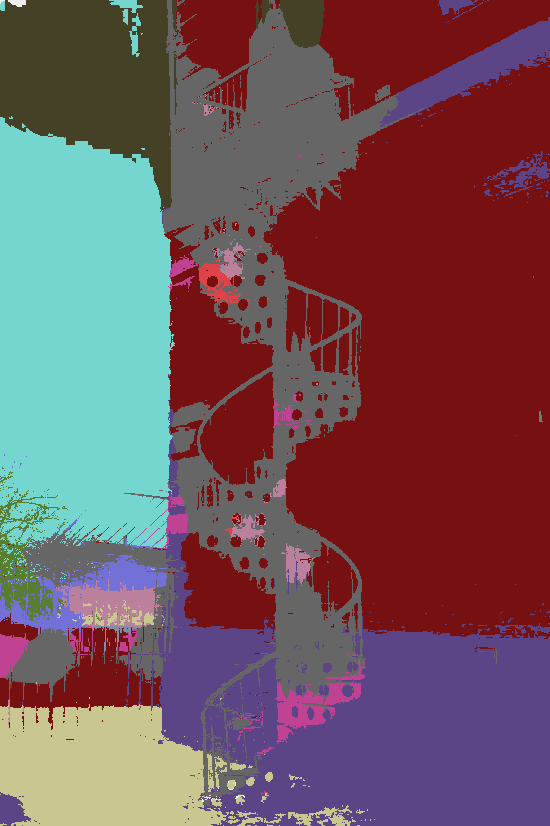}
  }
  \mycaption{Material Segmentation}{Example results of material segmentation.
  (c) depicts the unary results before application of MF, (d) after two steps of MF with Gaussian edge CRF potentials, (e) after
  two steps of MF with learned edge CRF potentials.}
    \label{fig:material_visuals}
    \vspace{-0.3cm}
\end{figure*}

\begin{table*}[h]
\scriptsize
  \centering
    \begin{tabular}{L{2.5cm} L{2.8cm} C{1.6cm} C{0.8cm} C{0.6cm} C{0.7cm} C{0.7cm} C{0.7cm} C{1.6cm} C{0.7cm} C{0.7cm} C{0.7cm}}
      \toprule
& & & & & \multicolumn{3}{c}{\textbf{Data Statistics}} & \multicolumn{4}{c}{\textbf{Training Protocol}} \\

\textbf{Experiment} & \textbf{Feature Types} & \textbf{Feature Scales} & \textbf{Filter Size} & \textbf{Filter Nbr.} & \textbf{Train}  & \textbf{Val.} & \textbf{Test} & \textbf{Loss Type} & \textbf{LR} & \textbf{Batch} & \textbf{Epochs} \\
      \midrule
      \multicolumn{2}{c}{\textbf{Single Bilateral Filter Applications}} & & & & & & & & & \\
      \textbf{2$\times$ Color Upsampling} & Position$_{1}$, Intensity (3D) & 0.13, 0.17 & 65 & 2 & 10581 & 1449 & 1456 & MSE & 1e-06 & 200 & 94.5\\
      \textbf{4$\times$ Color Upsampling} & Position$_{1}$, Intensity (3D) & 0.06, 0.17 & 65 & 2 & 10581 & 1449 & 1456 & MSE & 1e-06 & 200 & 94.5\\
      \textbf{8$\times$ Color Upsampling} & Position$_{1}$, Intensity (3D) & 0.03, 0.17 & 65 & 2 & 10581 & 1449 & 1456 & MSE & 1e-06 & 200 & 94.5\\
      \textbf{16$\times$ Color Upsampling} & Position$_{1}$, Intensity (3D) & 0.02, 0.17 & 65 & 2 & 10581 & 1449 & 1456 & MSE & 1e-06 & 200 & 94.5\\
      \textbf{Depth Upsampling} & Position$_{1}$, Color (5D) & 0.05, 0.02 & 665 & 2 & 795 & 100 & 654 & MSE & 1e-07 & 50 & 251.6\\
      \textbf{Mesh Denoising} & Isomap (4D) & 46.00 & 63 & 2 & 1000 & 200 & 500 & MSE & 100 & 10 & 100.0 \\
      \midrule
      \multicolumn{2}{c}{\textbf{DenseCRF Applications}} & & & & & & & & &\\
      \multicolumn{2}{l}{\textbf{Semantic Segmentation}} & & & & & & & & &\\
      \textbf{- 1step MF} & Position$_{1}$, Color (5D); Position$_{1}$ (2D) & 0.01, 0.34; 0.34  & 665; 19  & 2; 2 & 10581 & 1449 & 1456 & Logistic & 0.1 & 5 & 1.4 \\
      \textbf{- 2step MF} & Position$_{1}$, Color (5D); Position$_{1}$ (2D) & 0.01, 0.34; 0.34 & 665; 19 & 2; 2 & 10581 & 1449 & 1456 & Logistic & 0.1 & 5 & 1.4 \\
      \textbf{- \textit{loose} 2step MF} & Position$_{1}$, Color (5D); Position$_{1}$ (2D) & 0.01, 0.34; 0.34 & 665; 19 & 2; 2 &10581 & 1449 & 1456 & Logistic & 0.1 & 5 & +1.9  \\ \\
      \multicolumn{2}{l}{\textbf{Material Segmentation}} & & & & & & & & &\\
      \textbf{- 1step MF} & Position$_{2}$, Lab-Color (5D) & 5.00, 0.05, 0.30  & 665 & 2 & 928 & 150 & 1798 & Weighted Logistic & 1e-04 & 24 & 2.6 \\
      \textbf{- 2step MF} & Position$_{2}$, Lab-Color (5D) & 5.00, 0.05, 0.30 & 665 & 2 & 928 & 150 & 1798 & Weighted Logistic & 1e-04 & 12 & +0.7 \\
      \textbf{- \textit{loose} 2step MF} & Position$_{2}$, Lab-Color (5D) & 5.00, 0.05, 0.30 & 665 & 2 & 928 & 150 & 1798 & Weighted Logistic & 1e-04 & 12 & +0.2\\
      \midrule
      \multicolumn{2}{c}{\textbf{Neural Network Applications}} & & & & & & & & &\\
      \textbf{Tiles: CNN-9$\times$9} & - & - & 81 & 4 & 10000 & 1000 & 1000 & Logistic & 0.01 & 100 & 500.0 \\
      \textbf{Tiles: CNN-13$\times$13} & - & - & 169 & 6 & 10000 & 1000 & 1000 & Logistic & 0.01 & 100 & 500.0 \\
      \textbf{Tiles: CNN-17$\times$17} & - & - & 289 & 8 & 10000 & 1000 & 1000 & Logistic & 0.01 & 100 & 500.0 \\
      \textbf{Tiles: CNN-21$\times$21} & - & - & 441 & 10 & 10000 & 1000 & 1000 & Logistic & 0.01 & 100 & 500.0 \\
      \textbf{Tiles: BNN} & Position$_{1}$, Color (5D) & 0.05, 0.04 & 63 & 1 & 10000 & 1000 & 1000 & Logistic & 0.01 & 100 & 30.0 \\
      \textbf{LeNet} & - & - & 25 & 2 & 5490 & 1098 & 1647 & Logistic & 0.1 & 100 & 182.2 \\
      \textbf{Crop-LeNet} & - & - & 25 & 2 & 5490 & 1098 & 1647 & Logistic & 0.1 & 100 & 182.2 \\
      \textbf{BNN-LeNet} & Position$_{2}$ (2D) & 20.00 & 7 & 1 & 5490 & 1098 & 1647 & Logistic & 0.1 & 100 & 182.2 \\
      \textbf{DeepCNet} & - & - & 9 & 1 & 5490 & 1098 & 1647 & Logistic & 0.1 & 100 & 182.2 \\
      \textbf{Crop-DeepCNet} & - & - & 9 & 1 & 5490 & 1098 & 1647 & Logistic & 0.1 & 100 & 182.2 \\
      \textbf{BNN-DeepCNet} & Position$_{2}$ (2D) & 40.00  & 7 & 1 & 5490 & 1098 & 1647 & Logistic & 0.1 & 100 & 182.2 \\
      \bottomrule
      \\
    \end{tabular}
    \mycaption{Experiment Protocols} {Experiment protocols for the different experiments presented in this work. \textbf{Feature Types}:
    Feature spaces used for the bilateral convolutions. Position$_1$ corresponds to un-normalized pixel positions whereas Position$_2$ corresponds
    to pixel positions normalized to $[0,1]$ with respect to the given image. \textbf{Feature Scales}: Cross-validated scales for the features used.
     \textbf{Filter Size}: Number of elements in the filter that is being learned. \textbf{Filter Nbr.}: Half-width of the filter. \textbf{Train},
     \textbf{Val.} and \textbf{Test} corresponds to the number of train, validation and test images used in the experiment. \textbf{Loss Type}: Type
     of loss used for back-propagation. ``MSE'' corresponds to Euclidean mean squared error loss and ``Logistic'' corresponds to multinomial logistic
     loss. ``Weighted Logistic'' is the class-weighted multinomial logistic loss. We weighted the loss with inverse class probability for material
     segmentation task due to the small availability of training data with class imbalance. \textbf{LR}: Fixed learning rate used in stochastic gradient
     descent. \textbf{Batch}: Number of images used in one parameter update step. \textbf{Epochs}: Number of training epochs. In all the experiments,
     we used fixed momentum of 0.9 and weight decay of 0.0005 for stochastic gradient descent. ```Color Upsampling'' experiments in this Table corresponds
     to those performed on Pascal VOC12 dataset images. For all experiments using Pascal VOC12 images, we use extended
     training segmentation dataset available from~\cite{hariharan2011moredata}, and used standard validation and test splits
     from the main dataset~\cite{voc2012segmentation}.}
  \label{tbl:parameters}
\end{table*}

\end{document}